%% file: main.tex
\documentclass{article}
\pdfoutput=1
\usepackage{microtype}
\usepackage{graphicx}
\usepackage{subfigure}
\usepackage{booktabs}

\usepackage[accepted]{icml2024}

\usepackage{amssymb}
\usepackage{mathtools}
\usepackage{amsthm}
\usepackage{diagbox}

\usepackage[capitalize,noabbrev]{cleveref}

\usepackage[textsize=tiny]{todonotes}
\input{def}
\input{math_common}

\icmltitlerunning{Generative Models are Self-Watermarked: Declaring Model Authentication through Re-Generation}

\begin{document}

\twocolumn[
\icmltitle{Generative Models are Self-Watermarked: 
\\
Declaring Model Authentication through Re-Generation}



\icmlsetsymbol{equal}{*}

\begin{icmlauthorlist}
\icmlauthor{Aditya Desu}{equal,UOM}
\icmlauthor{Xuanli He}{equal,UCL}
\icmlauthor{Qiongkai Xu}{MQ}
\icmlauthor{Wei Lu}{SUTD}

\end{icmlauthorlist}

\icmlaffiliation{UOM}{School of Computing and Information System, University of Melbourne, Melbourne, Australia}
\icmlaffiliation{UCL}{Department of Computer Science, University College London, United Kingdom}
\icmlaffiliation{MQ}{School of Computing, FSE, Macquarie University, Sydney, Australia}
\icmlaffiliation{SUTD}{StatNLP Research Group, Singapore University of Technology and Design, Singapore}

\icmlcorrespondingauthor{Qiongkai Xu}{qiongkai.xu@mq.edu.au}


\icmlkeywords{Machine Learning, ICML}

\vskip 0.3in
]



\printAffiliationsAndNotice{\icmlEqualContribution} 


\input{sec_abstract}

\input{sec1_introduction}

\input{sec2_related_work}

\input{sec_background}

\input{sec3_method}

\input{sec4_experiment}

\input{sec6_conclusion}


\bibliography{custom}
\bibliographystyle{icml2024}

\newpage
\appendix
\onecolumn

\input{sec_appendix}

\end{document}

%% file: def.tex
\usepackage{stmaryrd}
\usepackage{amsfonts}
\usepackage{amsmath}
\usepackage{bbm}
\usepackage{multirow,tabularx}
\usepackage{xspace}
\usepackage{todonotes}
\usepackage{booktabs}
\usepackage{subcaption}
\usepackage{algorithm}
\usepackage{algpseudocode}
\usepackage{enumerate}
\usepackage{makecell}
\usepackage{amsfonts}
\usepackage{svg}
\usepackage{colortbl}

\usepackage{graphicx}
\usepackage{wrapfig}

\usepackage{mathtools}

\usepackage{soul}

\newtheorem{theorem}{Theorem}

\newtheorem{definition}{Definition} 

\def\eg{{\em e.g.,}\xspace}
\def\ie{{\em i.e.,}\xspace}

\def\etc{{\em etc.}\xspace}

\def\mone{{M2M}\xspace}
\def\mtwo{{mBART}\xspace}
\def\mthree{{GPT3.5-turbo}\xspace}
\def\mfour{{Cohere}\xspace}
\def\mfive{{GPT4}\xspace}

\def\SDvTwoOne{{SDv~2.1}\xspace}
\def\SDXLOneZero{{SDXL~1.0}\xspace}
\def\SDvTwo{{SD~v2}\xspace}
\def\SDvTwoOneB{{SD~v2.1B}\xspace}
\def\SDXLZeroNineB{{SDXL0.9B}\xspace}
\def\SDOneFour{{SD~v1.4}\xspace}
\def\BKSDM{{BKSDM}\xspace}

%% file: math_common.tex
\usepackage{amsmath,amsfonts,bm}

\def\Figref#1{Figure~\ref{#1}}

\def\Tabref#1{Table~\ref{#1}}

\def\Secref#1{Section~\ref{#1}}

\def\eqref#1{equation~\ref{#1}}

\def\Algref#1{Algorithm~\ref{#1}}

\def\1{\bm{1}}

\def\vx{{\bm{x}}}
\def\vy{{\bm{y}}}

\DeclareMathAlphabet{\mathsfit}{\encodingdefault}{\sfdefault}{m}{sl}
\SetMathAlphabet{\mathsfit}{bold}{\encodingdefault}{\sfdefault}{bx}{n}

\newcommand{\R}{\mathbb{R}}



%% file: sec_abstract.tex
\begin{abstract}
As machine- and AI-generated content proliferates, protecting the intellectual property of generative models has become imperative, yet verifying data ownership poses formidable challenges, particularly in cases of unauthorized reuse of generated data. The challenge of verifying data ownership is further amplified by using Machine Learning as a Service (MLaaS), which often functions as a black-box system. 

Our work is dedicated to detecting data reuse from even an individual sample. Traditionally, watermarking has been leveraged to detect AI-generated content. However, unlike watermarking techniques that embed additional information as triggers into models or generated content, potentially compromising output quality, our approach identifies latent fingerprints inherently present within the outputs through re-generation. We propose an explainable verification procedure that attributes data ownership through re-generation, and further amplifies these fingerprints in the generative models through iterative data re-generation. This methodology is theoretically grounded and demonstrates viability and robustness using recent advanced text and image generative models. 
Our methodology is significant as it goes beyond protecting the intellectual property of APIs and addresses important issues such as the spread of misinformation and academic misconduct. It provides a useful tool to ensure the integrity of sources and authorship, expanding its application in different scenarios where authenticity and ownership verification are essential.


\end{abstract}

%% file: sec1_introduction.tex
\section{Introduction}

In recent years, the emergence of Artificial Intelligence Generated Content (AIGC), including tools like ChatGPT, Claude, DALL-E, Stable Diffusion, Copilot, has marked a significant advancement in the quality of machine-generated content. These generative models are increasingly being offered by companies as part of pay-as-you-use services on cloud platforms. While such development has undeniably accelerated the advancement and dissemination of AI technology, it has simultaneously raised substantial concerns regarding the misuse of these models. A key challenge lies in authenticating the author (or source) of the content generated by these models, which encompasses two primary aspects: (1) protecting the Intellectual Property (IP) of the authentic generator when content is misused and (2) tracing the responsibility of the information source to ensure accountability.

\begin{figure*}[t]
    \centering
    \includegraphics[width=0.99\linewidth]{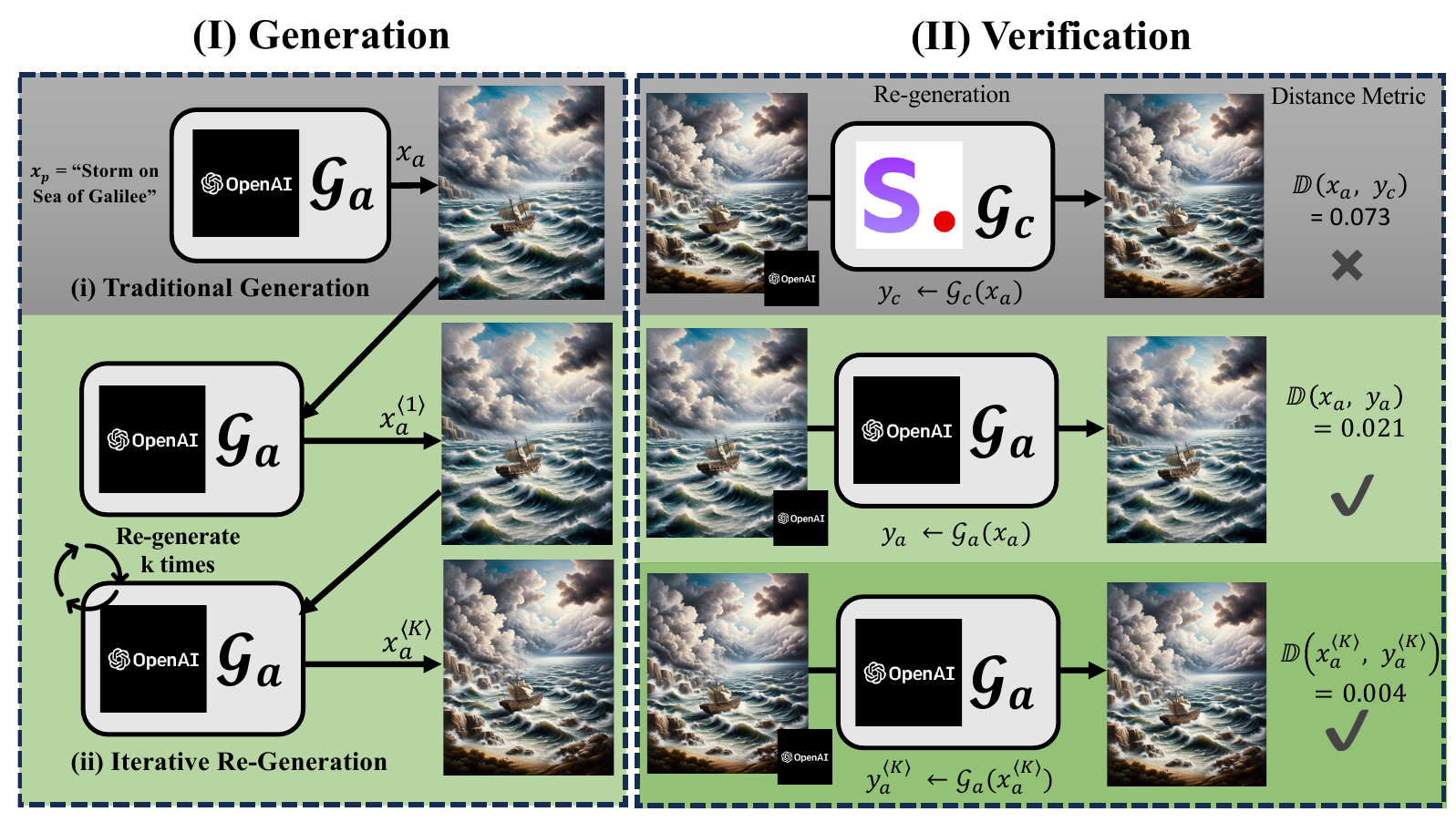}
    \caption{The two-stage framework leveraging fingerprints in generative models. In \textit{(I)} \textit{Generation Stage}, models generate output in traditional ways and optionally re-generate the output $k$ times prior to release. In \textit{(II)} \textit{Verification Stage}, the authentication of data ownership is established by assessing the distance between the data and its re-generated version. This is illustrated using authentic generator ($\mathcal G_a$) and contrasting generator ($\mathcal G_c$), exemplified by models from OpenAI and Stability AI respectively.} 
    \label{fig:teaser}
    \vspace{-4mm}
\end{figure*}


Traditionally, the primary approach for safeguarding IP has involved embedding subtle but verifiable watermarks into the generated content, such as text~\citep{He_Xu_Lyu_Wu_Wang_2022,he2022cater, kirchenbauer2023watermark}, images~\citep{zear2018proposed,zhao2023recipe} and code~\citep{lee2023wrote}. These watermarking techniques typically involve adding supplementary information to the deep learning model's parameters and architectures or direct post-processing alternations to the generated outputs. However, these alternations could potentially degrade the quality of the generated content. An alternative strategy has been the classification of data produced by a specific model to distinguish it from content generated by other models or humans~\citep{DBLP:journals/corr/abs-1908-09203,ippolito-etal-2020-automatic}. Nonetheless, this often requires training additional classifiers to verify authorship, raising concern about their ability to generalize and maintain robustness across evolving generative models, especially with limited training resources. 

In response to the challenges of author authentication and IP protection, our approach is strategically designed to exploit the inherent characteristics of generative models. Firstly, we recognize that generative models possess unique attributes - akin to model fingerprints - such as specific styles and embedded knowledge. In the Verification Stage (I) in our framework, we utilize these implicit fingerprints by measuring the \textit{distance} between the genuine data samples with content re-generated by the authentic and contrasting models. 
Secondly, to reinforce the distinctive nature of these fingerprints, our approach in the Generation Stage (II) involves using the original model to iteratively re-generate outputs from previous iterations. This process is grounded in fixed-point theory~\citep{granas2003fixed} and its practical applications. Through this iterative re-generation, the model's inherent fingerprints become more pronounced, enhancing the effectiveness of our verification process.

In Figure~\ref{fig:teaser}, we present a conceptual framework for authorship verification through re-generation.
\paragraph{Stage I: Generation} The authentic generator targets on producing outputs that involve stealthy but significant signatures that are distinguishable from other generative models or humans. We consider two distinct approaches: \textit{(i)}~`Traditional Generation' produces the authentic outputs $\vx_a$ from a given text input as a prompt, \ie $\vx_a = \mathcal G_a(\vx_p)$, where $\vx_p=\text{``Storm on Sea of Galilee''}$; and \textit{(ii)}~`Iterative Re-Generation' enhances the model's unique signature by re-generating the data multiple times using a `re-painting' mode, \ie $\vx_a^{\langle k+1\rangle}=\mathcal G_a(\vx_a^{\langle k\rangle})$. Here \( \mathcal{G}_{a} \) is the authentic generative model which is the actual creator of output (Dalle from OpenAI in this example)\cite{ramesh2021zero}. 

\paragraph{Stage II: Verification} In this stage, authentic model $\mathcal G_a$ verifies the origin of its artefact $\vx_a$ by comparing the `distance' $\mathbb D$ between $\vx_a$ and its re-generation by the authentic model $\mathcal G_a (\vx_a)$ or other contrasting models $\mathcal G_c (\vx_a)$. Intuitively, the one-step regeneration distance of an image originally generated by the authentic model, such as DALL$\cdot$E by OpenAI, is expected to be smaller when compared to itself than to a contrasting model not involved in its initial generation, \ie $\mathbb D (\vx_a, \mathcal G_a (\vx_a)) < \mathbb D (\vx_a, \mathcal G_c (\vx_a))$. Furthermore, the more re-generations an image undergoes during the Generation Stage, the lower its one-step regeneration of the authentic model becomes at the Verification Stage, \ie $\mathbb D (\vx_a^{\langle i \rangle}, \mathcal G_a (\vx_a^{\langle i \rangle})) < \mathbb D (\vx_a^{\langle j \rangle}, \mathcal G_a (\vx_a^{\langle j \rangle}))$, when $i>j$.

We summarize the key advantages and contributions of our work as follows:

\begin{itemize}

\item We validate the effectiveness of using re-generated data as a key indicator for authorship verification. This approach is designed to be functional in black-box settings and is applicable across varying generative applications, such as Natural Language Generation (NLG) and Image Generation (IG). 

\item We introduce an iterative re-generation technique to enhance the inherent fingerprints of generative models. We use fixed-point theory to demonstrate that modifications achieved through re-generation converge to minimal edit distances. This ensures a distinct separation between the outputs of authentic models and those generated by other models. 

\item We have developed a practical verification protocol that streamlines the process of data ownership validation in generative models. This protocol is particularly valuable in legal contexts as it obviates the need for generators to reveal their model parameters or specific watermarking strategies, thus maintaining confidentiality and proprietary integrity.

\item A notable advantage of our approach is its reliance solely on the standard generative models, without resorting to additional interventions, including (1) manipulating or fine-tuning generative model parameters, (2) post-processing alternations to the outputs, or (3) additional independent classification models for verification. This simplicity in design not only preserves the original quality of the generated content but also enhances the feasibility and accessibility of our verification method.

\end{itemize}

%% file: sec2_related_work.tex
\section{Related Work}
Recent advancements in the field of generative modeling, exemplified by innovations like DALL·E ~\citep{ramesh2022hierarchical}, Stable Diffusion~\citep{Rombach_2022_CVPR}, ChatGPT, Claude, Gemini, and others. This proliferation of synthetic media, however, has concurrently raised ethical concerns. These concerns include the potential for misuse in impersonation~\cite{voicescamnpr, scamwashingtonpost}, dissemination of misinformation~\citep{Pantserev2020, hazell2023large, mozes2023use, scamforbes}, academic dishonesty~\cite{lund2023chatgpt}, and copyright infringement \citep{brundage2018malicious, rostamzadeh2021ethics, He_Xu_Lyu_Wu_Wang_2022,xu2023security}. In response, there is an increasing focus on the need to trace and authenticate the origins of such content to prevent its illegitimate use in Artificial Intelligence-Generated Content (AIGC). Considering the distinct nature of image and text, we will review the authorship identification in image and text generation models separately.

\paragraph{Authorship Identification for Image Generation Models} Image watermarking, recognized as a standard approach for verifying ownership and safeguarding the copyright of a model, involves imprinting a unique watermark onto generated images. Conventional methods encompass direct alterations to pixel values, for instance, in the spatial domain, or the incorporation of watermarks into altered forms of the image, such as in the frequency domain~\citep{cox2008digital}.

With the advancements of deep learning techniques, multiple works have suggested leveraging neural networks to seamlessly encode concealed information within images in a fully trainable manner~\cite{zhu2018hidden,10.1007/978-3-030-11389-6_5, ahmadi2020redmark, you2020siamese}. Inspired by this idea, \cite{9746058} incorporate watermarks into the latent spaces formulated by a self-supervised network like DINO~\citep{caron2021emerging}. This approach modulates the features of the image within a specific region of the latent space, ensuring that subsequent transformations applied to watermarked images preserve the integrity of the embedded information. Subsequently, watermark detection can be conducted within this same latent space. Similarly, \cite{fernandez2023stable} introduces a binary signature directly into the decoder of a diffusion model, resulting in images containing an imperceptibly embedded binary signature. This binary signature can be accurately extracted using a pre-trained watermark extractor during verification.

Given the escalating concerns regarding the misuse of deep fakes, as highlighted in the literature~\cite{brundage2018malicious, Harris2019DeepfakesFP}, several studies have proposed methodologies for attributing the origin of an image, specifically discerning between machine-generated and authentic images. This task is rendered feasible through the identification of subtle, yet machine-detectable, patterns unique to images generated by Generative Adversarial Networks (GANs), as evidenced in recent research~\cite{8695364, afchar2018mesonet, 8639163, yu2019attributing}. Furthermore, detecting deep fakes is enhanced by analyzing inconsistencies in the frequency domain or texture representation between authentic and fabricated images, as indicated in recent studies ~\cite{9035107, durall2020watch,liu2020global}.

\paragraph{Authorship Identification for Natural Language Generation Models} Likewise, content generated by text generation models is increasingly vulnerable to various forms of misuse, including the spread of misinformation and the training of surrogate models~\citep{wallace-etal-2020-imitation, xu2021beyond}. Consequently, a growing interest has been in protecting the authorship (or IP) of text generation models through watermarks. However, unlike images, textual information is composed of discrete tokens, making the watermarking process for text a difficult endeavor due to the potential for inadvertent alternation that can change its semantic meaning~\citep{doi:10.1201/1079/43263.28.6.20001201/30373.5}. One solution for preserving semantic integrity during watermarking involves synonym substitution~\citep{10.1145/1161366.1161397,10.1162/COLI_a_00176, He_Xu_Lyu_Wu_Wang_2022}. Nevertheless, the simplistic approach to synonym substitution is vulnerable to detection through statistical analyses. In response, \cite{he2022cater} have proposed a conditional synonym substitution method to enhance both the stealthiness and robustness of substitution-based watermarks. Moreover, \cite{venugopal-etal-2011-watermarking} adopted bit representation to encode semantically similar sentences, enabling the selection of watermarked sentences through bit manipulation. With the advent of neural language models, traditional rule-based watermarking has evolved towards neural methodologies. For instance, \cite{kirchenbauer2023watermark} successfully watermarks output sentences by prompting the model to yield more watermarked tokens. Rather than producing discrete watermarked tokens, one can modify the output distribution to embed a subtle watermark within a continuous space~\citep{zhao2023protecting}.

The methods previously described are susceptible to rewrite attacks, where the watermarked sentence may be altered through automated or manual paraphrasing, thereby undermining the effectiveness of authorship identification due to watermark distortion. To counteract this, \cite{kuditipudi2023robust} suggests a technique that involves mapping a series of random numbers, generated via a randomized watermark key, onto outputs from a language model. This approach effectively embeds a distinct watermark in the text, capable of being traced to its origin. This ensures the watermark's detectability, even when the text is subjected to modifications like substitutions, insertions, or deletions. Alternatively, robust watermarking can be achieved by fine-tuning a BERT-based infill model~\cite{yoo-etal-2023-robust}. This approach incorporates keyword preservation and syntactically invariant corruptions, enhancing the model's robustness.


%% file: sec_background.tex
\section{Background}

The most recent advanced large generative models support two modes for generating outputs:

\paragraph{Prompt-based Generation:} The authentic generator $\mathcal G$ produces outputs $\vx_a$ conditioned on the given prompts ($\vx_{p}$), expressed as:
\begin{equation}
    \vx_{a} = \mathcal{G}(\vx_{p})
\end{equation}
The selection of prompt inputs $\vx_{p}$ can vary widely, ranging from textual descriptions to images.

\paragraph{Paraphrasing Content:} The generators can also ``paraphrase'' outputs, including texts or images, by reconstructing the content.
\begin{equation}
    \vx_{a}^{\langle\text{new}\rangle} = \mathcal{G}(\vx_{a}^{\langle\text{old}\rangle})
\end{equation}
As a Natural Language Generation (NLG) example, we can rely on round-trip translation to ``paraphrase'' a sentence~\citep{gaspari-2006-look}. As an image generation (IG) example, given a generated image $\vx$, the ``paraphrasing'' process will be (1) rebuilding partial images by randomly masked regions $\vx_a^{\langle\text{new}\rangle}{[t]}=\mathcal G(\vx_a^{\langle\text{old}\rangle}, M{[t]})$ ~\citep{von-platen-etal-2022-diffusers} and (2) merge the new sub-images as a whole new image, $\vx_a^{\langle\text{new}\rangle}=\text{Merge}(\{\vx_a^{\langle\text{new}\rangle}{[t]}\}_{t=1}^T)$. This iterative re-generation uses prior outputs as inputs, 
where $x_a^{\langle 0 \rangle}$is the initial output from the first round of 
prompt-based conditional generation. 

Our method will use \textit{Prompt-based Generation} mode for initial output generation and \textit{Paraphrasing Content} mode for iteratively polishing the generated contents.

%% file: sec3_method.tex
\section{Methodology}
\label{sec:method}

Our research primarily focuses on the threat posed by malicious users who engage in unauthorized usage of generated content. Specifically, we examine the scenario where the owner of authentic generative models, referred to as $\mathcal G_a$, grants access to their models in a black-box fashion, permitting others to query their API for content generation. 
Unfortunately, there exists a risk that these malicious users may exploit the generated content $\vx_a$ without acknowledging the authentic generator's license. Furthermore, they might even falsely attribute the content to unauthorized parties, represented as contrasting generators $\mathcal G_c$. 
To address this issue, API providers can actively monitor the characteristics of publicly available data to identify potential cases of plagiarism. This can be accomplished by applying the re-generation and measuring the corresponding edit distance in the verification stage, as described in Section~\ref{subsec:regeneration}. To enhance verification accuracy, the authentic model $\mathcal G_a$ can employ an iterative re-generation approach to bolster the fingerprinting signal, as introduced and proved converged in Section~\ref{subsec:iter_regeneration}. If there are suspicions of plagiarism, the company can initiate legal proceedings against the alleged plagiarist through a third-party arbitration, following the verification protocol (Algorithm~\ref{alg:verify}) on the output generated by Algorithm~\ref{alg:gen_iter}, as discussed in Section~\ref{subsec:protocol}. The motivation and intuition of the approach proposed in Section~\ref{subsec:protocol} are explained in Section~\ref{subsec:regeneration} and ~\ref{subsec:iter_regeneration}.

\subsection{Data Generation and Verification Protocol} \label{subsec:protocol}

Hereby, the defense framework is comprised of two key components, responding to Generation and Verification Stages in Figure~\ref{fig:teaser}: \textit{i)} Iterative Generation (\Algref{alg:gen_iter}), which progressively enhances the fingerprint signal in the generated outputs; and \textit{ii)} Verification (\Algref{alg:verify}) is responsible for confirming the authorship of the data sample through a one-step re-generation process using both authentic model $\mathcal G_a$ and suspected contrasting model $\mathcal G_c$, with a confidence margin $\delta>0$.



\begin{algorithm}[t]

\caption{Generation Algorithm for \textit{Stage I}.}\label{alg:gen_iter}
\textbf{Input:} Prompt input $\vx_p$ for generation and number of iterations $K$.\\
\textbf{Output:} Image, Text, etc. content with implicit fingerprints.
\begin{algorithmic}[1]

\State $\vx_a^{\langle 0\rangle} \gets \mathcal G(\vx_p)$ \Comment{Initial generation.}
\For{$k \gets 1$ to $K$} \Comment{Iterate $K$ steps.}
    \State $\vx_a^{\langle k\rangle} \gets \mathcal G(\vx_a^{\langle k-1\rangle})$ \Comment{Re-generation.}
\EndFor
\State \Return $\vx_a^{\langle K\rangle}$

\end{algorithmic}
\label{alg:generate_process}
\end{algorithm}
\begin{algorithm}[t]
\caption{Verification Algorithm for \textit{Stage II}. }\label{alg:verify}
\textbf{Input:} Data sample $\vx_a^{\langle K\rangle}$ generated by $\mathcal G_{a}$ and misused by $\mathcal G_{c}$. \\ The threshold $\delta$ for confident authentication.\\

\textbf{Output:} Unauthorized usage according to a contrasting generator.
\begin{algorithmic}[1]
\State $\vy_{a} \gets \mathcal G_{a}(\vx_a^{\langle K\rangle})$ \Comment{Regenerate data by model $\mathcal G_{a}$.}
\State $\vy_{c} \gets \mathcal G_{c}(\vx_a^{\langle K\rangle})$ \Comment{Regenerate data by model $\mathcal G_{c}$.}
\State $r \gets \mathbb D(\vy_{c}, \vx_a^{\langle K\rangle}) / \mathbb D(\vy_a, \vx_a^{\langle K\rangle})$ \Comment{Calc. exceed distance ratio.}
\State \Return $ r > 1+\delta $

\end{algorithmic}
\label{alg:verify_process}

\end{algorithm}

\textbf{Intuition:} Imagine human artists refining their distinct writing or painting styles during each artwork replication. Similarly, the unique `style' of a generative model becomes more defined during image re-generation, as it converges to its representation of what the output should be. This mirrors iterative functions which converge to fixed points (Theorem~\ref{thm:euclid_converge_next}). Each re-generation brings the image closer to the model's fixed point of output, and the distinct fingerprint facilitates authorship verification for AIGC.

\subsection{Authorship Verification through Contrastive Re-Generation} \label{subsec:regeneration}
An authentic generative model $\mathcal G_{a}$ that aims to distinguish between the data samples it generated, denoted as $x_{a}$, with the benign samples $x_c$ generated by other models $\mathcal G_c$ for contrast. To verify the data, the authentic model \textit{i)} re-generates the data  $\mathcal G_a (\vx)$ and \textit{ii)} evaluates the distance between the original sample and the re-generated sample, defined as $d(\vx, \mathcal G) \triangleq \mathbb D (\mathcal G(\vx), \vx)$.  
In essence, during the re-generation process, samples produced by the authentic model are expected to exhibit lower `self-edit' distance, as they share the same generative model $\mathcal G_a$, which uses identical knowledge, such as writing or painting styles, effectively serving as model fingerprints. In mathematical terms, we have
\begin{align}
    \mathbb{D} (\mathcal G_a(\vx_a), \vx_a) & < \mathbb{D} (\mathcal G_a(\vx_{c}), \vx_{c}), \\ \text{ \ \ie \ } d(\vx_a, \mathcal G_a) & < d(\vx_c, \mathcal G_a).
\end{align}
Consequently, authentic models can identify their samples by evaluating this `self-edit' distance, which can be viewed as a specialized form of a classification function for discriminating the authentic and contrasting models. Additionally, the re-generation process and corresponding `edit' distances can serve as explainable and comprehensible evidence to human judges.

\subsection{Enhancing Fingerprint through Iterative Re-Generation} \label{subsec:iter_regeneration}

While we have claimed that the data samples generated through the vanilla generative process can be verified, there is no theoretical guarantee regarding the `self-edit' distance for certifying these samples using the authentic model. To address this limitation, we introduce an iterative re-generation method that improves the fingerprinting capability of the authentic generative model, as it manages to reduce the `self-edit' distances. This property will be utilized in verification, as it serves as the $(K+1)$-th re-generation step for the authentic model.
This section provides a theoretical foundation for examining the characteristics of re-generated samples by exploring the convergence behavior of the fixed points by iterative functions.


\label{sec:fixed_point}
\begin{definition} [The Fixed Points of a Lipschitz Continuous Function] \label{thm:lipschiz_continuous}
    Given a multi-variable function $f: \R^m \rightarrow \R^m $ is $L$-Lipschitz continuous, \ie $\|f(\vx) - f(\vy)\| \leq L \cdot \|\vx-\vy\|$, where $L \in (0,1)$. 
    For any initial point $\vx^{\langle 0\rangle}$, the sequence $\{\vx^{\langle k\rangle}\}_{k=0}^{\infty}$ is acquired by the recursion $\vx^{\langle k+1\rangle}=f(\vx^{\langle k\rangle})$. The sequence converges to a fixed point $\vx^*$, where $\vx^*=f(\vx^*)$.
\end{definition}

\begin{theorem} [The Convergence of Step Distance for $k$-th Re-generation] \label{thm:euclid_converge_next}
The distance between the input and output of the $k$-th iteration is bounded by 
\begin{equation}
    \|\vx^{\langle K+1\rangle} - \vx^{\langle K\rangle}\| \leq L^K \cdot \|\vx^{\langle 1 \rangle} - \vx^{\langle 0\rangle}\|,
\end{equation}
and the distance converges to 0 given $L \in (0,1)$.\footnote{We estimated $L$ for Stable Diffusion models with results presented in Appendix ~\ref{app:lipschitz_empirical_est}.}
\end{theorem}
\begin{proof} We apply $L$-Lipschitz continuous property recursively,
\begin{align}
    \nonumber &\|\vx^{\langle K+1\rangle} - \vx^{\langle K\rangle}\| = \|f(\vx^{\langle K \rangle}) - f(\vx^{\langle K-1 \rangle})\| \\
    \nonumber & \leq L \cdot \|\vx^{\langle K\rangle} - \vx^{\langle K-1 \rangle}\| = L \cdot \|f(\vx^{\langle K-1 \rangle}) - f(\vx^{\langle K-2 \rangle})\| \\
     & \leq L^2 \cdot \|\vx^{\langle K-1 \rangle} - \vx^{\langle K-2 \rangle}\| \leq \cdots \leq L^k \cdot \|\vx^{\langle 1\rangle} - \vx^{\langle 0 \rangle}\| . 
\end{align}
\end{proof}
The theory can be extended to distance metrics in Banach space.
\begin{theorem} [Banach Fixed-Point Theorem]~\citep{banach1922operations, ciesielski2007stefan} Let $(\mathcal X,\mathbb D)$ be a complete metric space, and $f: \mathcal X \rightarrow \mathcal X$ is a contraction mapping, if there exists $L \in (0,1)$ such that for all $\vx, \vy \in \mathcal X$ 
\begin{equation}
\label{eq:lipschitz}
    \mathbb D(f(\vx),f(\vy))\leq L \cdot \mathbb D(\vx,\vy).
\end{equation}
Then, we have the following convergence of the distance sequence similar to Theorem~\ref{thm:euclid_converge_next},
\begin{equation}
    \mathbb D(\vx^{\langle K+1\rangle } - \vx^{\langle K \rangle}) \leq L^K \cdot \mathbb D(\vx^{\langle 1\rangle} - \vx^{\langle 0\rangle}).
\end{equation}
\end{theorem}

\begin{figure*}[!htb]
    \centering
    \includegraphics [width=0.9\linewidth] {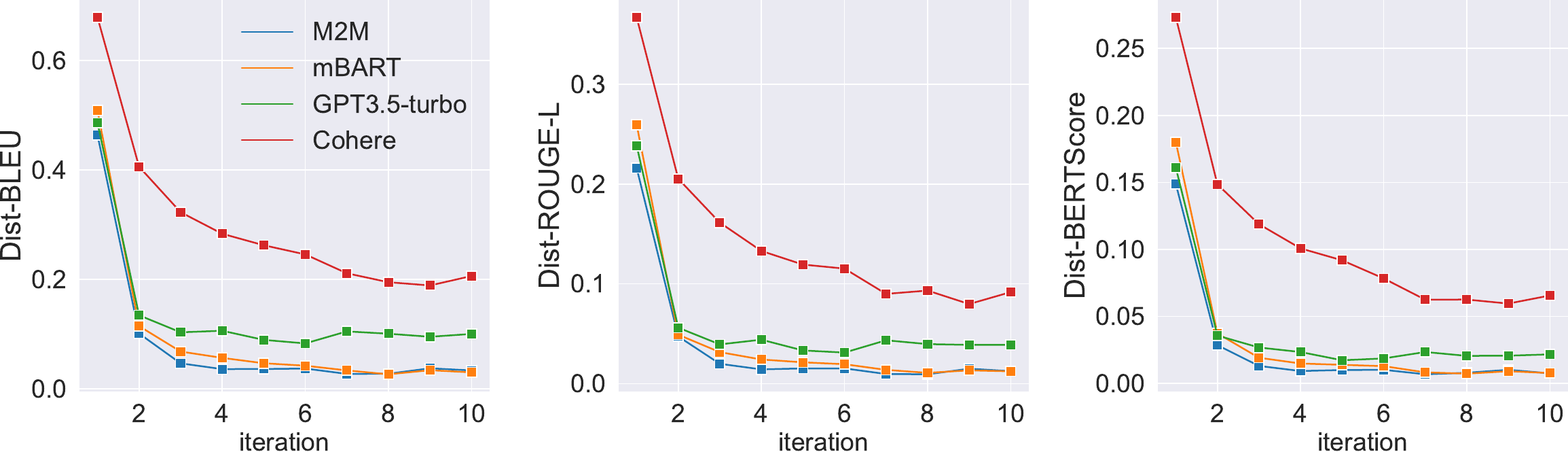}
    \caption{The convergence analysis of the distances in iterations based on various metrics on the re-generated text of 200 samples from in-house datasets. } 
    \label{fig:self_iter_nlp}
    \vspace{-2mm}
\end{figure*}

%% file: sec4_experiment.tex
\section{Experiments} \label{sec:experiments}


This section aims to demonstrate the efficacy of our re-generation framework of authorship authentication on generated text and image separately. For both Natural Language Generation (NLG) and Image Generation (IG) scenarios, Sections~\ref{subsec:nlg_exp} and~\ref{subsec:ig_exp}, we first generate the initial intended data samples followed by several steps of `paraphrasing' (for NLG) or `inpainting' (for IG) as re-generation, detailed in \Algref{alg:gen_iter}. Then, we test the properties of these samples by three series of experiments. 

\begin{enumerate}
    \item \textbf{Distance Convergence}: We verify the convergence of the distance between one-step re-generation. 
    \item \textbf{Discrepancy}: We illustrate the discrepancy between the distances by the authentic generative models and the `suspected' models for contrast. 
    \item \textbf{Verification}: We report the verification performance using precision and recall of the identified samples from the authentic model.
\end{enumerate}


\subsection{Experimental Setup}
Despite falling under the same framework for authorship verification, Natural Language Generation (NLG) and Image Generation (IG) use distinct generative models and varying metrics for measuring distance.

\paragraph{Generative Models.}
\label{sec:gen-models}
For text generation, we consider four generative models: 1) \textbf{M2M} ~\citep{fan2021beyond}: a multilingual encoder-decoder model trained for many-to-many multilingual translation; 2) \textbf{mBART-large-50}~\citep{tang-etal-2021-multilingual}: a model fine-tuned on mBART for multilingual machine translation between any pair of 50 languages; \textbf{Cohere}: a large language model developed by Cohere; 4) \textbf{GPT3.5-turbo} (version: 0613): a chat-based GPT3.5 model developed by OpenAI.\footnote{We have also studied GPT4 and present its corresponding results in the Appendix~\ref{app:gpt4}.}  


For image generation, we examine five primary generative models based on the Stable Diffusion (SD) architecture~\citep{Rombach_2022_CVPR}. All models are trained on a subset of the LAION-2B dataset \citep{schuhmann2022laion} consisting of CLIP-filtered image-text pairs. These models are:
1) \textbf{SDv2.1}; 
2) \textbf{SDv2}; 
3) \textbf{SDv2.1 Base}; 
4)\textbf{SDXLv1.0}~\citep{podell2023sdxl}, which distinguishes itself by employing an ensemble of experts pipeline for latent diffusion, where the base model first generates noisy latent that are refined using a specialized denoising model
5) \textbf{SDXL Base0.9}~\citep{podell2023sdxl}.\footnote{Additionally, we explored other variants, including \textbf{SDv1.5}, \textbf{SDv1.4}, and \textbf{Nota-AI's BK-SDM Base}.  The corresponding experiments and results are provided in Appendix \ref{app:cv_models}.} More information on model architecture and training, and the quality of re-generated images is demonstrated in Appendix \ref{app:cv_models}.

\paragraph{Distance Metrics.} To gauge the similarity between inputs and outputs, we employ three popular similarity metrics for NLG experiments, 1) \textbf{BLEU}: calculate the precision of the overlap of various n-grams (usually 1 to 4) between the candidate and the reference sentences~\citep{papineni-etal-2002-bleu}; 2) \textbf{ROUGE-L}: calculate the F1 of longest common subsequence between the candidate and reference sentences~\citep{lin-2004-rouge} and 3) \textbf{BERTScore}: compute the similarity between candidate and reference tokens using cosine similarity on their embeddings. Then, the token-level scores are aggregated to produce a single score for the whole text~\citep{Zhang2020BERTScore}. 
For IG experiments, we consider the following distance metrics: 1) \textbf{CLIP Cosine Distance} measures the semantic similarity using pre-trained CLIP image-text embeddings, where a lower distance indicates more closely aligned high-level content~\citep{radford2021learning}; 2) \textbf{LPIPS} compares perceptual style differences~\citep{zhang2018unreasonable}; 3) \textbf{Mean Squared Error (MSE)} serves as a pixel-level metric that compares raw image values; 4) \textbf{Structural Similarity Index (SSIM)} assesses image degradation based on luminance, contrast, and structure~\citep{wang2004image}. 
 We transform all similarity scores $s \in [0,1]$ to distances $d$ by $d=1-s$.
 
Note that we gauge the success of our approach based on the satisfactory performance of any of these metrics when applied to the verification of generative models.

\input{sec4_exp_NLP}

\input{sec4_exp_CV}

\input{sec4_exp_robustness}

%% file: sec4_exp_NLP.tex
\begin{figure*}[!htb]
     \centering
     \begin{subfigure}
         \centering
         \includegraphics[width=0.9\textwidth]{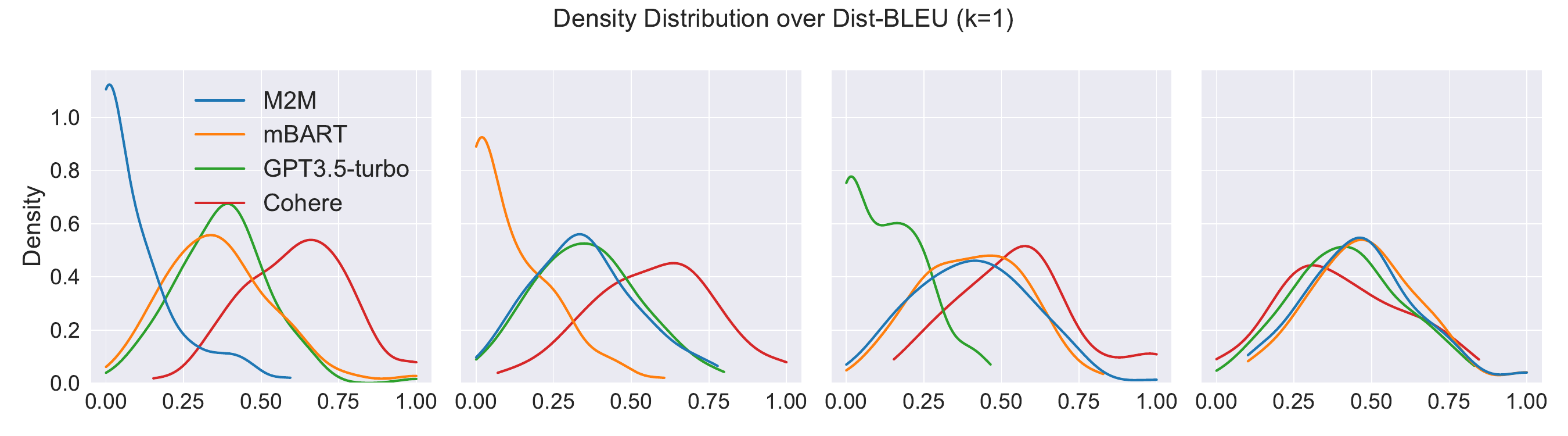}
     \end{subfigure}
     \vspace{0.02cm}
     \begin{subfigure}
         \centering
         \includegraphics[width=0.9\textwidth]{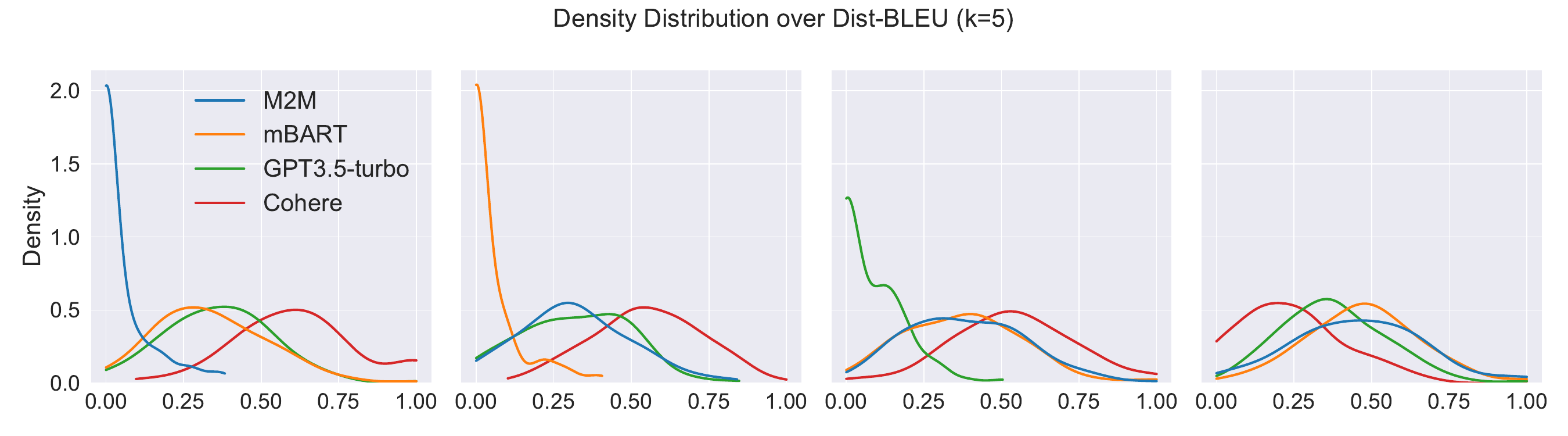}
     \end{subfigure}
        \caption{Density distribution of one-step re-generation among four text generation models, where the input to the one-step re-generation is the $k$-th iteration from the authentic models. The authentic models from left to right are: 1) M2M, 2) mBART, 3) GPT3.5-turbo, and 4) Cohere.}
        \label{fig:bleu_one_step}
\end{figure*}

\subsection{Natural Language Generation Experiments} \label{subsec:nlg_exp}

In the study of text generation, the primary objective is to produce sentences that fulfill specific requirements, including translation, paraphrasing, summarization, \etc. Then, one can re-generate intended output sentences via `paraphrasing' them. This study focuses on machine translation (specifically from French to English) as its primary generation task.\footnote{In addition to machine translation, we examine paraphrasing and summarization as the generation tasks in \Secref{sec:robustness}} Considering that \mone and \mtwo are restricted to machine translation tasks, we utilize round-trip translation to paraphrase English sentences. Specifically, for each round-trip translation step, an English input sentence is first translated into French and then back to English. 
We repeat this process multiple times to observe the change of the `edit' distances.
For \mthree and \mfour, we perform all experiments in a zero-shot prompting. We use the following prompts:
\begin{itemize}
    \item Translate to English: \textit{You are a professional translator. You should translate the following sentence to English and output the final result only: \{\textbf{INPUT}\}}
    \item Translate to French: \textit{You are a professional translator. You should translate the following sentence to French and output the final result only: \{\textbf{INPUT}\}}
\end{itemize}

\input{tables/tab_nlp_verification_diff_rate}

\input{tables/tab_nlp_verification}

It has been known that text generative models exhibit superior performance when test data inadvertently overlaps with pre-training data, a phenomenon referred to as data contamination~\citep{magar2022data}. To address this potential issue, we sample 200 sentences from the in-house data as the starting point for each model. To mitigate biases stemming from varied sampling strategies, we set the temperature and top p to 0.7 and 0.95 for all models and experiments. 

\paragraph{Distance Convergence.} The dynamics of the distance change across three metrics and various generative models are depicted in \Figref{fig:self_iter_nlp}. We observe a remarkable distance reduction between the first and second iterations in all settings. Subsequently, the changes in distance between consecutive iterations exhibit a diminishing pattern, tending to converge after approximately 5-7 rounds of re-generation. These observed trends are consistent with the fixed-point theorem as elaborated in \Secref{sec:fixed_point}.

\paragraph{Discrepancy.} In our study of iterative re-generation using the same model, we observe convergence which can be utilized to distinguish the authentic model from its counterparts. As delineated in \Secref{subsec:regeneration}, for each sentence $\vx$ from our corpus $X$, we apply the authentic model to yield $\vx_a$ via a translation. Both authentic and contrasting models then perform a one-step re-generation of $\vx_a$ to obtain $y$. Finally, we can measure the distance between $\vx_a$ and $\vy$ (including $\vy_a$ and $\vy_c$) as described in \Secref{subsec:regeneration}. According to \Figref{fig:bleu_one_step}, the density distribution associated with the authentic model demonstrates a noticeable divergence from those of models for contrast, thus affirming the hypothesis discussed in \Secref{subsec:regeneration}. This consistent pattern holds when using ROUGE-L and BERTScore for distance measures, as evidenced by their respective density distributions in Appendix~\ref{app:density}.

\begin{figure*}[ht]
    \centering
    \includegraphics[width=0.9\linewidth]{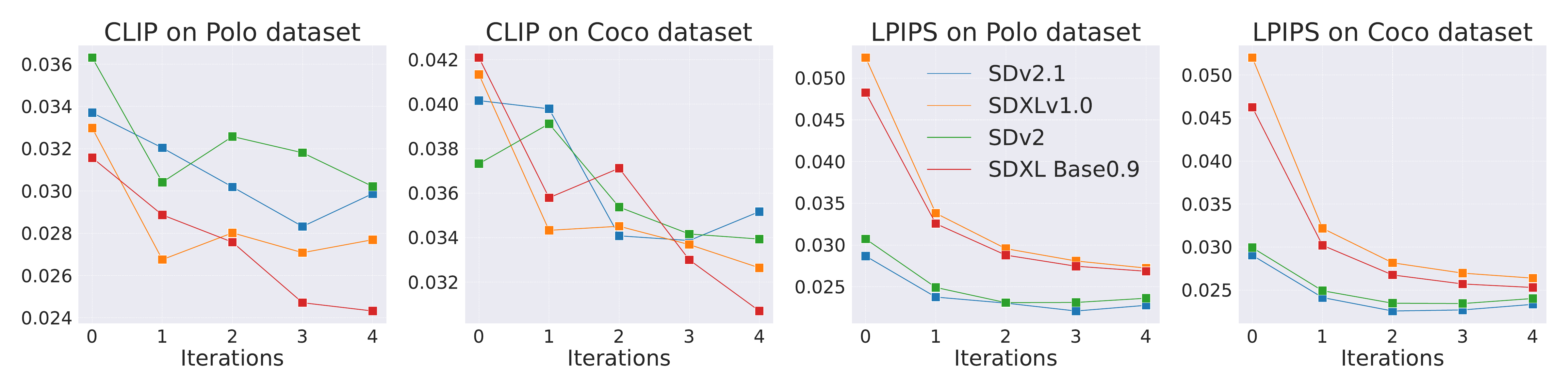}
    \caption{The convergence analysis of the distances in iterations based on CLIPS and LPIPS metrics on re-generated images of 200 samples on Coco and Polo datasets.}
    \label{fig:convergence-analysis-regen}
\end{figure*}

We observe the difficulty of distinguishing Cohere from other models merely based on the outputs from the first iteration, \ie without re-generation. We attribute the distinctive behavior of Cohere to its relatively slower convergence than other models.\footnote{The underlying reason behind the slow convergence for Cohere is investigated in Appendix~\ref{app:text_regen}.} However, as depicted in \Figref{fig:self_iter_nlp}, the distances observed in subsequent iterations of the same model experience a substantial reduction. This implies the re-generated outputs in the Generation Stage possess enhanced fingerprints. In particular, we can derive $x_a^{\langle k \rangle}$ from $5$ iterations of $\mathcal{G}_a(\cdot)$. \Figref{fig:bleu_one_step} shows that increasing $k$ enhances the differentiation of the authentic model's density distribution compared to other models. This distinction becomes evident even in the case of Cohere, which remains indistinguishable at $k=1$. This enhancement for Cohere is also corroborated in the verification part.

\paragraph{Verification.} We employ \Algref{alg:verify_process} to ascertain if a given sentence originates from the authentic models. The determination hinges on the threshold parameter, $\delta$. Thus, we designate \mone as the authentic model, while using \mtwo, \mfour, and \mthree as contrast models to determine the optimal value of $\delta$. As indicated in \Tabref{tab:m2m_verification}, a threshold of 0.05 certifies that the three contrast models can validate that more than 93\% of the samples are derived from \mone. As anticipated, an increase in the value of $\delta$ augments the stringency criteria, resulting in a reduction in verification precision. Therefore, we fix $\delta$ at 0.05 for ensuing evaluations unless stated otherwise.


As indicated in \Tabref{tab:ACC_all_nlp_models}, our approach undertakes the validation of authorship, yielding a precision exceeding 85\% and a recall rate surpassing 88\% across \mone, \mtwo, and \mthree. Moreover, both precision and recall metrics improve with additional re-generation iterations. 

As expected, Cohere's slower convergence results in comparatively less effective performance. Nevertheless, enhancements in both precision and recall are observed when $k=5$.


%% file: tables/tab_nlp_verification_diff_rate.tex
\begin{table*}[!htb]
    \centering
      \caption{The precision of differentiating contrast models from the authentic model ($\mathcal G_a=$M2M) using various $\delta\in\{0.05, 0.1, 0.2, 0.4\}$, BLUE, ROUGE-L, and BERTScore, respectively.}
    \scalebox{1}{
    \begin{tabular}{c|ccc|ccc|ccc}
    \toprule
    \multirow{2}{*}{\textbf{$\delta$}} & \multicolumn{3}{c|}{\textbf{mBART}} &  \multicolumn{3}{c|}{\textbf{Cohere}} & \multicolumn{3}{c}{\textbf{GPT3.5-turbo}}\\
    & BLEU & ROUGE & BERT & BLEU & ROUGE & BERT& BLEU & ROUGE & BERT\\
    \midrule
          0.05 & 94.0 & 92.0 & 91.0  & 99.0 & 99.0 & 99.0 &93.0 & 89.0  & 92.0\\
          0.10   & 91.0 & 92.0& 90.0 & 98.0 & 99.0 & 99.0 & 92.0 & 88.0 & 92.0\\
          0.20  & 91.0 & 92.0& 90.0& 98.0 & 99.0 & 99.0&  91.0 & 87.0 & 90.0  \\
          0.40 & 88.0 & 88.0& 85.0 & 95.0 & 97.0 & 99.0& 87.0 & 85.0 & 85.0\\
    \bottomrule
    \end{tabular}
    }
    \label{tab:m2m_verification}
\end{table*}

%% file: tables/tab_nlp_verification.tex
\begin{table*}
    \centering
    \caption{The precision and recall of verifying the authentic models ($\mathcal G_{a}$) using different contrasting models ($\mathcal G_c$).}
    \scalebox{1}{
    \begin{tabular}{c cccccccc}
    \toprule
    \multirow{2}{*}{\backslashbox{$\mathcal G_{a}$}{$\mathcal G_c$}}& \multicolumn{2}{c}{\textbf{M2M}} & \multicolumn{2}{c}{\textbf{mBART}} &  \multicolumn{2}{c}{\textbf{Cohere}} & \multicolumn{2}{c}{\textbf{GPT3.5-turbo}}\\
    & Precision $\uparrow$ & Recall $\uparrow$ &Precision $\uparrow$ & Recall $\uparrow$ & Precision $\uparrow$ & Recall $\uparrow$ &  Precision $\uparrow$ & Recall $\uparrow$\\
    \bottomrule
    \\[-0.8em]
    \multicolumn{9}{c}{(a) $k=1$}
    \\[0.1em]
    \toprule
          \textbf{M2M}  & - &-& 94.0 & 98.0 &\ \ 99.0 & \ \ 99.0 & 93.0 & 95.0 \\
          \textbf{mBART} & 85.0 & 89.0&- & - & 100.0 & 100.0 & 89.0  & 92.0 \\
          \textbf{Cohere} & 60.0 & 64.0 & 63.0  & 57.0& - & - & 65.0 & 51.0\\
          \textbf{GPT3.5-turbo} & 87.0 & 92.0& 90.0 & 92.0 & \ \ 99.0 & \ \ 99.0 & - & - \\
    \bottomrule
    \\[-0.8em]
    \multicolumn{9}{c}{(b) $k=3$}
    \\[0.1em]
    \toprule
   \textbf{M2M}& - & -& 94.0 & 99.0  & 100.0 & 100.0 & 95.0 & 97.0\\
\textbf{mBART}&85.0 & 93.0 & - &-  & 100.0 & 100.0 & 90.0 & 96.0 \\
 \textbf{Cohere} & 75.0 & 83.0 & 73.0 & 76.0  & - & - & 63.0 & 69.0\\
 \textbf{GPT3.5-turbo} & 89.0 & 92.0 & 93.0 & 95.0  & \ \ 97.0& \ \ 98.0 & - & -\\
 \bottomrule
    \\[-0.8em]
    \multicolumn{9}{c}{(c) $k=5$}
    \\[0.1em]
    \toprule
\textbf{M2M} & - & - & 94.0 & 98.0  & \ \ 98.0 & \ \ 99.0 & 95.0 & 99.0 \\
\textbf{mBART} & 91.0 & 96.0 & - & -  & 100.0 & 100.0& 88.0 & 94.0 \\
 \textbf{Cohere} & 71.0& 81.0 & 83.0 & 87.0  & - & - & 69.0 & 73.0\\
 \textbf{GPT3.5-turbo} & 90.0 & 94.0 & 94.0 & 98.0 & \ \ 95.0 & \ \ 97.0& - & - \\
    \bottomrule
    \end{tabular}
    }
    \label{tab:ACC_all_nlp_models}
\end{table*}


%% file: sec4_exp_CV.tex
\begin{figure*}[htb]
    \centering
    \includegraphics[width=0.9\linewidth]{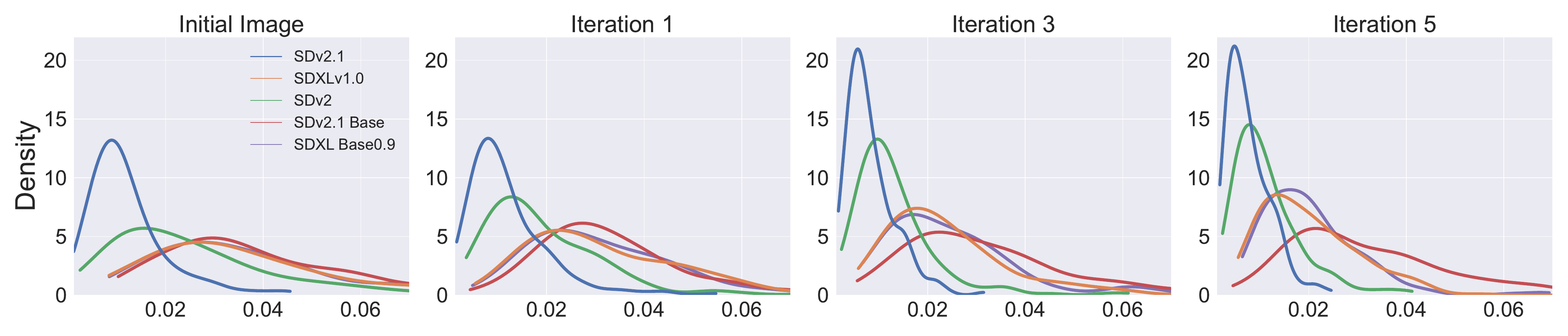}
    \includegraphics[width=0.9\linewidth]{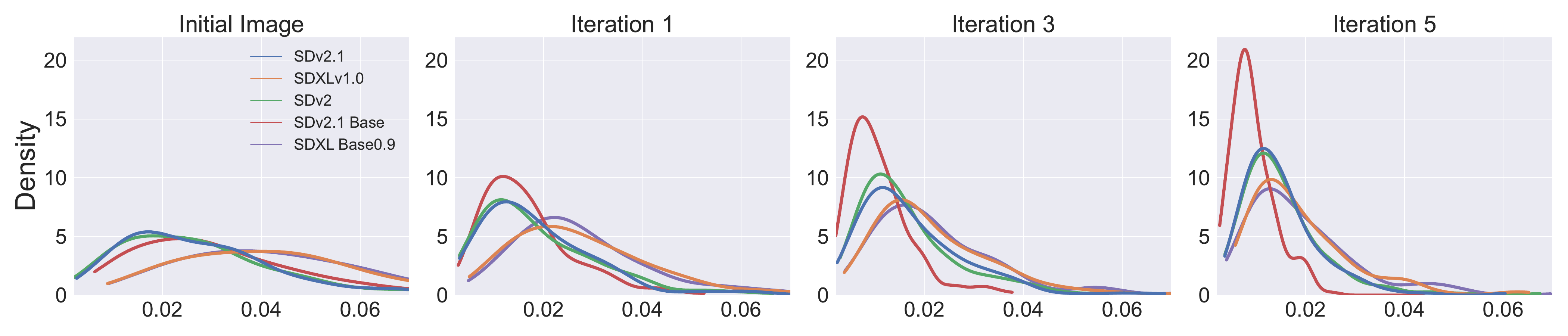}
    \caption{Verifying data generated by authentic $\mathcal G_a$ SDv2.1 and SDv 2.1 Base on Coco Dataset at various iterations using CLIP distance.}
    \label{fig:enhanced_disc_m2}
    \vspace{-3mm}
\end{figure*}

\subsection{Image Generation Experiments} \label{subsec:ig_exp}
The main objective of the image generation task is to produce an image given a text prompt using a generative model. To generate the initial proposal of images, we sample 200 prompts each from the \textbf{MS-COCO dataset (COCO)}~\citep{lin2014microsoft} and \textbf{Polo Club Diffusion DB dataset (POLO)}~\citep{wang2022diffusiondb}, then we generate initial images corresponding to the prompts using all assigned models. We consider two settings in re-generating: {(1)} \textit{watermarking} images through inpainting and {(2)} \textit{fingerprinting} enhancement through full image re-generation. The subsequent paragraphs detail the methodologies for each.

\paragraph{Watermarking through Inpainting}

When re-generating authentic image $\vx$, we mask $1/N$-th ($N=10$) of its pixels with fixed positions as the watermark pattern. All pixels in the masked regions are then re-generated by inpainting using a generative model. 
Similar to text generation, the inpainting step is iterated multiple times, following Algorithm \ref{alg:gen_iter}. The comprehensive description of the watermarking procedure is provided in Appendix~\ref{app:watermark-detailed}. 

\begin{figure*}[t]
    \centering
    \includegraphics[width=0.95\linewidth]{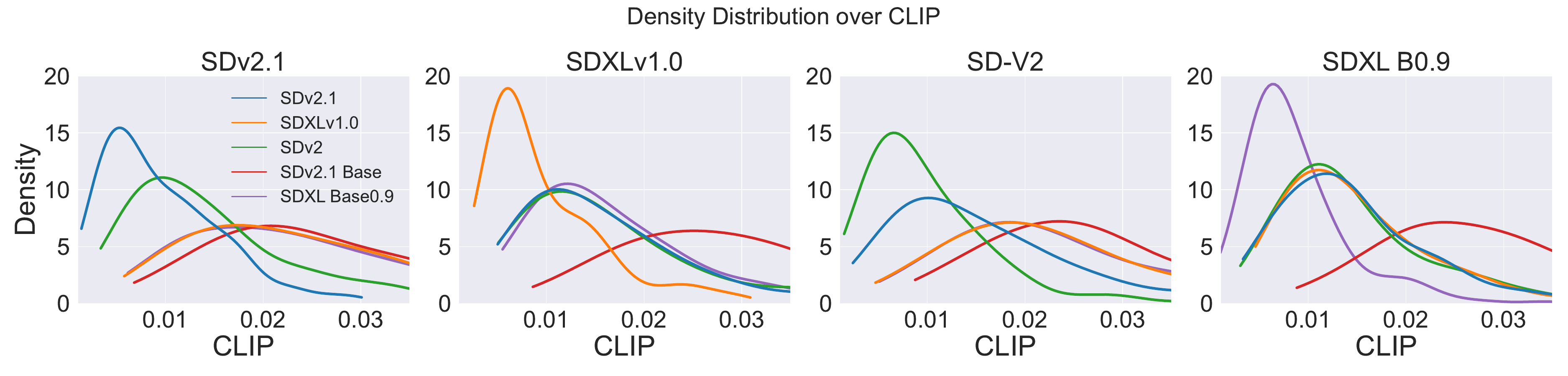}
    \caption{Density distribution of one-step re-generation among four image generation models on Polo Dataset. The authentic models from left to right are: 1) SD 2.1, 2) SDXL 1.0, 3) SD 2, 4) SDXL Base 0.9.}
    \label{fig:clip-one-step}
\end{figure*}

\input{tables/tab_cv_verification_main}

\begin{figure*}[!htb]
    \centering
    \includegraphics[width=0.9\linewidth]{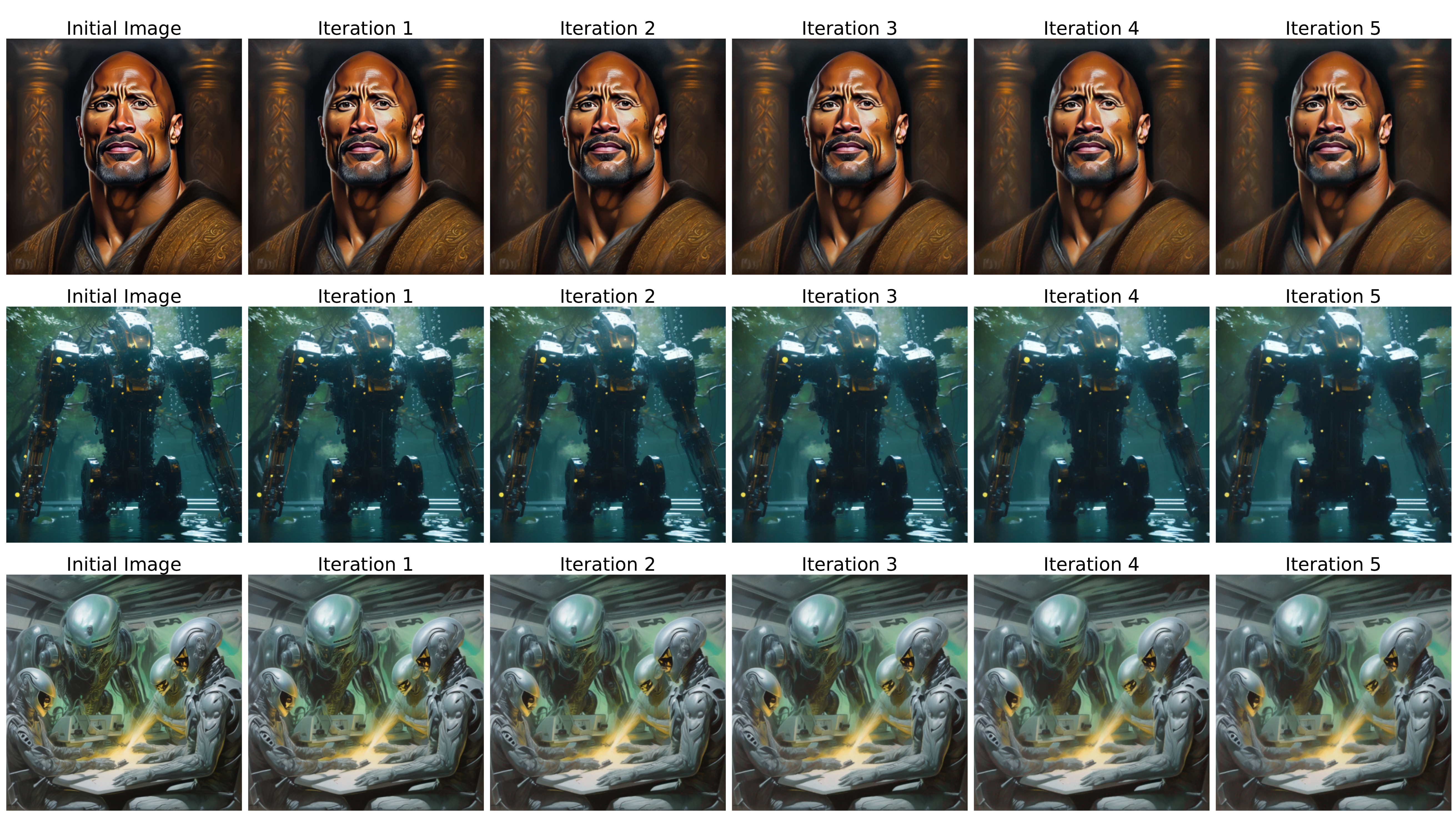}
    \caption{The quality of fingerprinted images over multiple iterations of re-generation using SDXL 0.9B.}
    \label{fig:sdxl-0.9b-main}
\end{figure*}


\paragraph{Fingerprinting through Re-generation}  In the re-generation setting, we first split all possible mask positions into 8 non-overlapped sets, coined \textit{segments}, each with $1/8$ pixel positions for a given image $\vx_a$. We parallelly reconstruct each segment based on the rest $7/8$ pixels of the image as context. Then, we reassemble the 8 generated segments into a new full image $\vy$. When analyzing reconstructions of $\vy_a=\mathcal{G}_a(\vx_a) $ by contrasting models $\vy_c=\mathcal{G}_c(\vx_a)$, there is an expectation that models may exhibit variations in painting masked regions due to their inherent biases and training. In the verification stage, by comparing $\vy_a$ and $\vy_c$ to the original $\vx_a$, we aim to identify subtle model fingerprints based on the `edit-distance' from the original data to re-generated content.
For this study, each model generates 200 images, which are then re-generated over one to five iterations. We compute LPIPS and CLIP similarity between $\vx_a$ and $\vy$ to quantify the model behavior.\footnote{The performance of MSE and SSIM is reported in Appendix~\ref{app:convergence-regeneration}} The detailed discussion on the quality of re-generated images is provided in Appendix~\ref{app:image_regen_quality}.

\textbf{Distance Convergence.} 
Similar to text re-generation, the re-generation distance converges for both watermarking and fingerprinting settings, as illustrated in ~\Figref{fig:convergence-analysis-regen} and ~\ref{fig:watermark-coco}, indicated by consistent downward trends across datasets and distance metrics. This implies that the re-generation process converges, with subsequent iterations yielding images that more closely resemble their predecessors. While our methodology proves efficacious in both scenarios, subsequent sections will prioritize the fingerprinting setting. Details about the watermarking setting can be found in Appendix \ref{app:convergence-watermark}.


\textbf{Discrepancy.}
In our study, we focus on the setting of iterative re-generation through inpainting and observe convergence that can reveal differences between models. Like the NLG section both authentic and contrast models perform a one-step re-generation of $\vx_a$ to obtain $\vy$ and the CLIP distance is measured between the outputs. 
As illustrated in \Figref{fig:clip-one-step}, the one-step re-generation density distribution of the authentic model predominantly peaks at lower values across $\mathcal{G}$. This indicates that authentic models can effectively distinguish their own re-generated outputs from those of non-authentic models. The divergence in CLIP distances after multiple rounds of inpainting highlights our method's ability to distinguish between generative models. Using images from later inpainting iterations (\eg $k>1$) further emphasizes each model's unique characteristics. A similar trend is evident with the Coco dataset and LPIPS distance metric, the details of which can be found in Appendix \ref{app:density_cv}.

The effectiveness of this method is illustrated in \Figref{fig:enhanced_disc_m2} where we can observe even generative models that are not adept at inpainting like SDv2.1 Base over iterations can successfully enhance their fingerprints by 5 iterations. While most models, on average, may require fewer rounds of re-generation to deeply impart their fingerprints, with more rounds of re-generation, even older models can successfully enhance their fingerprints. The quality of the images is illustrated in \Figref{fig:sdxl-0.9b-main}, showcasing a high retention of visual quality through five iterations of re-generation, with the subject's features and the surrounding details remaining sharp and consistent. Simultaneously, the strength of the fingerprint of the generating model is reinforced over iterations, enhancing the model's traceability without degrading the image integrity, as indicated by the trend in \Figref{fig:enhanced_disc_m2}.

\textbf{Verification.} In accordance with the text generation framework, we utilize \Algref{alg:verify_process} to evaluate verification performance, selecting optimal parameters $k=5$ and $\delta=0.05$ based on the results of the text generation experiment. As demonstrated in \Tabref{tab:m2m_verification_clip}, models reliably identify images from the authentic model, with precision often exceeding 85\% and recall around 85\%.

%% file: tables/tab_cv_verification_main.tex
\begin{table*}[htb]
    \centering
    \caption{The precision and recall of verifying the authentic models \(\mathcal G_{a}\) using different contrasting models \(\mathcal G_c\) on Polo Dataset, given $k=5$ and $\delta=0.05$.}
    \scalebox{0.85}{
    \begin{tabular}{c | cccccc | cccccccc}
    \toprule
    \multirow{2}{*}{\backslashbox{\(\mathcal G_{a}\)}{\(\mathcal G_c\)}} & \multicolumn{2}{c}{\textbf{SD v2.1}} & \multicolumn{2}{c}{\textbf{SD v2}} & \multicolumn{2}{c | }{\textbf{SD v2.1B}} & \multicolumn{2}{c}{\textbf{SDXL 0.9}} & \multicolumn{2}{c}{\textbf{SDXL 1.0}} \\
    & Precision \(\uparrow\) & Recall $\uparrow$ & Precision \(\uparrow\) & Recall $\uparrow$ & Precision \(\uparrow\) & Recall $\uparrow$ & Precision \(\uparrow\) & Recall $\uparrow$ & Precision \(\uparrow\) & Recall $\uparrow$ \\
    \midrule
    \textbf{SD v2.1} & - & - & 80.0 & 86.0 & \ \ 99.5 & \ \ 99.5 & 98.5 & \ \ 99.5 & 99.0 & 100.0 \\
    \textbf{SD v2} & 77.5 & 81.5 & - & - & \ \ 99.0 & 100.0 & 96.5 & \ \ 97.0 & 95.5 & \ \ 97.5 \\
    \textbf{SD v2.1B} & 81.5 & 83.5 & 82.5 & 16.5 & - & - & 88.5 & \ \ 92.0 & 89.5 & \ \ 91.0 \\
    \midrule
    \textbf{SDXL 0.9} & 97.5 &  97.5 & 96.5 & 98.0 & \ \ 99.5 & \ \ 99.5 & - & - & 92.0 &  \ \ 93.5 \\
    \textbf{SDXL 1.0} & 95.5 & 96.0 & 92.5 & 95.5 & 100.0 & 100.0 & 94.5 & \ \ 96.0 & - & - \\
    \bottomrule
    \end{tabular}
    }
    \label{tab:m2m_verification_clip}
\end{table*}

%% file: sec4_exp_robustness.tex
\subsection{Generalization and Robustness Evaluation}
\label{sec:robustness}
\paragraph{Task Generalization} To demonstrate the generalization of our approach, we further examine two popular text generation tasks using prompting: paraphrasing and summarization. The corresponding prompts are:
\begin{itemize}
    \item \textbf{Paraphrasing}: \textit{You are a professional language facilitator. You should paraphrase the following sentence and output the final result only:};
    \item \textbf{Summarization}: \textit{You are a professional language facilitator. You should summarize the following document using one sentence:}
\end{itemize}
We use prompt-based paraphrasing for one-step re-generation, opting for a $k$ value of 5 due to its superior performance. As for backbone models for text generation tasks, we consider \mthree, Llama2-chat-7B~\cite{touvron2023llama}, and Mistral~\cite{jiang2023mistral}.

\Tabref{tab:add_nlp_tasks} demonstrates that while paraphrasing and summarization tasks exhibit performance degradation compared to machine translation, our approach effectively distinguishes IPs among various models. This supports the overall effectiveness of our approach across a wide range of text-generation tasks.

Until now, we assumed that malicious users employed the authentic model to generate text and images, subsequently utilizing and disseminating them directly. Nevertheless, in order to circumvent potential IP infringements, malicious users may choose to modify the generated content prior to its public dissemination. This situation poses a challenge to the effectiveness of our approach in protecting the IP claims associated with the altered content.

\paragraph{Robustness Evaluation} To assess the robustness of our verification process, we simulate the modification process using paraphrasing and perturbation attacks.

\input{tables/tab_paraphrasing}

Our robustness assessment commenced with a paraphrasing attack targeting image generation. This experiment was designed to test the efficacy of our verification process in identifying plagiarized images. Specifically, we use a set of images designated as $A\_B$, originally generated by one model (Model A) and then re-generated or paraphrased by another model (Model B). We experiment with SD v2.1 as the authentic model and SDXL 1.0 as the paraphrasing model.

We re-generate these paraphrased $A\_B$ images using both the authentic and paraphrasing models to analyze the distance density distribution variations. Furthermore, we assess this paraphrasing attack at different rounds of generation $\vx_a^{\langle K\rangle}$ where K is 1, 3, and 5 by the authentic model. 

\begin{figure}[ht]
    \centering
    \includegraphics[width=\linewidth]{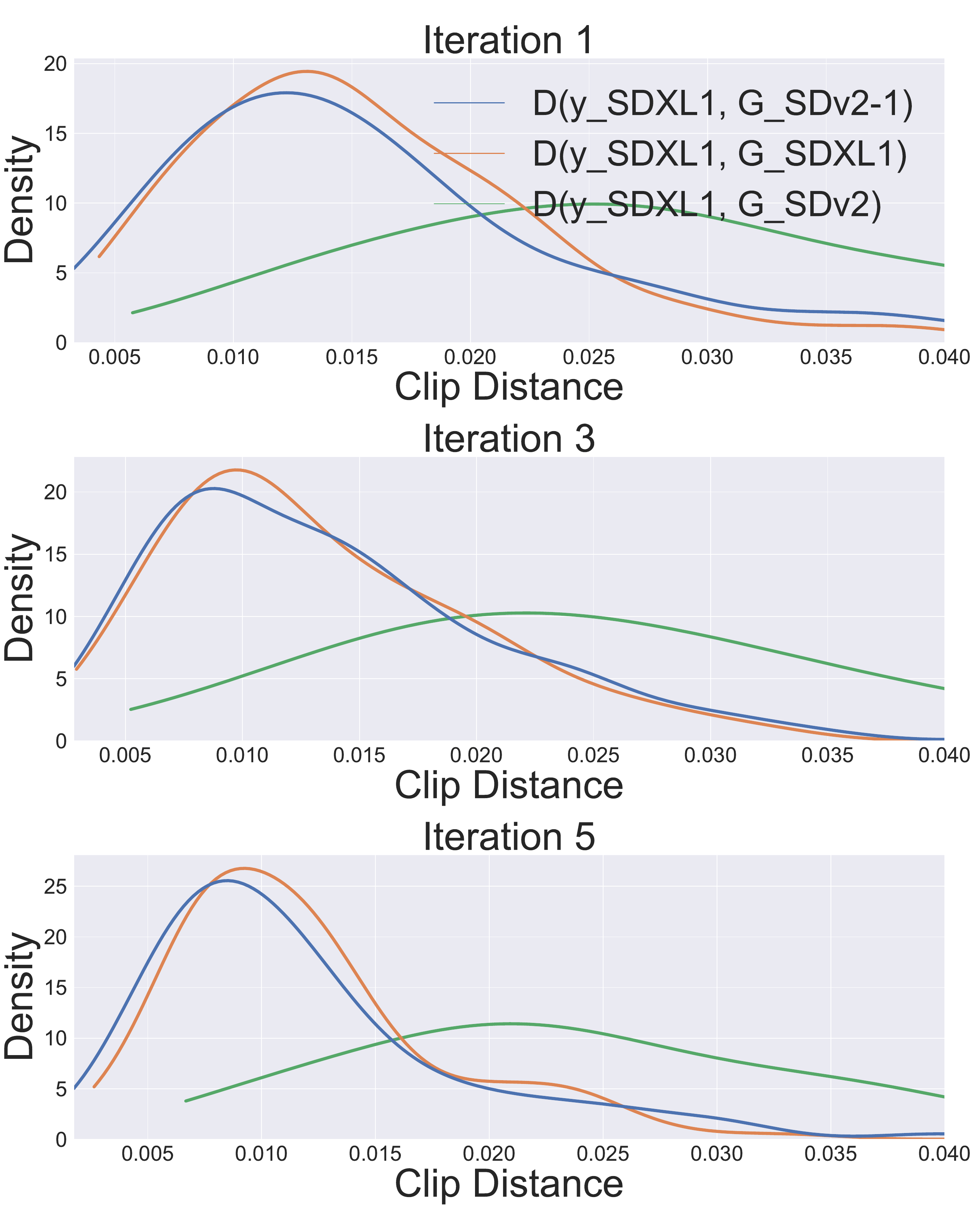}
    \caption{The variations in distance density distribution between paraphrased images and the authentic model, paraphrasing model and a contrast model which is not involved in the generation process of the paraphrased images. The paraphrased images are initially generated by SDv2.1 (authentic model) and then paraphrased by models SDXL 1.0 at different rounds of K (1, 3 \& 5) by the authentic model. Additionally, SDv2 acts as a contrast model for the set of images which is produced by SDv2.1 and then paraphrased by SDXL 1.0. }
    \label{fig:density-distribution-comparison-paraphrase}
    \vspace{-4mm}
\end{figure}

The results, represented in \Figref{fig:density-distribution-comparison-paraphrase}, show comparisons such as $\mathbb{D}(A\_B, \mathcal{G}_A(A\_B))$ and $\mathbb{D}(A\_B, \mathcal{G}_B(A\_B))$, indicating the distance between the paraphrased image (created by A then paraphrased by B) and re-generated images.

We observe a significant challenge in attributing authorship to the original creator model once the image is plagiarized (even at higher values of re-generation iteration K in Algorithm~\ref{alg:gen_iter}). The distance between the paraphrased image to the authentic model and the paraphrasing model cannot be differentiated making it increasingly difficult to distinguish whether the SDv2.1 model or another paraphrasing model generated the image. 
This overlap in the density distributions is a limitation in our methodology, which relies on discerning the nuances in these distances for authorship attribution. This is a common weakness of any fingerprinting or watermarking methods where there is no provable robustness guarantee ~\cite{10.1145/3576915.3623189}.

However, the results also reveal a noteworthy strength in the methodology. There is a discernible and consistently large distance when comparing paraphrased images to those generated by a contrast model (\ie an unrelated model, one which was neither the authentic nor the paraphrasing model). This finding is crucial as it suggests that despite the challenges in direct attribution, our approach can reliably detect when an external model, not involved in the image's original creation or its paraphrasing, is falsely claimed as the author. Therefore, this method could potentially be leveraged to establish joint authorship in scenarios where the paraphrased image is claimed by a third party, ensuring proper credit is given to the authentic and paraphrasing models involved in the image's creation.

We investigate perturbation attacks in both text and image generation tasks. For our NLP experiments, we randomly perturb $X$\% of the words preceding the one-step re-generation procedure by substituting them with random words. The value of $X$ ranges from 10\% to 50\%. We designate \mthree as the authentic model. We set $k=5$, due to its outperforming performance. As depicted in \Tabref{tab:nlp_robustness_evaluation}, as the perturbation rate increases, we observe a gradual decline in the verification performance of our method. In general, the efficacy of watermark verification remains reasonably positive. Nevertheless, we argue that perturbations in the range of 30\% to 50\% have a notable adverse impact on the quality of the generated texts, rendering them significantly less valuable for potential attackers.

\input{tables/tab_nlp_robustness}

Regarding image generation, we introduce two types of perturbations: Gaussian Noise and Brightness alterations.
In the first robustness experiment, Gaussian noise is applied to a random subset of pixels (r) in the original image before re-generation, using a fixed mean ($\mu$ = 0) and standard deviation ($\sigma$ = 4). The second robustness assessment involves modifying the brightness of 10\% of randomly selected pixels in the images. As expected, as the perturbation intensity increases, our verification precision decreases. When we introduced Gaussian noise to different proportions of pixels, we observed minimal impact on verification accuracy across various models, as evidenced in \Tabref{tab:robustness_evaluation}. Even at a 50\% noise level, the models maintained a precision range of 89-99\%. This highlights the robustness of the proposed approach, even in the presence of high levels of random noise. In contrast, brightness perturbations reveal that the watermark is more sensitive to changes in brightness than to Gaussian noise. Examining \Tabref{tab:brightness_evaluation}, we observe a sharp decline in precision for SDXL 1.0, which remains relatively stable until a perturbation rate of 1.02. Beyond this threshold, the model's robustness is significantly compromised.

\input{tables/tab_cv_robustness_noise}

\input{tables/tab_cv_robustness_brightness}

\begin{figure}[t]
    \centering
    \includegraphics[width=0.9\linewidth]{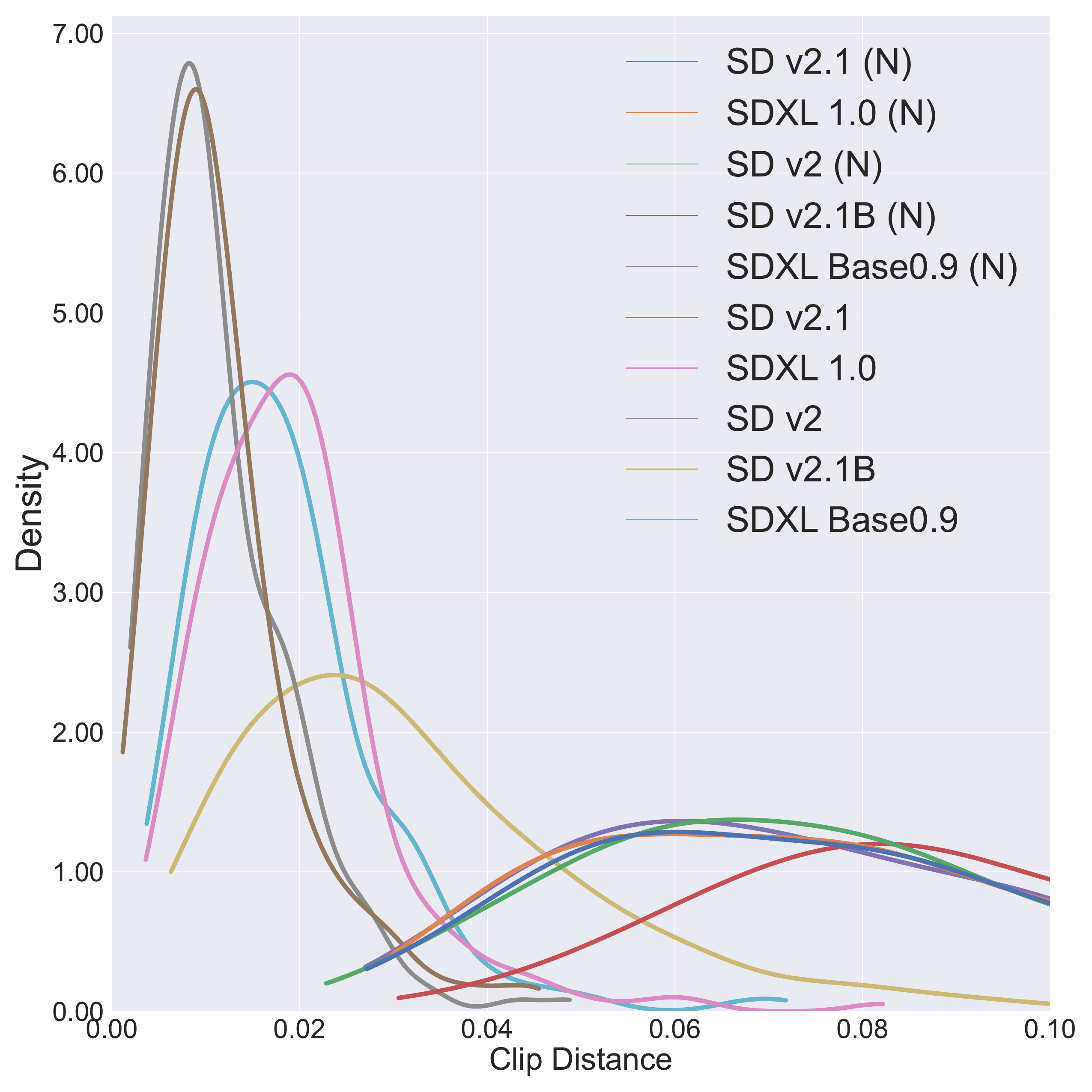}
    \caption{Comparison of Density Distribution of Re-generation Distances in Verification Stage, acquired by Natural vs AI Image Re-generations.} 
    
    \label{fig:Comparison of Density Distribution of Distances Natual vs AI Image Regenerations}
\end{figure}

\paragraph{Distance Differentiation between Natural and AI-Generated Images} In order to assess the viability of our framework to detect plagiarism on naturally made images (i.e., shot on camera), we re-generate natural images from the MS COCO 2017 Evaluation set \cite{lin2014microsoft} with previously used image generation models for one iteration. We compare the Clip distances between these natural images and their respective re-generations to the distance between the AI-generated images and their re-generation for one iteration by the corresponding models as presented in ~\Figref{fig:Comparison of Density Distribution of Distances Natual vs AI Image Regenerations}. There is a clear distinction in the distance density distribution when AI-generated images are re-generated using AI our Generation Algorithm \ref{alg:gen_iter}, the distance remains low whereas using a natural image for re-generation results in further distances even with the most advanced AI models.

%% file: tables/tab_paraphrasing.tex
\begin{table}
    \centering
    \caption{The precision of verifying the authentic models ($\mathcal G_{a}$) using different contrasting models ($\mathcal G_c$). The target tasks are paraphrasing and summarization.}
    \scalebox{0.9}{
    \begin{tabular}{c ccc}
    \toprule
    \multirow{1}{*}{\backslashbox{$\mathcal G_{a}$}{$\mathcal G_c$}}& \multicolumn{1}{c}{\textbf{GPT3.5-turbo}} & \multicolumn{1}{c}{\textbf{Zephyr}} &  \multicolumn{1}{c}{\textbf{Mistral}} \\
    & Precision $\uparrow$ &Precision $\uparrow$ & Precision $\uparrow$ \\
    \bottomrule
    \\[-0.8em]
    \multicolumn{4}{c}{\texttt{Paraphrasing}}
    \\[0.1em]
    \toprule
    \textbf{GPT3.5-turbo} &     - & 75.0 & 69.0	 \\
    \textbf{Zephyr}	 & 84.0	& -	& 69.0 \\
\textbf{Mistral} & 74.0	 & 76.0	 & - \\

    \bottomrule
    \\[-0.8em]
    \multicolumn{4}{c}{\texttt{Summarization}}
    \\[0.1em]
    \toprule
\textbf{GPT3.5-turbo} & -& 92.0	 & 68.0	\\
\textbf{Zephyr}	& 88.0 & -	& 67.0 \\
\textbf{Mistral}	& 86.0	& 86.0	& - \\
 \bottomrule
    \end{tabular}
    }
    \vspace{-0.3cm}
    \label{tab:add_nlp_tasks}
\end{table}

%% file: tables/tab_nlp_robustness.tex

\begin{table}
    \centering
    \caption{The precision of robustness evaluation for text generation with perturbed substitution when \(k=5\).}
    \scalebox{1}{
    \begin{tabular}{c | c  c  c}
    \toprule
    \textbf{Perturbation rates} & {\textbf{M2M}} & {\textbf{mBART}} & {\textbf{Cohere}} \\
    \midrule
    \ \ 0\%       & 90.0 & 94.0 & 95.0 \\
    10\%    & 91.0 & 91.0 & 97.0  \\
    20\%     & 84.0 & 84.0 & 95.0 \\
    30\%     & 84.0  & 74.0 &  94.0 \\
    40\%     & 79.0 & 75.0 & 93.0 \\
    50\%        & 69.0 &  65.0 &  92.0 \\
    \bottomrule
    \end{tabular}}
    \label{tab:nlp_robustness_evaluation}
\end{table}

%% file: tables/tab_cv_robustness_noise.tex

\begin{table}
    \centering
    \caption{The precision of robustness evaluation for image generation with Gaussian Noise Perturbation with Gaussian Noise \(\mu=0, \sigma=4\) when \(k=5\).}
    \scalebox{1}{
    \begin{tabular}{c | c | c | c}
    \toprule
    \textbf{Perturbation rates} & {\textbf{SD v2}} & {\textbf{SD v2.1B}} & {\textbf{SDXL 1.0}} \\
    \midrule
    0.00       & \ \ 76.0 &  \ \ 97.5 & 94.0 \\
    0.01    & \ \ 82.0 &  \ \ 94.5  & 82.0 \\
    0.05    & 100.0  &  \ \ 98.5  & 91.0 \\
    0.10     & \ \ 97.0  & \ \ 99.0  & 95.0  \\
    0.20     & \ \ 99.5 & 100.0&  93.5  \\
    0.50     & \ \ 99.0 & \ \ 99.0  & 89.5  \\
    1.00       & \ \ 99.5  & \ \ 99.0 & 66.5  \\
    \bottomrule
    \end{tabular}}
    \label{tab:robustness_evaluation}
\end{table}

%% file: tables/tab_cv_robustness_brightness.tex

\begin{table}
    \centering
    \caption{The precision of robustness evaluation for image generation with Brightness Perturbation when \(k=5\).}
    \scalebox{1}{
    \begin{tabular}{c | c | c | c}
    \toprule
    \textbf{Perturbation rate} & {c}{\textbf{SD v2}} & {\textbf{SD v2.1B}} & {\textbf{SDXL 1.0}} \\
    \midrule
    0.00       & 76.0 & 97.5 &  94.0  \\
    1.01    & 35.0 & 97.0 & 85.0  \\
    1.02    & 41.5 & 91.0 & 87.0 \\
    1.05    & 34.5 & 94.0 &  59.5  \\
    1.10     & 35.0 & 93.0 & 44.0  \\
    1.50     & 23.5 & 91.5  & 41.5  \\
    \bottomrule
    \end{tabular}}
    \label{tab:brightness_evaluation}
\end{table}

%% file: sec6_conclusion.tex
\section{Conclusion}

In this work, we observe the intrinsic fingerprint in both text and vision generative models, which can be identified and verified by contrasting the re-generation of the suspicious data samples by authentic models and contrasting models. Furthermore, we propose iterative re-generation in the Generation Stage to enhance the fingerprints and provide a theoretical framework to ground the convergence of one-step re-generation distance by the authentic model. Our research paves the way towards a generalized authorship authentication for deep generative models without (1) modifying the original generative model, (2) post-processing to the generated outputs, or (3) an additional model for identity classification.


%% file: sec_appendix.tex

\clearpage
\section{Experimental setup for image generation}
\label{app:cv_generation}

\subsection{Embedding Watermark through Inpainting}
\label{app:watermark-detailed}
To analyze the stability and convergence properties of inpainting models, we perform an iterative masked image infilling procedure. Given an input image $\vx$ from model $\mathcal{G}$, we iteratively inpaint with mask $M$:

\begin{equation}
\vx^{\langle k+1\rangle} = \mathcal{G}(\vx^{\langle k\rangle}, M)
\end{equation}

Here, the mask $M$ not only guides the inpainting but also functions as the medium to embed our watermark. As we iteratively inpaint using a mask $M$, the watermark becomes more deeply embedded, serving as a distinctive signature to identify the authentic model $\mathcal{G}_a$. 

\subsubsection{Convergence of Watermarked Images}
\label{app:convergence-watermark}
The iterative masked inpainting procedure displays consistent convergence behavior across models. With a fixed binary mask covering $1/10$th of the image, the distance between successive image generations decreases rapidly over the first few iterations before stabilizing. This is evidenced by the declining trend in metrics like MSE, LPIPS, and CLIP similarity as iterations increase.

\begin{figure*}[!hb]
    \centering
    \includegraphics[width=\linewidth]
    {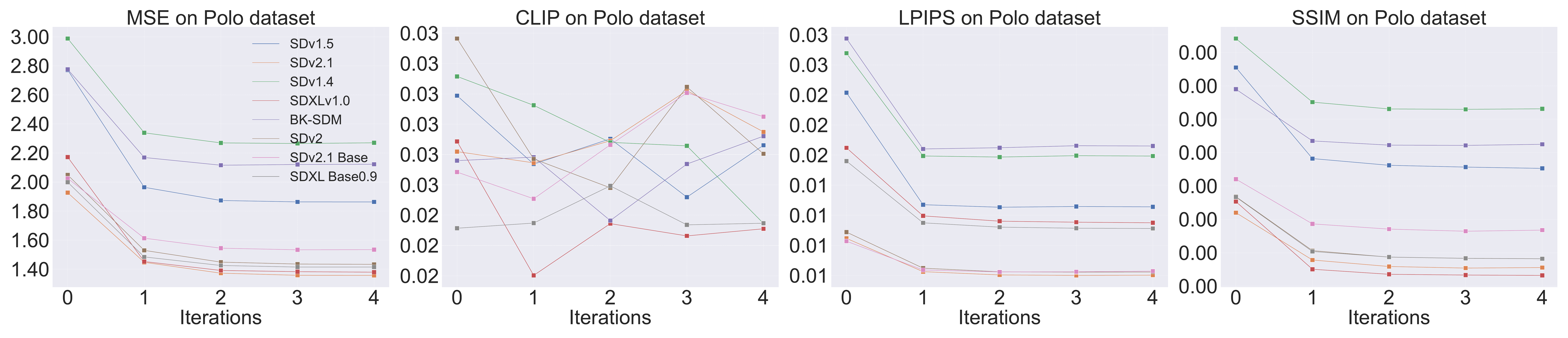}
    \caption{The convergence analysis of the distances in iteration based on various metrics on the watermarked images of 200 samples from Polo datasets.}
    \label{fig:watermark-polo}
\end{figure*}

\begin{figure*}[!hb]
    \centering
    \includegraphics[width=\linewidth]{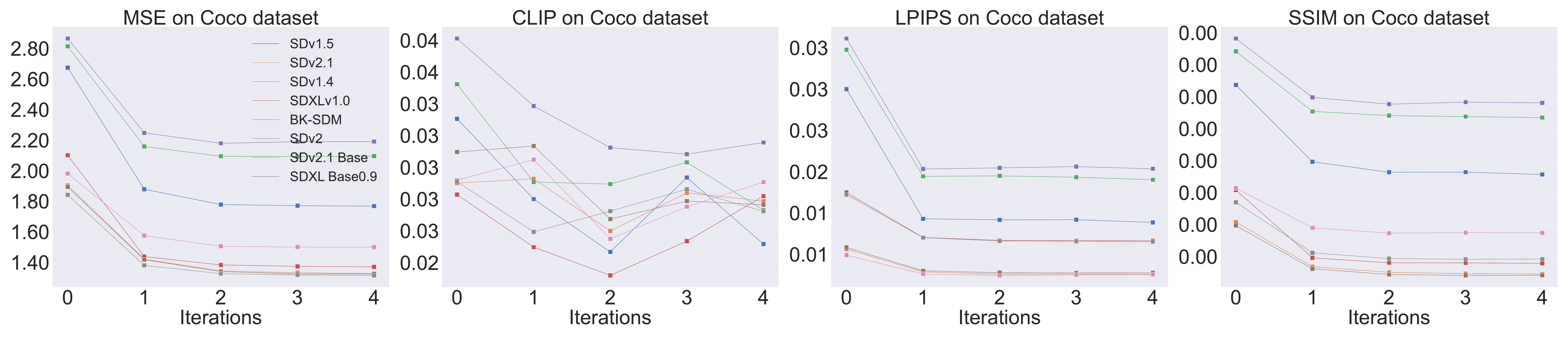}
    \caption{The convergence analysis of the distances in iterations based on various metrics on the watermarked images of 200 samples from Coco dataset}
    \label{fig:watermark-coco}
\end{figure*}

The early convergence suggests the generative models are effectively reconstructing the masked regions in a coherent manner. While perfect reconstruction is infeasible over many passes, the models appear to reach reasonable stability within 5 iterations as shown in \Figref{fig:watermark-polo} and \ref{fig:watermark-coco}.

Convergence to a stable equilibrium highlights latent fingerprints in the model behavior. The consistent self-reconstruction statistics form the basis for distinguishing authentic sources in the subsequent fingerprinting experiments. The watermarking convergence analysis highlights model stability and confirms that iterative inpainting effectively removes embedded watermarks without degrading image quality (see \Figref{app:image_regen_quality}).

\subsection{Enhancing and Verifying Fingerprint through Re-generation}

We divide an image into non-overlapping segments and have models reconstruct masked regions to expose inherent fingerprints. Given an image $x$, we generate segment masks $M_1,..., M_T$ covering $x$ and apply $\mathcal{G}$ to reconstruct each part:

\begin{equation}
\vy{[t]} = \mathcal{G}(\vx, M[t])
\end{equation}

By analyzing reconstructed segments $\vy{[t]}$ and the composited image $\vy$, model-specific artifacts can be quantified without additional pattern information. The non-overlapping reconstructed segments are merged to form the composite image $y$, we can obtain this by using the corresponding mask and extracting the reconstructed portion. A salient feature of our proposed re-generation paradigm is its independence from any additional information about the image partitioning pattern. The model's unique fingerprint emerges naturally during re-generation, regardless of how the image is divided or the components are merged. We experiment on 5 latest SD models - \SDvTwoOne, \SDvTwoOneB, \SDvTwo, \SDXLOneZero, and \SDXLZeroNineB using 8 segmented masks each covering $1/8$th of the image thereby totaling full image coverage. Model fingerprints are identified through LPIPS, and CLIP similarities between original $x_a$ and reconstructed $y$.

\subsubsection{Convergence of Re-generated Images}

\begin{figure*}[ht]
    \centering
    \includegraphics[width=\linewidth]{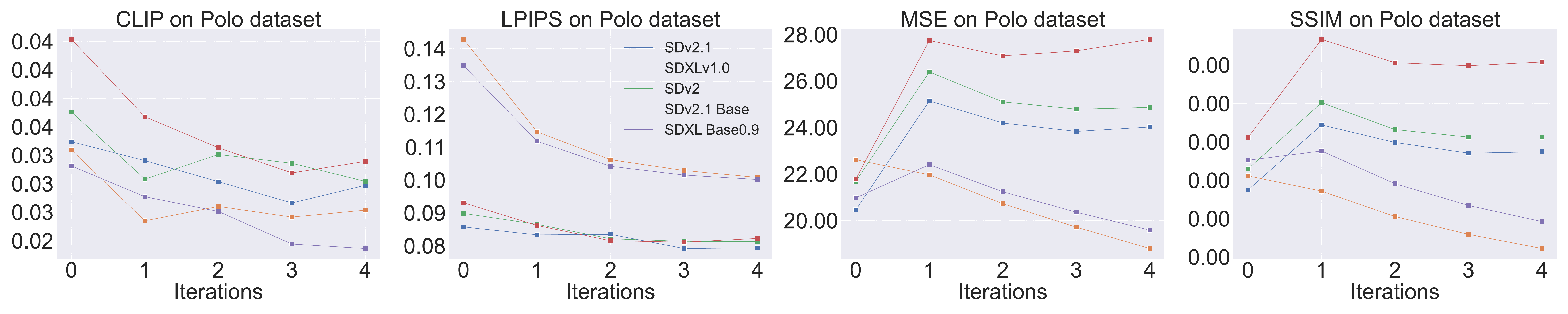}
    \caption{The convergence analysis of the distances in iteration based on various metrics on the re-generated images of 200 samples from Polo dataset.}
    \label{fig:regen-polo-all}
\end{figure*}

\begin{figure*}[ht]
    \centering
    \includegraphics[width=\linewidth]{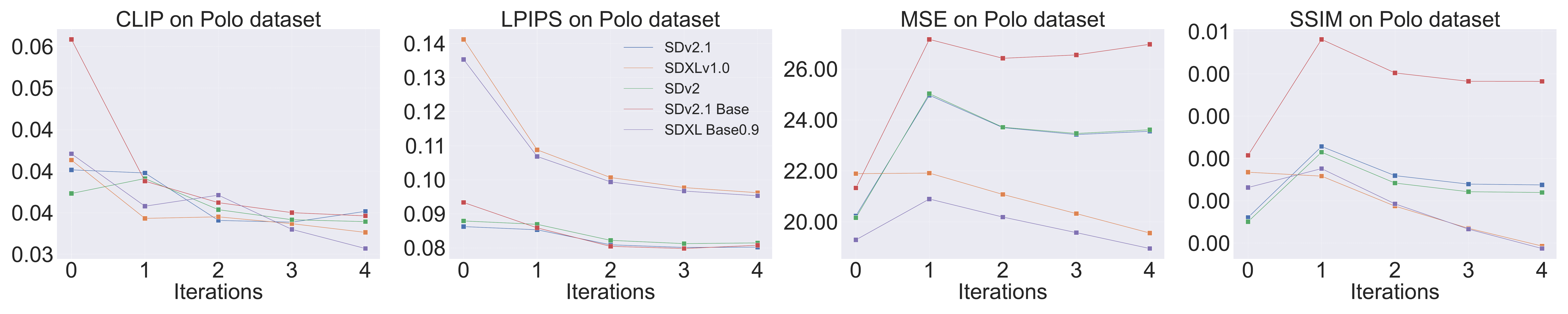}
    \caption{The convergence analysis of the distances in iteration based on various metrics on the re-generated images of 200 samples from Coco dataset.}
    \label{fig:regen-coco-all}
\end{figure*}

Enhancing fingerprint re-generation shows consistent convergence in perceptual metrics like CLIP and LPIPS within four iterations (see \Figref{fig:regen-polo-all} and \Figref{fig:regen-coco-all}). However, traditional metrics such as MSE and SSIM lack clear convergence, suggesting inpainting effectively captures visual content but not at the pixel level. The models converge to unique stable points, revealing inherent fingerprints based on their biases and training data. This divergence is important for model attribution. Overall,  re-generation effectively exposes these fingerprints while maintaining visual integrity, underscoring perceptual superiority over pixel-based metrics in evaluating generative model fingerprints.
\label{app:convergence-regeneration}

\section{Computer Vision Supplementary Experiments and Details}
\subsection{Computer Vision Generative Models}
\label{app:cv_models}
We consider eight models based on the Stable Diffusion architecture~\citep{Rombach_2022_CVPR}. These models leverage the architecture and differ primarily in their training schemes. The models are selected to span a range of architectures, training schemes, and dataset sizes. This diversity allows us to explore model-specific behaviors for attribution and stability analysis.

\begin{table}[ht]
    \centering
    \caption{Summary of models based on the Stable Diffusion architecture for IG experiments. The \textbf{bold} models are used in our primary experiments.}
    \begin{tabularx}{\linewidth}{c|X}
    \toprule
    \textbf{Model} & \textbf{Description} \\
    \midrule
    \textbf{Model\textsubscript{1}} & Stable Diffusion 1.5: Fine-tuned from SD1.4 for 595k steps~\cite{Rombach_2022_CVPR} \\
    \textbf{Model\textsubscript{2}} & Stable Diffusion 2.1 (\textbf{\SDvTwoOne}): Fine-tuned from SDvTwo for 210k steps~\cite{Rombach_2022_CVPR} \\
    \textbf{Model\textsubscript{3}} & Stable Diffusion 1.4 (\SDOneFour) : Initialized from SDv1.2 weights~\cite{Rombach_2022_CVPR} \\
    \textbf{Model\textsubscript{4}} & Stable Diffusion XL 1.0 \textbf{(\SDXLOneZero}): Larger backbone, trained on LAION-5B~\cite{podell2023sdxl} \\
    \textbf{Model\textsubscript{5}} & Block-removed Knowledge-distilled Stable Diffusion Model~\cite{kim2023architectural} \\
    \textbf{Model\textsubscript{6}} & Stable Diffusion 2 (\textbf{\SDvTwo})~\cite{Rombach_2022_CVPR} \\
    \textbf{Model\textsubscript{7}} & Stable Diffusion v 2.1 Base (\textbf{\SDvTwoOneB})~\cite{Rombach_2022_CVPR} \\
    \textbf{Model\textsubscript{8}} & Stable Diffusion XL 0.9 (\textbf{\SDXLZeroNineB})~\cite{podell2023sdxl} \\
    \bottomrule
    \end{tabularx}
    \label{tab:model_summary}
\end{table}

All models support inpainting, allowing images to be edited given a mask and image. We utilize the inpainting pipeline - StableDiffussion and StableDiffusionXL - provided by HuggingFace~\citep{von-platen-etal-2022-diffusers}.\footnote{\url{https://huggingface.co/docs/diffusers/api/pipelines/stable_diffusion/inpaint}} Both \SDOneFour and \SDvTwo checkpoints were initialized with the weights of the \SDOneFour checkpoints and subsequently fine-tuned on ``laion-aesthetics v2 5+''. \SDXLOneZero employs a larger UNet backbone and a more potent text encoder~\citep{podell2023sdxl}. \BKSDM is designed for efficient text-to-image synthesis, removing blocks from the U-Net and undergoing distillation pre-training on 0.22M LAION text-image pairs~\citep{kim2023architectural}. 

\paragraph{Image Inpainting:}
Image inpainting refers to filling in missing or masked regions of an image to reconstruct the original intact image. A fixed binary mask is applied to cover certain areas of the image. The binary masks are generated as blank images filled with random white pixels. To reconstruct images, the binary mask and original image are passed to the generative model's inpainting pipeline, where the values of the pixels in the masked areas are predicted based on the unmasked context. After inpainting, the reconstructed masked regions are merged back into the re-generated image for future iteration. We utilize the inpainting pipeline\footnote{\url{https://huggingface.co/docs/diffusers/api/pipelines/stable_diffusion/inpaint}} - StableDiffussion and StableDiffusionXL - provided by HuggingFace~\citep{von-platen-etal-2022-diffusers} that enables us to regenerate a given image multiple times by masking different parts of the image.

\subsection{Computer Vision Distance Metrics}

\paragraph{Contrastive Language-Image Pretraining (CLIP) Cosine Distance}
The CLIP model encodes images into high-dimensional feature representations that capture semantic content and meaning~\citep{radford2021learning}. With pre-trained image-text models, we can easily capture the semantic similarity of two images, which is their shared meaning and content, regardless of their visual appearance.

CLIP uses a vision transformer model as the image encoder. The output of the final transformer block can be interpreted as a semantic feature vector describing the content of the image. Images with similar content will have feature vectors close together or aligned in the embedding space.

To compare two images I1 and I2 using CLIP, we can encode them into feature vectors $f1$ and $f2$. The cosine distance between these semantic feature vectors indicates the degree of semantic alignment.
\begin{equation}
\text{cosine\_sim}(f_1, f_2) = \frac{f_1 \cdot f_2}{\|f_1\| \|f_2\|}
\end{equation}

\begin{equation}
\text {cosine\_dist}(f_1, f_2) = {1 - \text{cosine\_sim}(f_1, f_2) }    
\end{equation}

Where \( \cdot \) denotes the dot product and \( \|\cdot\| \) denotes the \( L_2 \) norm. The cosine distance ranges from 0 to 1, with 0 indicating perfectly aligned features. A lower distance between images implies more similar high-level content and meaning in the images as captured by the CLIP feature embeddings. We specifically use OpenClip with a ConvNext-XXLarge encoder pretrained on laion2b dataset.

\paragraph{Learned Perceptual Image Patch Similarity (LPIPS)}

LPIPS metric focuses on perceptual and stylistic similarities, by using a convolutional neural network pretrained on human judgements of image patch similarities. The distance is measured between the CNN's intermediate feature representations of two images~\citep{zhang2018unreasonable}.

\begin{equation}
d(x, x') = \sum_{l} \frac{1}{{H_l W_l}} \sum_{h,w} \left\| w_l \odot (\hat{y}_l^{hw} - \hat{y}_l'^{hw}) \right\|_2^2
\end{equation}

By comparing features across corresponding layers of the CNN, LPIPS can provide a fine-grained distance measuring subtle perceptual differences imperceptible in pixel space~\citep{zhang2018unreasonable}. For example, changes to color scheme or artistic style that maintain semantic content will have a low CLIP distance but a higher LPIPS distance. We use the original implementation of~\cite{zhang_2023}

\paragraph{Mean Squared Error (MSE)}
The mean squared error (MSE) between two images $x$ and $x'$ is calculated as:

$$MSE(x, x') = \frac{1}{mn} \sum_{i=0}^{m-1} \sum_{j=0}^{n-1} [x(i,j) - x'(i,j)]^2$$

Where $x$ and $x'$ are $m \times n$ images represented as matrices of pixel intensities. The MSE measures the average of the squared intensity differences between corresponding pixels in $x$ and $x'$. 

It provides a simple pixel-level similarity metric sensitive to distortions like noise, blurring, and coloring errors. However, MSE lacks perceptual relevance and is not robust to geometric/structural changes in the image.

\paragraph{Structural Similarity Index (SSIM)}
The structural similarity index (SSIM)~\citep{hore2010image} compares corresponding $8\times8$ windows in the images across three terms - luminance, contrast, and structure:

$$SSIM(x, x') = \frac{(2\mu_x\mu_{x'}+c_1)(2\sigma_{xx'}+c_2)}{(\mu_x^2+\mu_{x'}^2+c_1)(\sigma_x^2+\sigma_{x'}^2+c_2)}$$

Where $\mu_x$, $\mu_{x'}$ are mean intensities, $\sigma_x^2$, $\sigma_{x'}^2$ are variances, and $\sigma_{xx'}$ is covariance for windows in $x$ and $x'$. $c_1$, $c_2$ stabilize division.

This decomposes similarity into comparative measurements of structure, luminance, and contrast. As a result, SSIM better matches human perceptual judgments compared to MSE. Values range from -1 to 1, with 1 being identical local structure.

As research rapidly improves generation quality, we expect future use cases to leverage such advanced generators. Analyzing these models is thus more indicative of real-world conditions going forward compared to earlier versions. Furthermore, the marked quality improvements in recent SD models present greater challenges for attribution and fingerprinting. Subtle inter-model differences become more difficult to quantify amidst high-fidelity outputs. Distance metrics like MSE and LPIPS are sensitive to quality, so lower baseline distortion is a more rigorous test scenario. By evaluating cutting-edge models without inpainting specialization, we aim to benchmark model fingerprinting efficacy on contemporary quality levels. Our experiments on the latest SD variants at scale also assess generalization across diverse high-fidelity generators. Successful attribution and stability analysis under these conditions will highlight the viability of our proposed techniques in real-world deployment.

\subsection{Empirical Estimation of the Lipschitz Constant}
\label{app:lipschitz_empirical_est}
For the empirical estimation of L We use all pairs of images ($\vx$ and $\vy$) from Polo dataset, representing them as inputs and applying the re-generation function $f(\cdot)$. We measure their distances using $\mathbb D$, i.e. Euclidean and LPIPS. The ratio below is the transformation of Equation \ref{eq:lipschitz}:

\begin{equation}
    L \leq \frac{\mathbb D(\vx,\vy)}{\mathbb D(f(\vx),f(\vy))}
\end{equation}

\input{tables/tab_lipschitz_estimation}

From Table \ref{tab:lipschitz_est}, we csn observe LPIPS as a distance metric consistently produced lower L values compared to Euclidean distance. This demonstrates LPIPS as a superior metric for our framework, explaining why Euclidean distance fails for verification. Additionally, more advanced models exhibited improved L values, indicating better pertubation resilience and verification capacity. Our approach focuses on leveraging inherent model properties for verification rather than enforcing universal viability. For models currently unsupported, we suggest targeting enhancements to model robustness in line with these observations.

Overall, the experiment validated LPIPS as an optimal distance function and revealed a correlation between model advancement and verifiability via intrinsic fingerprints. 

\subsection{Sample Prompts and Images Reproduced}

In our computer vision experiments, we sample prompts from the MS-COCO 2017 Evaluation ~\citep{lin2014microsoft} and POLOCLUB (POLO) Diffusion Dataset ~\citep{wang2022diffusiondb} for image generation. We present a few example prompts and images produced for different models in the section below.

\subsubsection{COCO Dataset Prompts}

\begin{itemize}
  \item A motorcycle with its brake extended standing outside.
  \item Off white toilet with a faucet and controls.
  \item A group of scooters rides down a street.
\end{itemize}

\subsubsection{Polo Dataset Prompts}

\begin{itemize}
  \item A renaissance portrait of Dwayne Johnson, art in the style of Rembrandt!! Intricate. Ultra detailed, oil on canvas, wet-on-wet technique, pay attention to facial details, highly realistic, cinematic lightning, intricate textures, illusionistic detail.
  \item Epic 3D, become legend shiji! GPU mecha controlled by telepathic hackers, made of liquid, bubbles, crystals, and mangroves, Houdini SideFX, perfect render, ArtStation trending, by Jeremy Mann, Tsutomu Nihei and Ilya Kuvshinov.
  \item An airbrush painting of a cyber war machine scene in area 5 1 by Destiny Womack, Gregoire Boonzaier, Harrison Fisher, Richard Dadd.
\end{itemize}

\subsection{Density Distribution of a one-step re-generation}
\label{app:density_cv}

\begin{figure*}[!htb]
    \centering
    
    \begin{subfigure} 
        \centering
        \includegraphics[width=\textwidth]{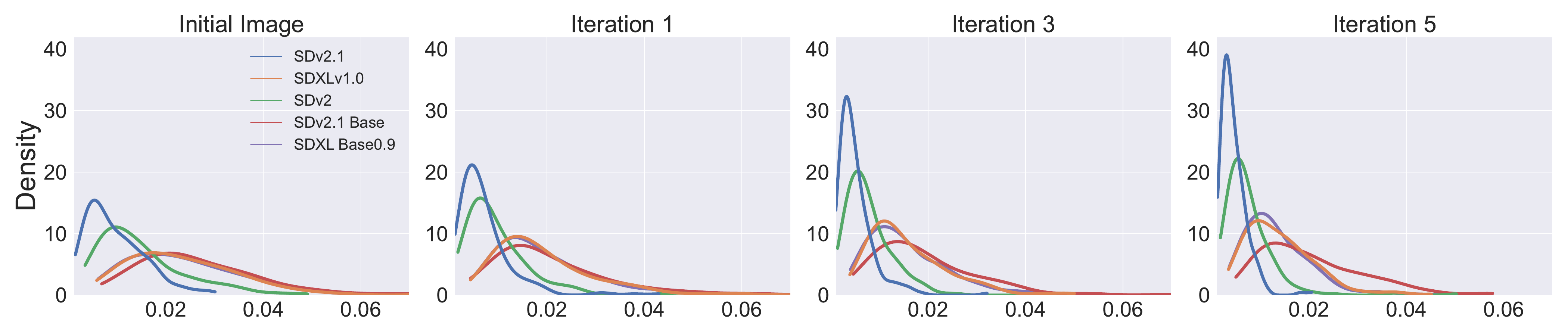}
        \label{fig:sdv2.1}
    \end{subfigure}
    \hfill

    \begin{subfigure}
        \centering
        \includegraphics[width=\textwidth]{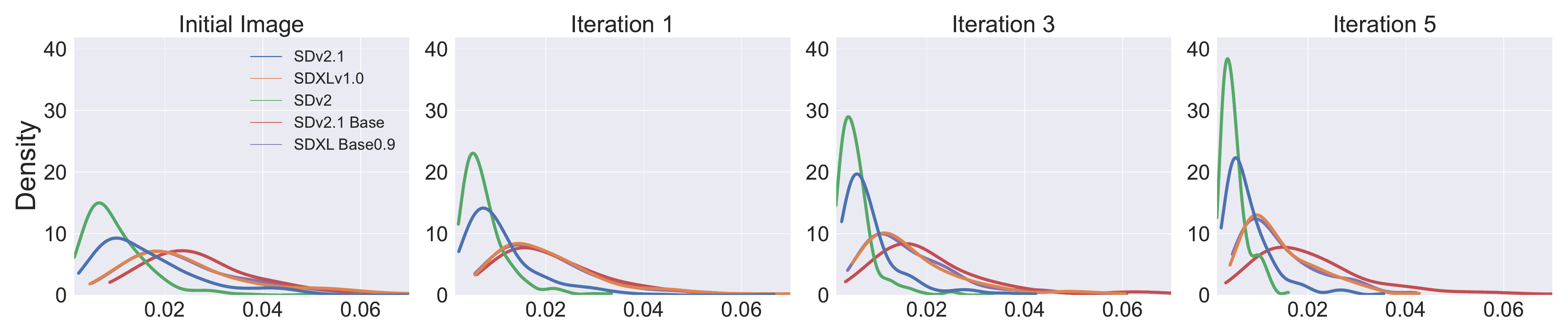}
        \label{fig:sdv2}
    \end{subfigure}
    \hfill
    \begin{subfigure}
        \centering
        \includegraphics[width=\textwidth]{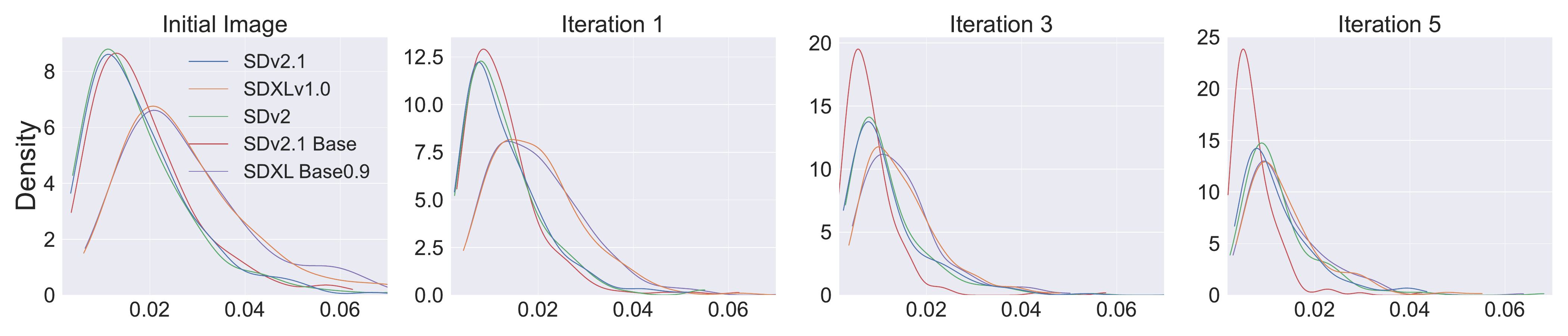}
        \label{fig:sdv2.1base}
    \end{subfigure}
    \hfill
    \begin{subfigure} 
        \centering
        \includegraphics[width=\textwidth]{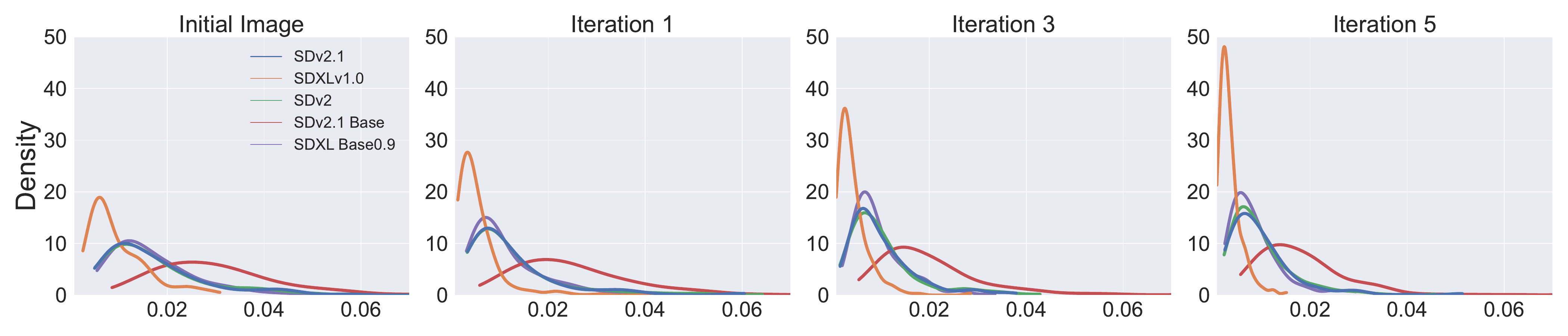}
        \label{fig:sdxl1.0}
    \end{subfigure}
    \hfill
    \begin{subfigure}  
        \centering
        \includegraphics[width=\textwidth]{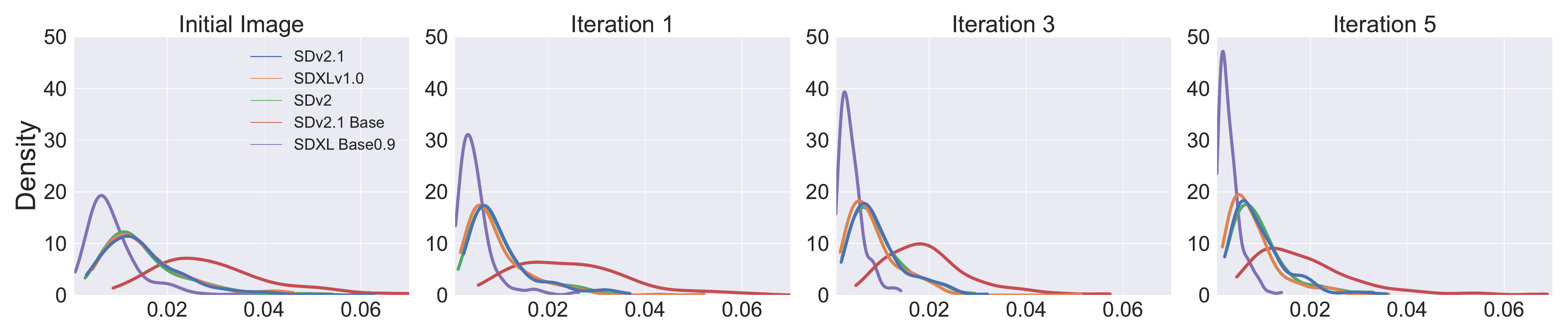}
        \label{fig:sdxl0.9}
    \end{subfigure}
    
    \caption{One Step Re-generation for various authentic models on the Polo Dataset using the CLIP metric at different iterations. The authentic models from top to bottom are: 1) \SDvTwoOne, 2) \SDvTwo, 3) \SDvTwoOneB,4) \SDXLOneZero, 5) \SDXLZeroNineB.
 }
    \label{fig:one-step-regeneration-cluster-polo}

\end{figure*}

\begin{figure*}[!htb]
    \centering
    \begin{subfigure}  
        \centering
        \includegraphics[width=\textwidth]{figures/One_Step_Discrimination/Enhanced_Regeneration_Authentic_Model_Model_2_Coco_Dataset_CLIP.pdf}
    \end{subfigure}
    \hfill
    \begin{subfigure} 
        \centering
        \includegraphics[width=\textwidth]{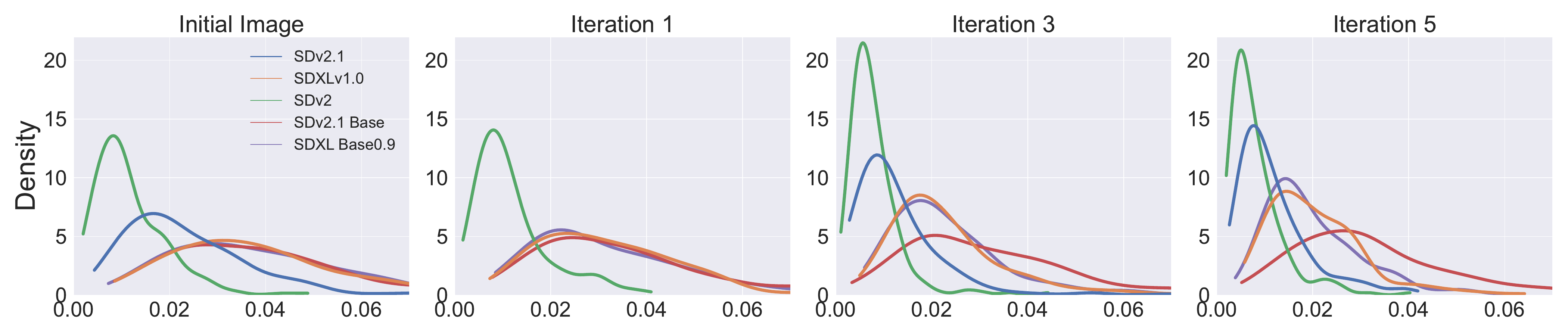}
    \end{subfigure}
    \hfill
    \begin{subfigure}  
        \centering
        \includegraphics[width=\textwidth]{figures/One_Step_Discrimination/Enhanced_Regeneration_Authentic_Model_Model_7_Coco_Dataset_CLIP.pdf}
    \end{subfigure}
    \hfill
    \begin{subfigure} 
        \centering
        \includegraphics[width=\textwidth]{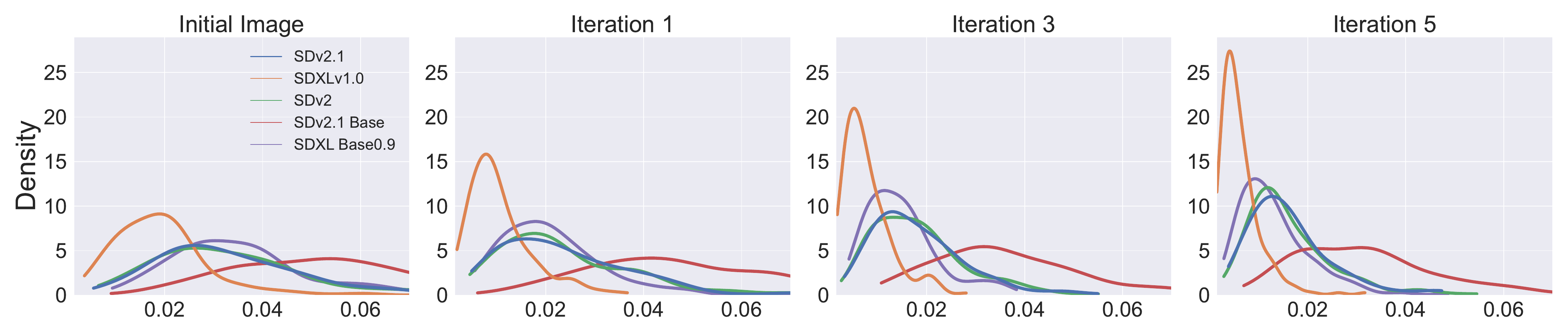}
    \end{subfigure}
    \hfill
    \begin{subfigure}  
        \centering
        \includegraphics[width=\textwidth]{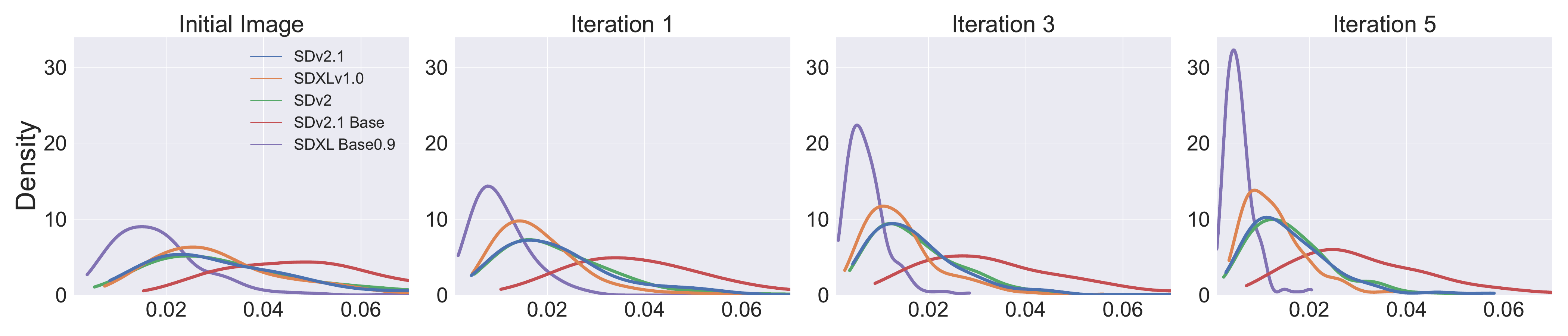}
    \end{subfigure}
    
    \caption{One Step Re-generation for various Authentic Models on the Coco Dataset using the CLIP metric at different iterations. The authentic models from top to bottom are: 1) \SDvTwoOne, 2) \SDvTwo, 3) \SDvTwoOneB, 4) \SDXLOneZero, 5) \SDXLZeroNineB.}
    \label{fig:one-step-regeneration-coco}
\end{figure*}

The one-step re-generation density distributions reveal distinct model-specific characteristics, enabling discrimination as seen in Figures \ref{fig:one-step-regeneration-cluster-polo} and \ref{fig:one-step-regeneration-coco}, Most models exhibit distinction, with each distribution showing unique traits. 

A notable exception in this behavior is observed for the \SDvTwoOneB (see Figures \ref{fig:one-step-regeneration-cluster-polo} and \ref{fig:one-step-regeneration-coco}). which initially demonstrates less discrimination. However, over extended iterations, \SDvTwoOneB shows marked improvement, highlighting the capacity for iterative refinement. By the 5th iteration, there is a noticiable improvement in the discriminative nature of its one-step re-generation. This improvement is crucial as it highlights the model's capacity to refine and enhance its re-generative characteristics over time.

\begin{figure*}[!htb]
    \centering
    
    \begin{subfigure}  
        \centering
        \includegraphics[width=\textwidth]{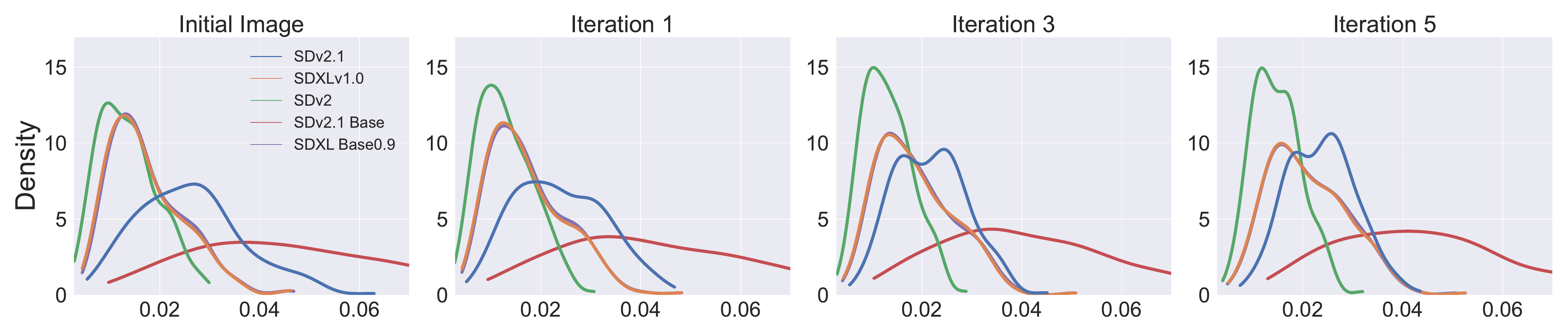}
    \end{subfigure}
    \hfill
    \begin{subfigure}  
        \centering
        \includegraphics[width=\textwidth]{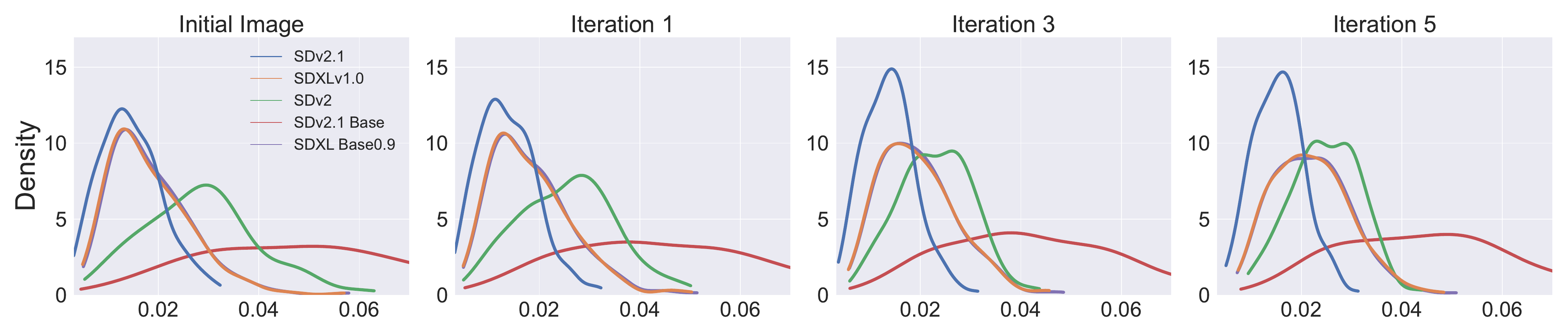}
    \end{subfigure}
    \hfill
    \begin{subfigure} 
        \centering
        \includegraphics[width=\textwidth]{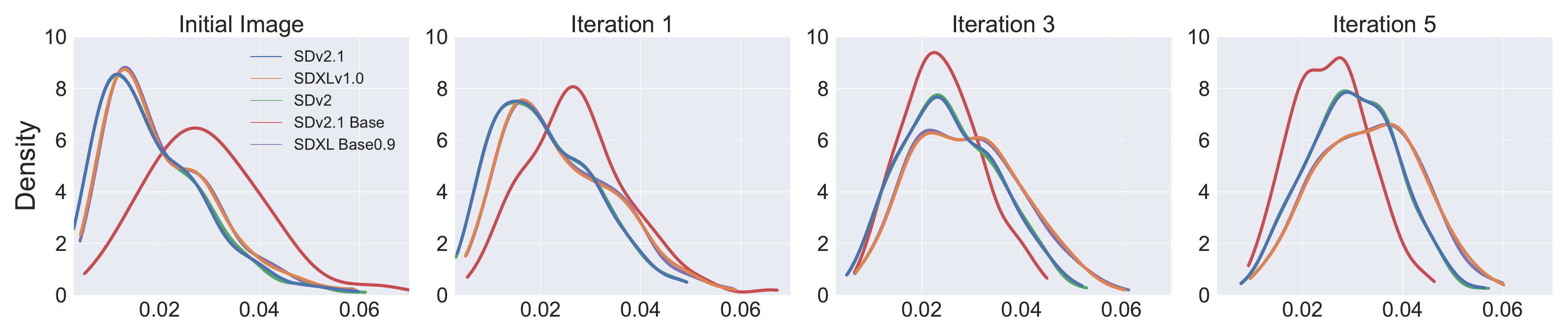}
    \end{subfigure}
    
    \hfill
    \begin{subfigure} 
        \centering
        \includegraphics[width=\textwidth]{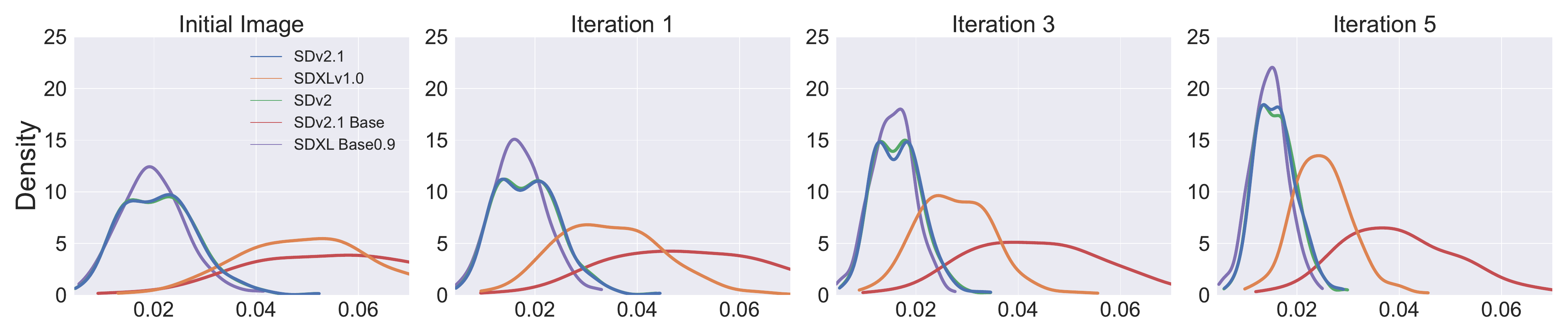}
    \end{subfigure}
    \hfill
    \begin{subfigure} 
        \centering
        \includegraphics[width=\textwidth]{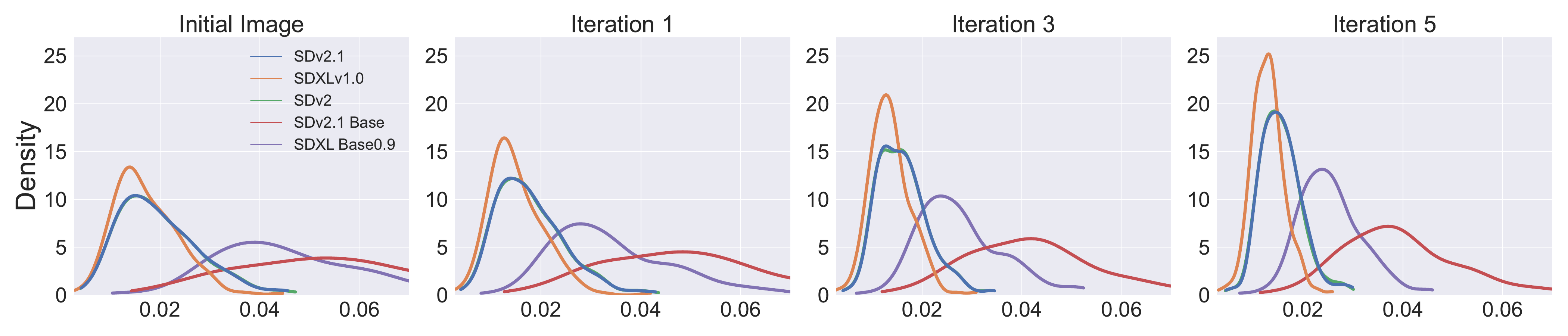}
    \end{subfigure}
    
    \caption{One Step Re-generation for various Authentic Models on the Polo Dataset using the LPIPS metric at different iterations. The authentic models from top to bottom are: 1) \SDvTwoOne, 2) \SDvTwo, 3) \SDvTwoOneB, 4) \SDXLOneZero, 5) \SDXLZeroNineB.}
    \label{fig:one-step-regeneration-lpips-polo}
\end{figure*}

\begin{figure*}[]
    \centering
    \begin{subfigure}  
        \centering
        \includegraphics[width=\textwidth]{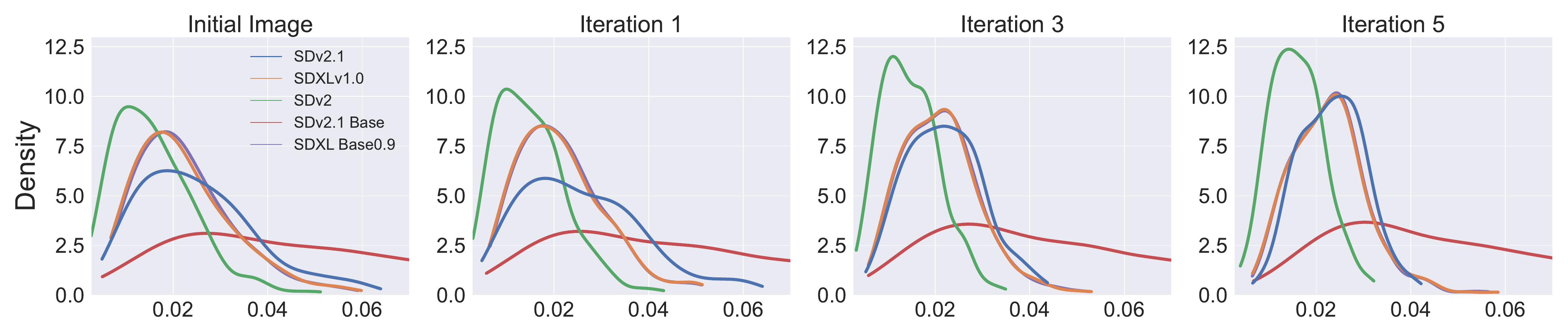}
    \end{subfigure}
    \hfill
    \begin{subfigure} 
        \centering
        \includegraphics[width=\textwidth]{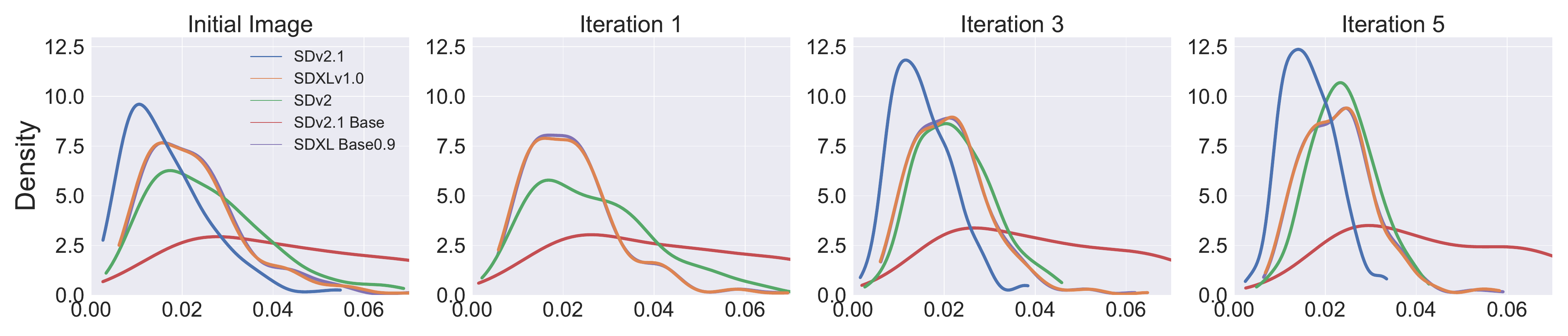}
    \end{subfigure}
    \hfill
    \begin{subfigure}  
        \centering
        \includegraphics[width=\textwidth]{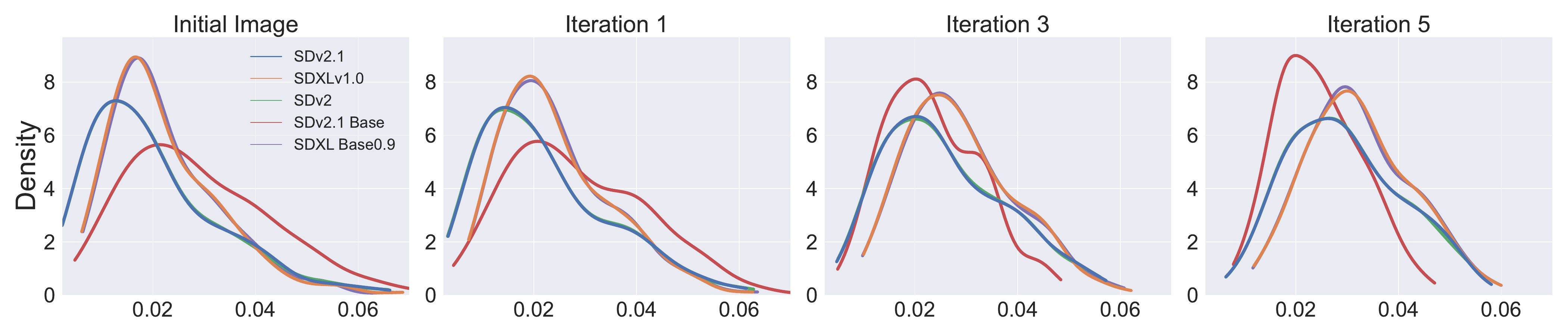}
    \end{subfigure}
    \hfill
    \begin{subfigure} 
        \centering
        \includegraphics[width=\textwidth]{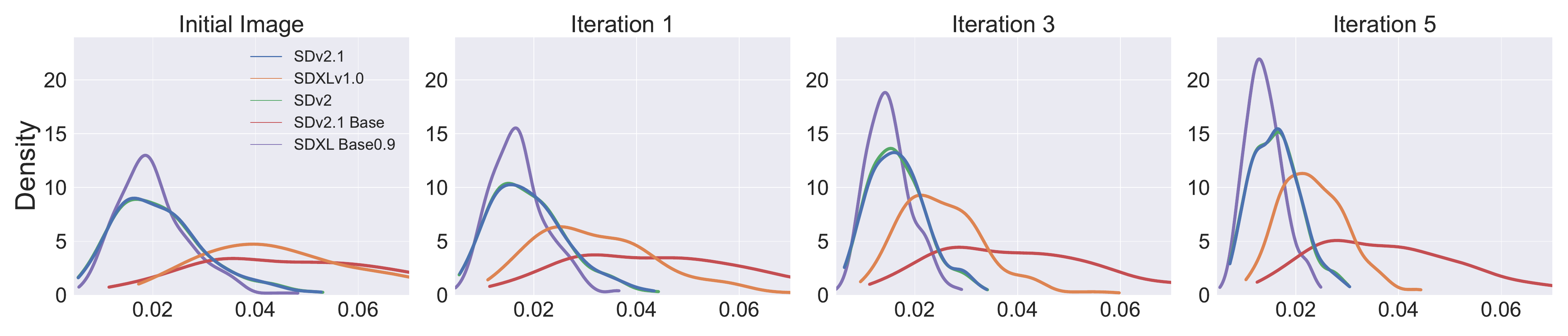}
    \end{subfigure}
    \hfill
    \begin{subfigure}  
        \centering
        \includegraphics[width=\textwidth]{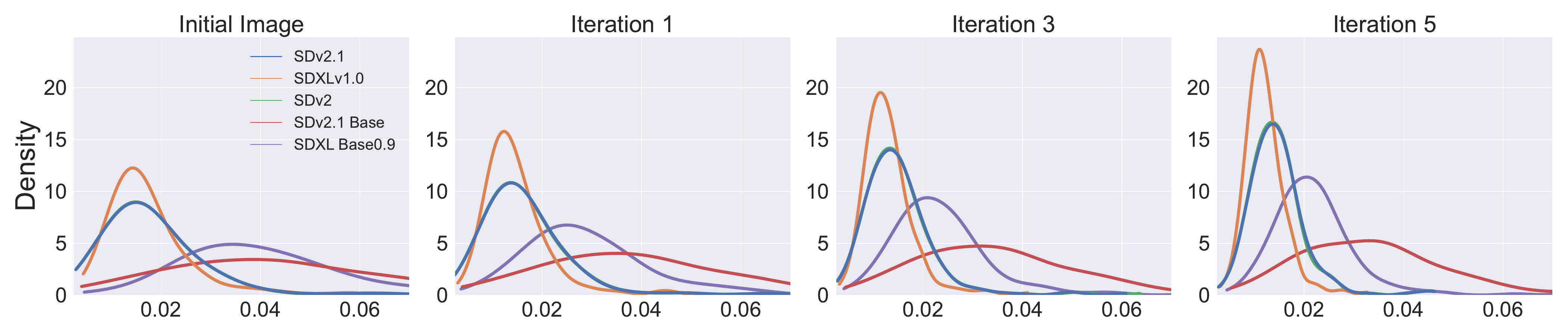}
    \end{subfigure}
    
    \caption{One Step Re-generation for various Authentic Models on the Coco Dataset using the LPIPS metric at different iterations. The authentic models from top to bottom are: 1) \SDvTwoOne, 2) \SDvTwo, 3) \SDvTwoOneB, 4) \SDXLOneZero, 5) \SDXLZeroNineB.}
    \label{fig:one-step-regeneration-coco-lpips}
\end{figure*}

While LPIPS is not immediately effective in pinpointing the authentic model at the very first step, it still offers a powerful mechanism to distinguish between models. LPIPS is effective at differentiating between families of models, such as the Stable Diffusion models and the Stable Diffusion XL models, as visualized in Figures \ref{fig:one-step-regeneration-lpips-polo} and \ref{fig:one-step-regeneration-coco-lpips} for the Polo and Coco dataset correspondingly. The lack of effectiveness of LPIPS in identifying the authentic model is a primary reason why it was not chosen for verification.

\input{tables/tab_cv_verification_coco}

\input{tables/tab_cv_verification_polo}

Further insights into the models' discriminative capabilities can be derived from Table \ref{tab:ACC_all_cv_models_coco} and \ref{tab:ACC_all_cv_models_polo}, while SDv2.1B starts with lower accuracy in distinguishing itself, a significant improvement in accuracy is seen across iterations. The initially anomalous behavior transitions into more discriminating re-generation.

\clearpage
\section{Supplementary experiments for text generation models}
\subsection{Density distribution of a one-step re-generation}
\label{app:density}
\begin{figure*}[h!]
     \centering
     \begin{subfigure}
         \centering
         \includegraphics[width=\textwidth]{figures/nlp/bleu_all_k1.pdf}
     \end{subfigure}
     \hfill
     \vspace{-0.02cm}
     \begin{subfigure}
         \centering
         \includegraphics[width=\textwidth]{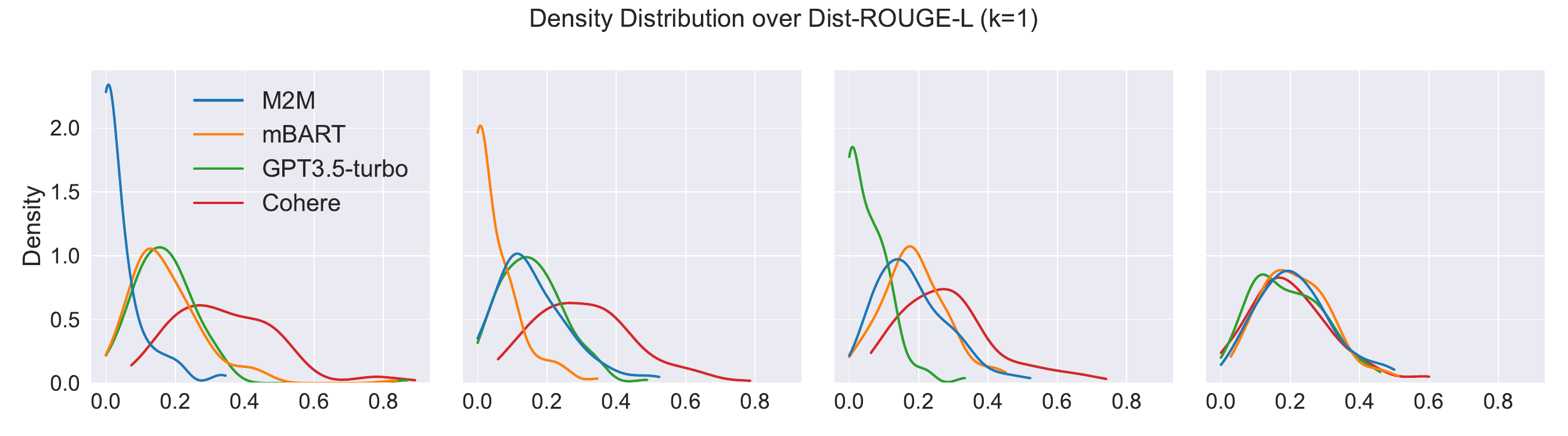}
     \end{subfigure}
     \hfill
     \vspace{-0.02cm}
     \begin{subfigure}
         \centering
         \includegraphics[width=\textwidth]{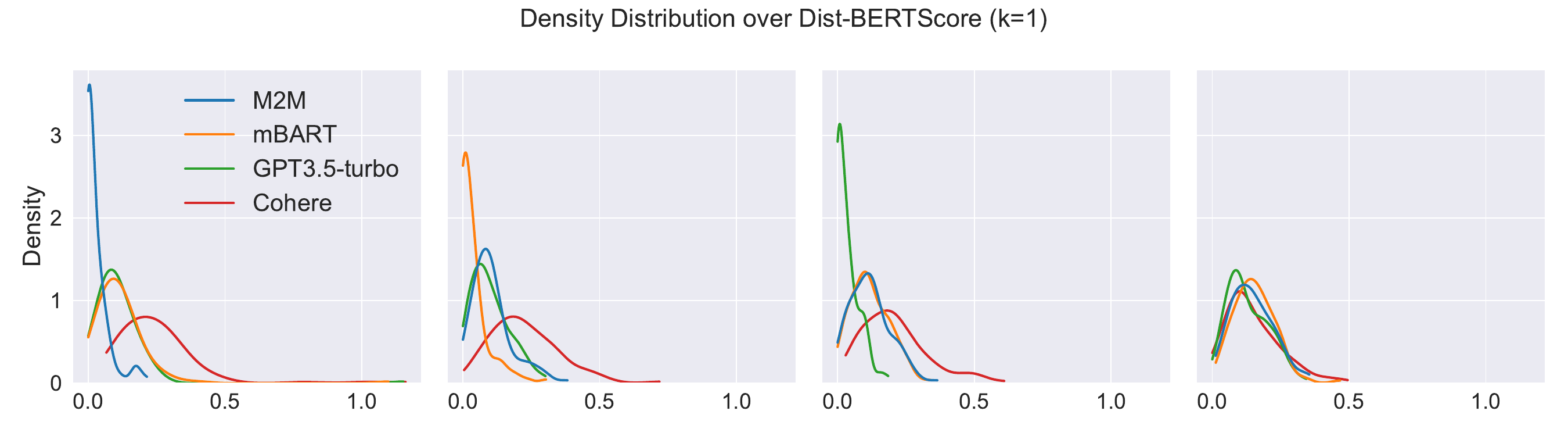}
     \end{subfigure}
        \caption{Density distribution of one-step re-generation among four text generation models, where the input to the one-step re-generation is the first iteration from the authentic models. The authentic models from left to right are: 1) M2M, 2) mBART, 3) GPT3.5-turbo, and 4) Cohere.}
        \label{fig:vary_metrics}
\end{figure*}

In \Figref{fig:vary_metrics}, we illustrate the density distribution of one-step re-generation for four text generation models, using the first iteration from the authentic models as input. Excluding Cohere, the density distributions of the authentic models are discernible from those of the contrast models across BLEU, ROUGE-L, and BERTScore metrics.

\begin{figure*}[h!]
     \centering
     \begin{subfigure}
         \centering
         \includegraphics[width=\textwidth]{figures/nlp/bleu_all_k1.pdf}
     \end{subfigure}
     \hfill
     \vspace{0.02cm}
     \begin{subfigure}
         \centering
         \includegraphics[width=\textwidth]{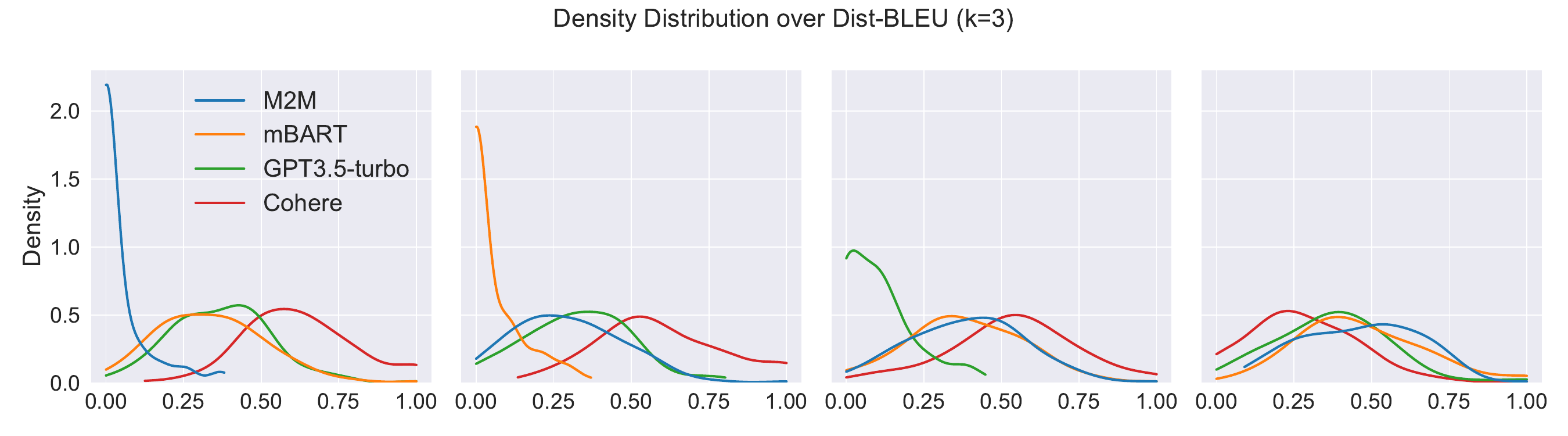}
     \end{subfigure}
     \hfill
     \vspace{0.02cm}
     \begin{subfigure}
         \centering
         \includegraphics[width=\textwidth]{figures/nlp/bleu_all_k5.pdf}
     \end{subfigure}
        \caption{Density distribution of one-step re-generation among four text generation models, where the input to the one-step re-generation is the $k$th iteration from the authentic models. The authentic models from left to right are: 1) M2M, 2) mBART, 3) GPT3.5-turbo, and 4) Cohere.}
        \label{fig:vary_k}
\end{figure*}

\Figref{fig:vary_k} illustrates the density distribution when the input is the $k$th iteration from the authentic models. As $k$ increases, the distinction between the authentic model's distribution and that of the contrast models becomes more pronounced.

\begin{figure*}[h!]
     \centering
     \begin{subfigure}
         \centering
         \includegraphics[width=\textwidth]{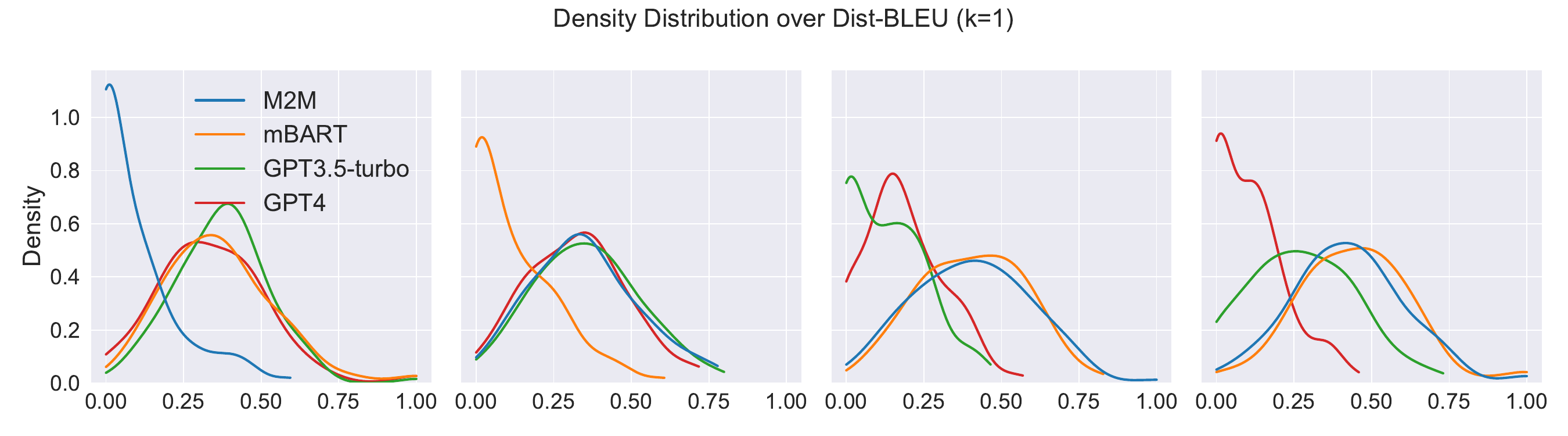}
     \end{subfigure}
     \hfill
     \vspace{0.02cm}
     \begin{subfigure}
         \centering
         \includegraphics[width=\textwidth]{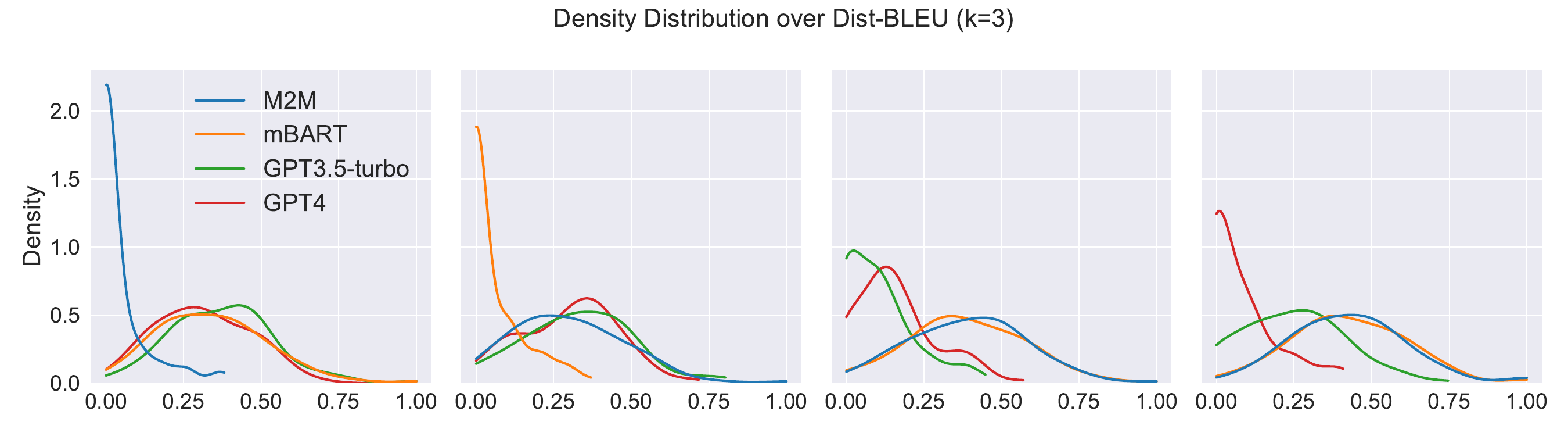}
     \end{subfigure}
     \hfill
     \vspace{0.02cm}
     \begin{subfigure}
         \centering
         \includegraphics[width=\textwidth]{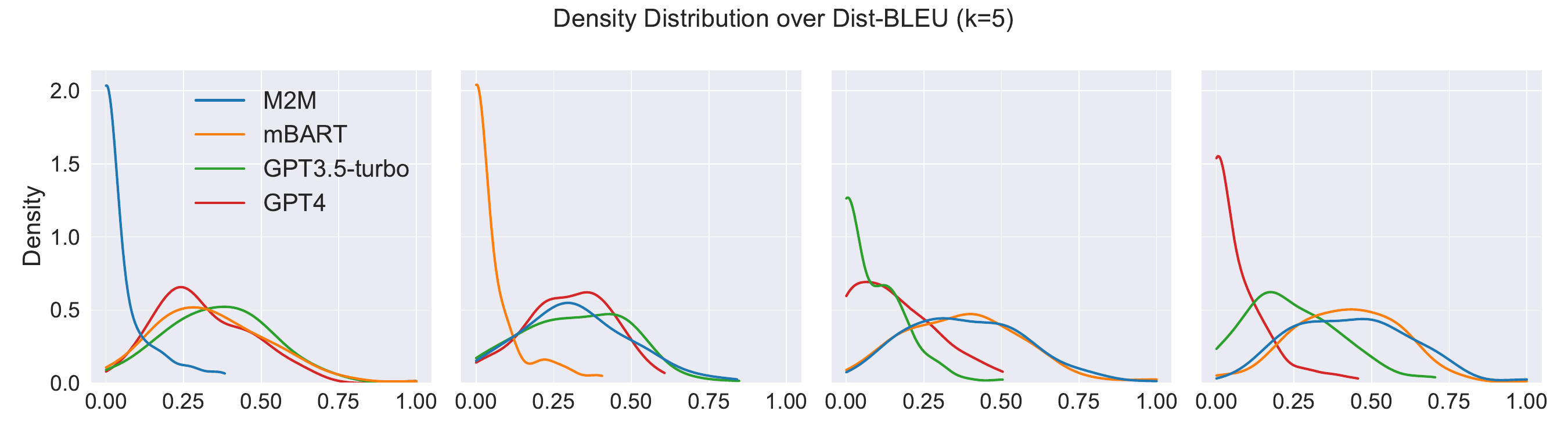}
     \end{subfigure}
        \caption{Density distribution of one-step re-generation among four text generation models, where the input to the one-step re-generation is the $k$th iteration from the authentic models. The authentic models from left to right are: 1) M2M, 2) mBART, 3) GPT3.5-turbo, and 4) GPT4.}
        \label{fig:vary_k_gpt4}
        \vspace{-0.2cm}
\end{figure*}

\input{tables/tab_app_gpt4}

\subsection{Experiments for GPT4}
\label{app:gpt4}
In this section, we analyze the characteristics of GPT4. We first present the density distribution when the input is the $k$th iteration from the authentic models in \Figref{fig:vary_k_gpt4}. Similarly, as $k$ increases, the difference between the authentic model's distribution and the contrast models intensifies. Nonetheless, distinguishing between \mthree and \mfive proves difficult, particularly when \mthree serves as the authentic model, even for larger values of $k$. Detailed examination reveals that \mfive's one-step re-generation bears resemblance to the 
$k$-th iteration of \mthree. This might stem from \mthree and \mfive originating from the same institution, suggesting potential similarities in architecture and pre-training data. Thus, we contend that models stemming from the same institution inherently share identical intellectual property, thus obviating the possibility of intellectual property conflicts. Interestingly, when \mfive is the authentic model, our methodology can differentiate it from \mthree, attributed to the marked difference between \mthree's one-step re-generation and \mfive's $k$-th iteration. This distinction, we surmise, is due to \mfive's superior advancement over \mthree.

In evaluating the verification of authentic models, our methodology, as depicted in \Tabref{tab:ACC_mis_gpt4_models}, consistently confirms the authorship of all models with an accuracy exceeding 85\%. Furthermore, the misclassification rate remains below 10\%.

\begin{figure*}[h!]
    \centering
    \includegraphics[width=\textwidth]{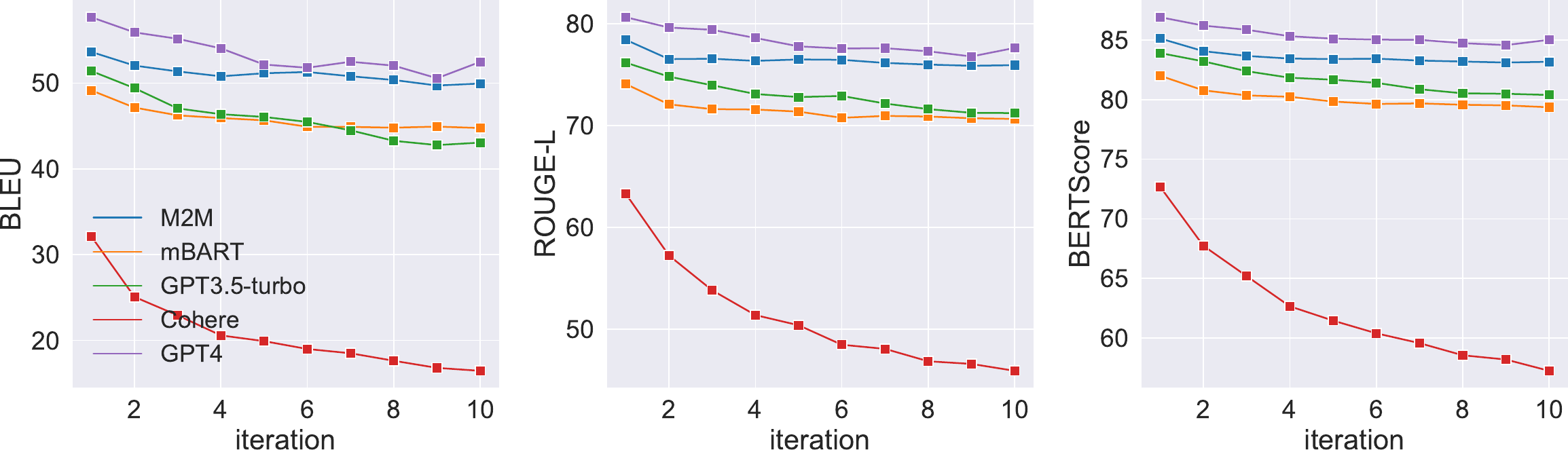}
    \caption{The quality between the original inputs and $k$th re-generation using BLEU, ROULGE-L and BERTScore.}
    \label{fig:regen_quality}
\end{figure*}


\subsection{Quality of Re-generated Sentences and Images}

\subsubsection{Text Re-generations}
\label{app:text_regen}

In this section, we examine the quality of re-generated sentences. As illustrated in \Figref{fig:regen_quality}, the re-generated sentences display high quality across most evaluation metrics, except for Cohere. The slower convergence observed for Cohere can be attributed to its subpar re-generation quality, which significantly deviates from that of other models. For further illustration, refer to the samples and their re-generations in Tables \ref{tab:example_m2m}-\ref{tab:example_cohere}.

\input{tables/tab_app_text_sample}

\subsubsection{Image Re-generations}
\label{app:image_regen_quality}
\begin{figure*}[!hb]
    \centering
    \includegraphics[width=0.95\linewidth]{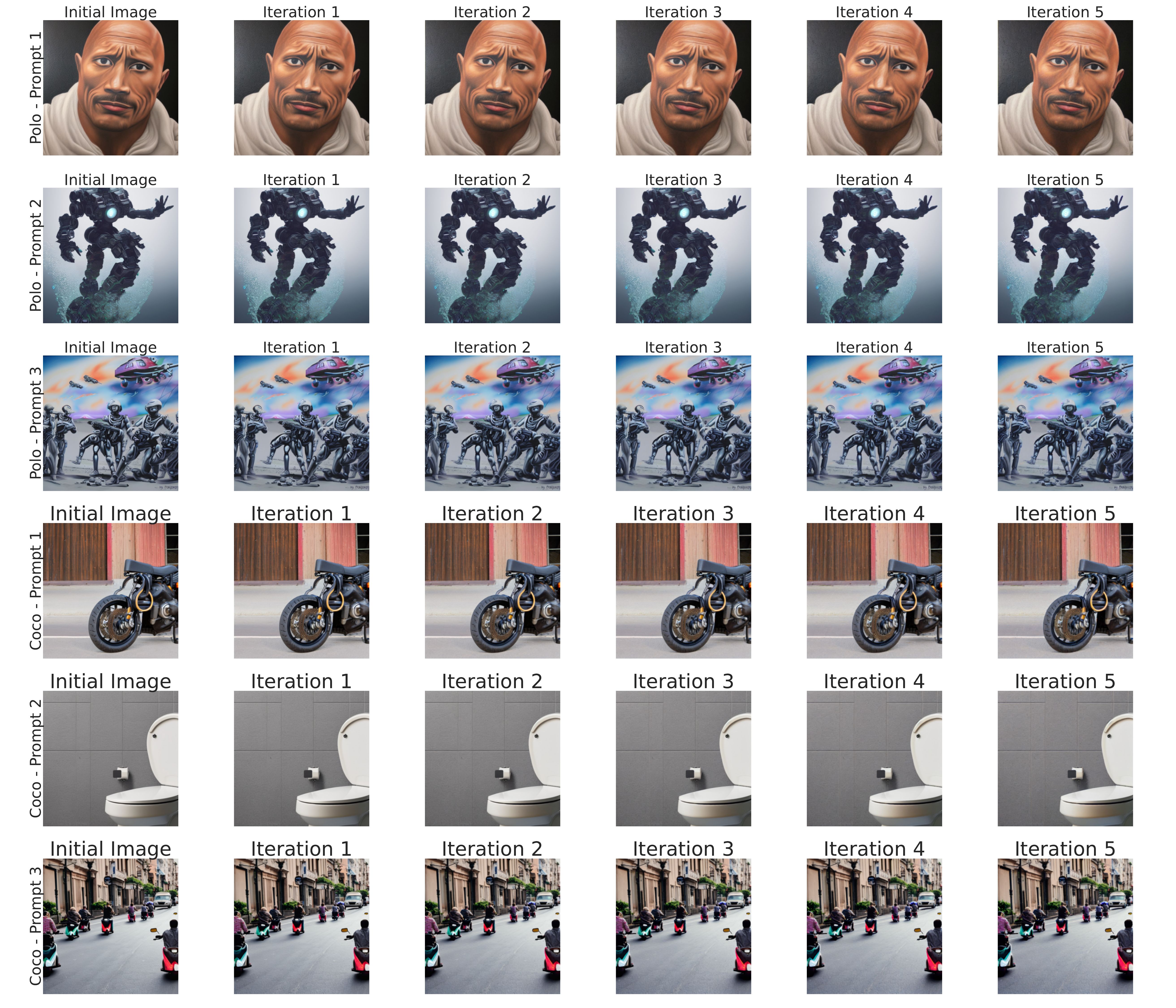}
    \caption{Initial image generation and subsequent re-generations by the SDv2.1 model on Coco and Polo datasets.}
    \label{fig:sdv2.1-regen}
\end{figure*}

\begin{figure*}
    \centering
    \includegraphics[width=0.95\linewidth]{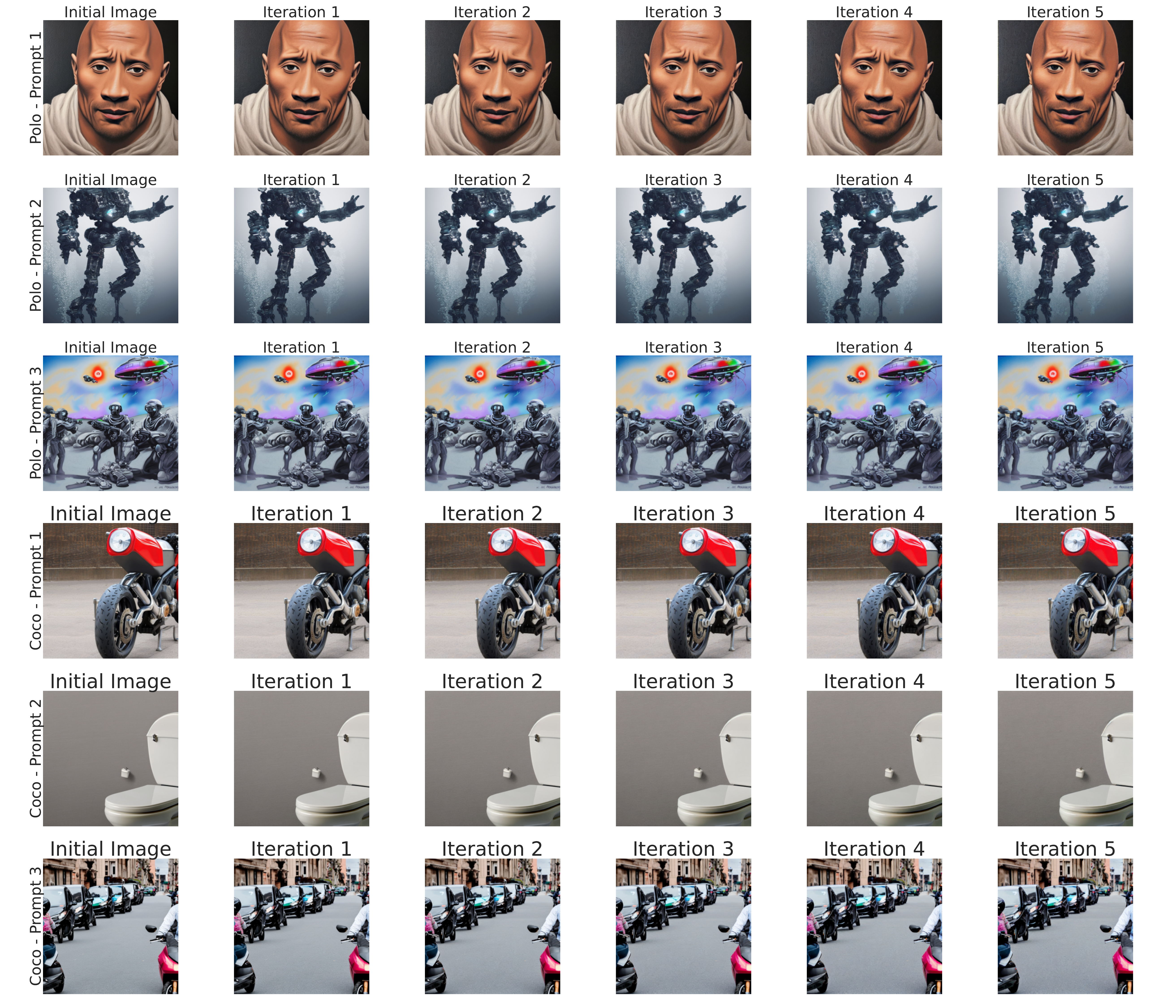}
    \caption{Initial image generation and subsequent re-generations by the SDv2 model on Coco and Polo datasets.}
    \label{fig:sdv2-regen}
\end{figure*}

\begin{figure*}
    \centering
    \includegraphics[width=0.95\linewidth]{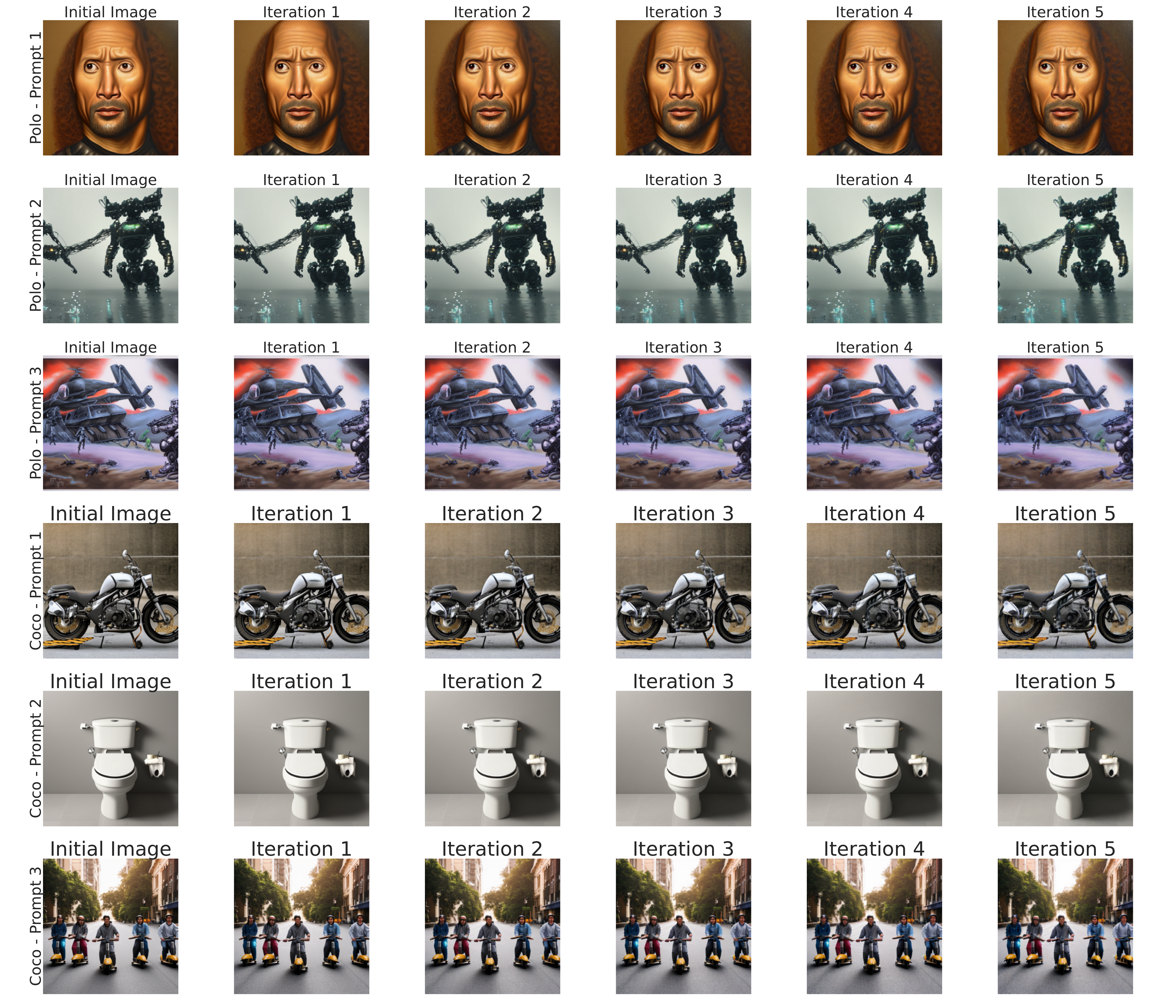}
    \caption{Initial image generation and subsequent re-generations by the SDv2.1B model on Coco and Polo datasets.}
    \label{fig:sdv2.1b-regen}
\end{figure*}

\begin{figure*}
    \centering
    \includegraphics[width=0.95\linewidth]{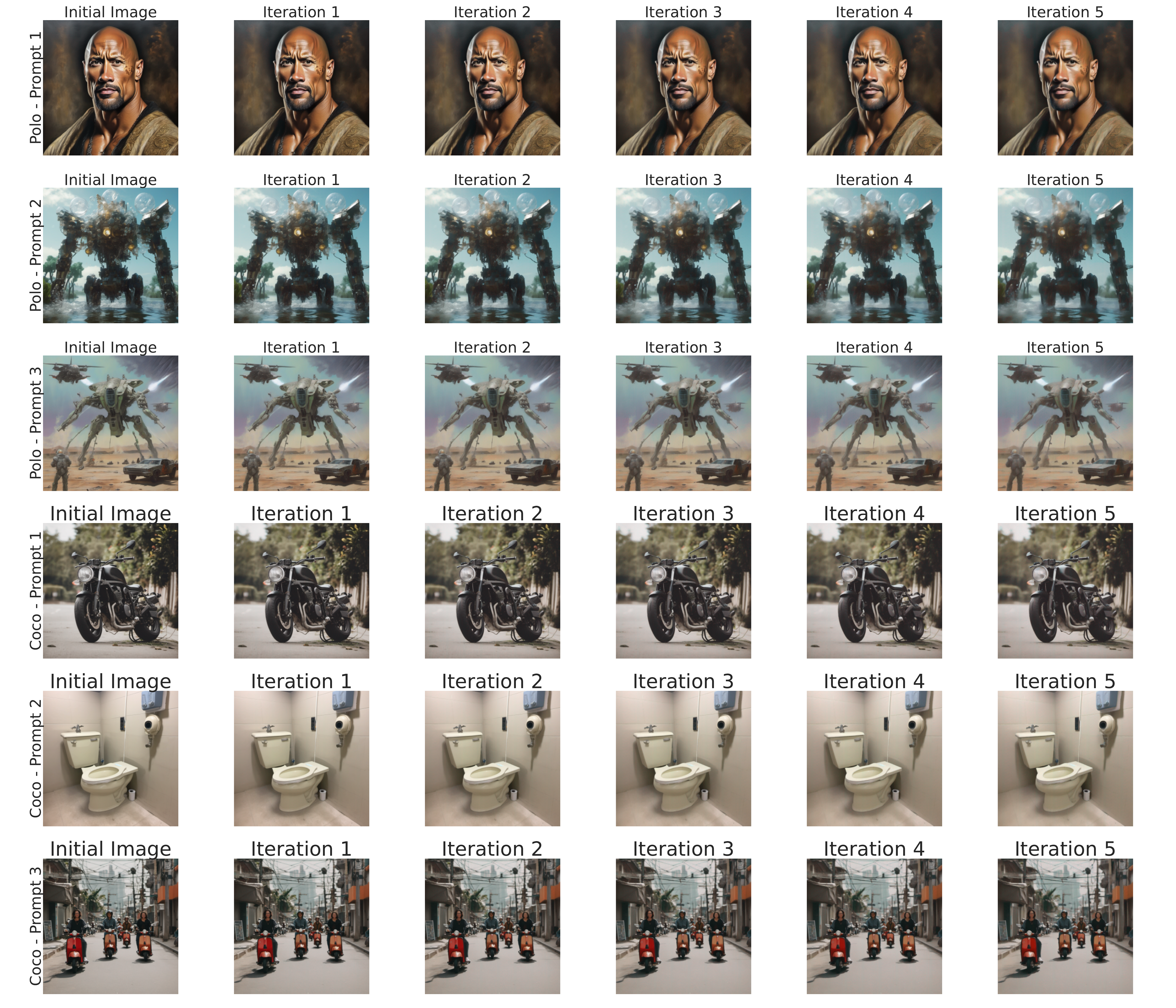}
    \caption{Initial image generation and subsequent re-generations by the SDXL1 model on Coco and Polo datasets.}
    \label{fig:sdxl1-regen}
\end{figure*}

\begin{figure*}
    \centering
    \includegraphics[width=0.95\linewidth]{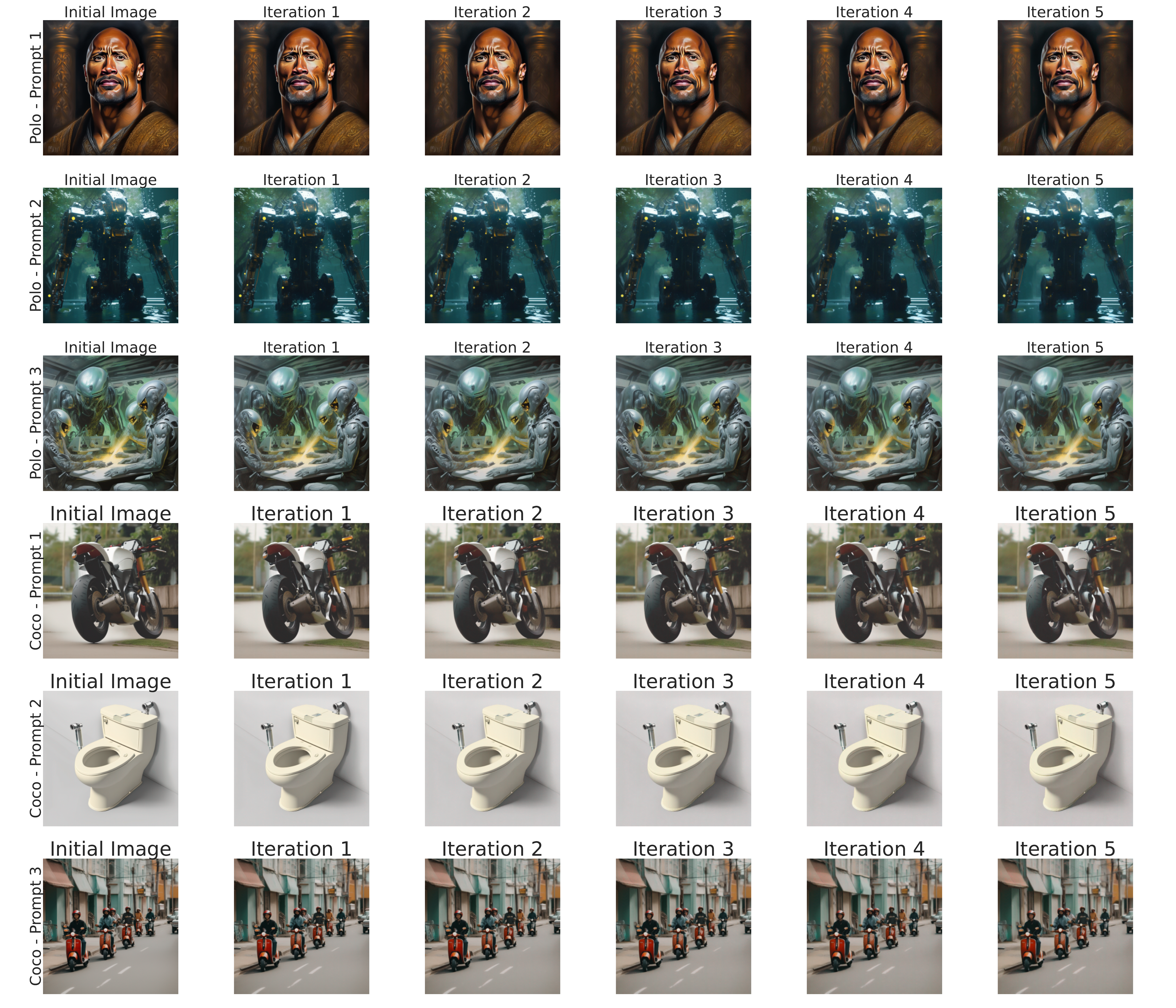}
    \caption{Initial image generation and subsequent re-generations by the SDXL0.9 model on Coco and Polo datasets.}
    \label{fig:sdxl09-regen}
\end{figure*}

An integral part of our evaluation process involves understanding the reproducibility of the images generated by different models across iterations. To this end, we present image re-generations for various authentic models in Figures \ref{fig:sdv2.1-regen}-\ref{fig:sdxl09-regen}. These re-generations provide insights into the consistency and stability of each model when tasked with reproducing the same visual content over multiple iterations.

\clearpage

%% file: tables/tab_lipschitz_estimation.tex
\begin{table*}[htb]
    \centering
    \caption{Lipschitz Constant Estimation}
    \scalebox{1}{
    \begin{tabular}{c | ccccc}
    \toprule
    Model & L\_LPIPS Mean & L\_LPIPS Std & L\_Euclidean Mean & L\_Euclidean Std \\
    \midrule
    \textbf{SD v2.1}   & 0.976 & 0.015 & 1.000 & 0.001 \\
    \textbf{SDXL 1.0}  & 0.943 & 0.21  & 1.000 & 0.004 \\
    \textbf{SD v2}     & 0.975 & 0.016 & 1.000 & 0.001 \\
    \textbf{SD v2.1B}  & 0.970 & 0.023 & 1.000 & 0.002 \\
    \textbf{SDXL 0.9}  & 0.946 & 0.020 & 0.999 & 0.005 \\
    \bottomrule
    \end{tabular}}
    \label{tab:lipschitz_est}
\end{table*}

%% file: tables/tab_cv_verification_coco.tex
\begin{table}[]
    \centering
    \caption{The precision and recall of verifying the authentic models \(\mathcal G_{a}\) using different contrast models \(\mathcal G_c\) on Coco Dataset.}
    \scalebox{0.8}{
    \begin{tabular}{ccccccccccccccc}
    \toprule
    \multirow{2}{*}{\backslashbox{\(\mathcal G_{a}\)}{\(\mathcal G_c\)}} & \multicolumn{2}{c}{\textbf{SD v2.1}} & \multicolumn{2}{c}{\textbf{SD v2.}} & \multicolumn{2}{c}{\textbf{SD v2.1B}} & \multicolumn{2}{c}{\textbf{SDXL 0.9}} & \multicolumn{2}{c}{\textbf{SDXL 1.0}} \\
    \cmidrule{2-11} 
    & Precision \(\uparrow\) & Recall $\uparrow$ & Precision \(\uparrow\) & Recall $\uparrow$ & Precision \(\uparrow\) & Recall $\uparrow$ & Precision \(\uparrow\) & Recall $\uparrow$ & Precision \(\uparrow\) & Recall $\uparrow$  \\
      \bottomrule
    \\[-0.8em]
    \multicolumn{11}{c}{(a) $k=1$}
    \\[0.1em]
    \toprule
    \textbf{SD v2.1} & - & - & 87.5 & \ \ 90.5 & \ \ 97.5 & \ \ 97.5 & 96.5 & 98.0 & 97.0 & 97.5 \\
    \textbf{SD v2.} & 89.0 &  91.5 & - & - & \ \ 97.0 & \ \ 98.0 & 96.0 & 97.0 & 95.5 & 97.5 \\
    \textbf{SD v2.1B} & 14.5 & 19.0 & \ \ 2.0 & \ \ \ \ 3.0 & - & - & 47.5 & 53.0 & 48.5 & 56.0 \\
    \textbf{SDXL 0.9} & 85.5 & 88.5 & 88.0 & \ \ 91.5 & \ \ 99.5 & 100.0 & - & - & 90.5 & 95.0 \\
    \textbf{SDXL 1.0} & 92.5 &  96.0 & 89.5 & \ \ 93.0 & \ \ 98.0 & \ \ 98.5 & 95.5 & 97.0 & - & - \\
    \bottomrule
    \\[-0.8em]
    \multicolumn{11}{c}{(b) $k=3$}
    \\[0.1em]
    \toprule
    \textbf{SD v2.1} & - & - & 73.0 & 77.0 & \ \ 97.5 & \ \ 98.5 & 94.5 &  96.5 & 95.5 &  96.5 \\
    \textbf{SD v2.} & 72.0 & 79.0 & - & - & \ \ 97.0 & \ \ 99.0 & 95.0 & 96.0 & 94.5 & 95.5 \\
    \textbf{SD v2.1B} & 74.0 & 79.0 & 74.5 & 79.5 & - & - & 82.0 & 86.0 & 81.5 & 85.5 \\
    \textbf{SDXL 0.9} & 88.5 & 90.0 & 90.0 & 91.5 & \ \ 98.0 & \ \ 98.5 & - & - & 86.5 & 88.0 \\
    \textbf{SDXL 1.0} & 91.5 & 94.5 & 92.5 &  95.0 & 100.0 & 100.0 & 89.0 & 91.0 & - & - \\
     \bottomrule
    \\[-0.8em]
    \multicolumn{11}{c}{(c) $k=5$}
    \\[0.1em]
    \toprule
    \textbf{SD v2.1} & - & - & 66.0 & 73.0 & \ \ 99.5 & 99.5 & 94.5 &  97.0 & 93.0 & 96.0 \\
    \textbf{SD v2.} & 72.5 & 81.5 & - & - & \ \ 96.0 & 96.5 & 91.0 &  92.5 & 94.0 &  95.0 \\
    \textbf{SD v2.1B} & 78.5 & 83.5 & 77.0 & 82.5 & - & - & 82.5 & 87.0 & 85.5 & 89.5 \\
    \textbf{SDXL 0.9} & 97.0 &  98.0 & 97.5 & 98.0 & 100.0 & 100.0 & - & - & 92.0 &  95.0 \\
    \textbf{SDXL 1.0} & 93.0 &  96.0 & 92.5 &  93.0 & \ \ 99.0 & \ \ 99.5 & 88.0 &  90.5 & - & - \\
    \bottomrule
    \end{tabular}}
    \label{tab:ACC_all_cv_models_coco}
\end{table}

%% file: tables/tab_cv_verification_polo.tex
\begin{table*}[htb!]
    \centering
    \caption{The precision and recall of verifying the authentic models \(\mathcal G_{a}\) using different contrast models \(\mathcal G_c\) on Polo Dataset.}
    \scalebox{0.85}{
    \begin{tabular}{ccccccccccccccc}
    \toprule
    \multirow{2}{*}{\backslashbox{\(\mathcal G_{a}\)}{\(\mathcal G_c\)}} & \multicolumn{2}{c}{\textbf{SD v2.1}} & \multicolumn{2}{c}{\textbf{SD v2.}} & \multicolumn{2}{c}{\textbf{SD v2.1B}} & \multicolumn{2}{c}{\textbf{SDXL 0.9}} & \multicolumn{2}{c}{\textbf{SDXL 1.0}} \\
    \cmidrule{2-11} 
     & Precision \(\uparrow\) & Recall $\uparrow$ & Precision \(\uparrow\) & Recall $\uparrow$ & Precision \(\uparrow\) & Recall $\uparrow$ & Precision \(\uparrow\) & Recall $\uparrow$ & Precision \(\uparrow\) & Recall $\uparrow$  \\
     \bottomrule
    \\[-0.8em]
    \multicolumn{11}{c}{(a) $k=1$}
    \\[0.1em]
    \toprule
    \textbf{SD v2.1} & - & - & 76.0 & 78.5 &  97.5 & 99.0 & 93.5 & 97.0 & 94.0 & 95.0 \\
    \textbf{SD v2.} & 79.0 & 84.0 & - & - &  98.0 & 98.5 & 92.0 & 95.5 & 94.0 &  95.5 \\
    \textbf{SD v2.1B} & 40.5 & 51.0 & 39.0 & 46.0 & - & - & 72.5 & 80.0 & 73.5 & 80.5 \\
    \textbf{SDXL 0.9} & 86.0 & 88.5 & 85.0 & 89.5 &  99.5 & 99.5 & - & - & 91.5 &  95.5 \\
    \textbf{SDXL 1.0} & 89.0 &  92.0 & 89.5 &  93.0 &  98.0 & 99.0 & 91.5 &  95.5 & - & - \\
    \bottomrule
    \\[-0.8em]
    \multicolumn{11}{c}{(b) $k=3$}
    \\[0.1em]
    \toprule
    \textbf{SD v2.1} & - & - & 77.0 & 83.0 & \ \ 99.5 & 100.0 & 95.5 &  98.0 & 96.5 & 97.5 \\
    \textbf{SD v2.} & 75.5 & 82.0 & - & - & \ \ 98.0 & \ \ 98.0 & 96.5 &  98.0 & 97.5 &  97.5 \\
    \textbf{SD v2.1B} & 70.0 & 76.5 & 71.5 & 78.5 & - & - & 84.0 &  90.5 & 84.5 & 89.5 \\
    \textbf{SDXL 0.9} & 92.5 &  95.0 & 91.0 &  94.0 & 100.0 & 100.0 & - & - & 91.0 & 94.0 \\
    \textbf{SDXL 1.0} & 90.0 & 92.5 & 89.5 &   91.5 & 100.0 & 100.0 & 90.0 & 93.0 & - & - \\
    \bottomrule
    \\[-0.8em]
    \multicolumn{11}{c}{(c) $k=5$}
    \\[0.1em]
    \toprule
    \textbf{SD v2.1} & - & - & 80.0 & 86.0 & \ \ 99.5 & \ \ 99.5 & 98.5 &  99.5 & 99.0 &  100.0 \\
    \textbf{SD v2.} & 77.5 & 81.5 & - & - & \ \ 99.0 & 100.0 & 96.5 & 97.0 & 95.5 & \ \ 97.5 \\
    \textbf{SD v2.1B} & 81.5 & 83.5 & 82.5 & 86.5 & - & - & 88.5 & 92.0 & 89.5 & \ \ 91.0 \\
    \textbf{SDXL 0.9} & 97.5 & 97.5 & 96.5 &  98.0 & \ \ 99.5 & \ \ 99.5 & - & - & 92.0 & \ \ 93.5 \\
    \textbf{SDXL 1.0} & 95.5 &  96.0 & 92.5 &  95.5 & 100.0 & \ \ \ \ 0.0 & 94.5 &  96.0 & - & - \\
    \bottomrule
    \end{tabular}}
    \label{tab:ACC_all_cv_models_polo}
\end{table*}

%% file: tables/tab_app_gpt4.tex
\begin{table*}[]
    \centering
    \caption{The precision and recall of verifying the authentic models ($\mathcal G_{a}$) using different contrast models ($\mathcal G_c$).}
    \scalebox{1}{
    \begin{tabular}{ccccccccc}
    \toprule
    \multirow{2}{*}{\backslashbox{$\mathcal G_{a}$}{$\mathcal G_c$}}& \multicolumn{2}{c}{\textbf{M2M}} & \multicolumn{2}{c}{\textbf{mBART}} &  \multicolumn{2}{c}{\textbf{GPT3.5-turbo}} & \multicolumn{2}{c}{\textbf{GPT4}}\\
    \cmidrule{2-9} 
    & Precision $\uparrow$ & Recall $\uparrow$ &Precision $\uparrow$ & Recall $\uparrow$ & Precision $\uparrow$ & Recall $\uparrow$ &  Precision $\uparrow$ & Recall $\uparrow$\\
   \bottomrule
    \\[-0.8em]
    \multicolumn{9}{c}{(a) $k=1$}
    \\[0.1em]
    \toprule
          \textbf{M2M}  & - &-& 94.0 & 98.0 & 93.0 & 95.0 &  87.0 & 92.0 \\
          \textbf{mBART} & 85.0 &89.0 &- & - & 89.0  &92.0&  80.0 & 85.0 \\
          \textbf{GPT3.5-turbo} & 87.0 & 92.0 & 90.0 & 92.0  & - & - & 51.0  & 69.0 \\
          \textbf{GPT4} & 92.0 & 98.0  & 94.0 &98.0  & 77.0 & 90.0 & - & - \\
   \bottomrule
    \\[-0.8em]
    \multicolumn{9}{c}{(b) $k=3$}
    \\[0.1em]
    \toprule
   \textbf{M2M}& - & -& 94.0 & 99.0  & 95.0 & 97.0  & 94.0 & 95.0 \\
\textbf{mBART}&85.0 & 93.0 & - &- & 90.0 & 96.0  & 89.0 & 95.0 \\
\textbf{GPT3.5-turbo} & 89.0 & 92.0 & 93.0 & 95.0 & - & - & 51.0 & 72.0 \\
 \textbf{GPT4} & 96.0 & 97.0 & 92.0 & 94.0 & 79.0 & 91.0 &- & - \\
 \bottomrule
    \\[-0.8em]
    \multicolumn{9}{c}{(c) $k=5$}
    \\[0.1em]
    \toprule
\textbf{M2M} & - & - & 94.0 & 98.0 & 95.0 & 99.0  & 94.0 & 96.0 \\
\textbf{mBART} & 91.0 & 96.0 & - & - & 88.0 & 94.0 & 89.0 & 94.0 \\
\textbf{GPT3.5-turbo} & 90.0 & 94.0 & 94.0 & 98.0 & - & - & 55.0 & 83.0 \\
 \textbf{GPT4} & 97.0 & 98.0 & 94.0 & 97.0 & 83.0 & 95.0 & - & - \\
    \bottomrule
    \end{tabular}
    }
    \label{tab:ACC_mis_gpt4_models}
\end{table*}

%% file: tables/tab_app_text_sample.tex
\begin{table*}[]
    \centering
    \caption{Three original samples and their corresponding $k$-iteration from \mone.}
    \begin{tabular}{p{0.95\textwidth}}
    \toprule
    Sample 1: \\
    \midrule
        \textbf{Original input}: Working in groups of 2 or 4 (even numbers may work better later on), suggest a heuristic to evaluate the utility of a given board configuration.\\
         \textbf{Iteration 1}: Working in groups of 2 or 4 (even numbers can work better later), suggests a heurist to evaluate the usefulness of a given table configuration. \\
          \textbf{Iteration 2}: Working in groups of 2 or 4 (even numbers can work better later), suggests a heurist to evaluate the usefulness of a given table configuration.\\
        \textbf{Iteration 3}: Working in groups of 2 or 4 (even numbers can work better later), suggests a heurist to evaluate the usefulness of a given table configuration.\\
        \textbf{Iteration 4}: Working in groups of 2 or 4 (even numbers can work better later), suggests a heurist to evaluate the usefulness of a given table configuration.\\
        \textbf{Iteration 5}: Working in groups of 2 or 4 (even numbers can work better later), suggests a heurist to evaluate the usefulness of a given table configuration.\\
        \midrule
     Sample 2: \\
    \midrule
        \textbf{Original input}: Therefore, to improve the ethical coping ability of Chinese NLP techniques and their application effectiveness in the field of ethics, the Chinese ethics knowledge base and KEPTMs for ethics domain are researched\\
        \textbf{Iteration 1}: Therefore, in order to improve the ethical treatment capacity of Chinese NLP techniques and their effectiveness of application in the field of ethics, the knowledge base of Chinese ethics and the KEPTMs for the field of ethics are studied.\\
        \textbf{Iteration 2}: Therefore, in order to improve the ethical treatment capacity of Chinese NLP techniques and their effectiveness of application in the field of ethics, the knowledge base of Chinese ethics and KEPTMs for the field of ethics are studied.\\
        \textbf{Iteration 3}: Therefore, in order to improve the ethical treatment capacity of Chinese NLP techniques and their effectiveness of application in the field of ethics, the knowledge base of Chinese ethics and KEPTMs for the field of ethics are studied.\\
         \textbf{Iteration 4}: Therefore, in order to improve the ethical treatment capacity of Chinese NLP techniques and their effectiveness of application in the field of ethics, the knowledge base of Chinese ethics and KEPTMs for the field of ethics are studied.\\
         \textbf{Iteration 5}: Therefore, in order to improve the ethical treatment capacity of Chinese NLP techniques and their effectiveness of application in the field of ethics, the knowledge base of Chinese ethics and KEPTMs for the field of ethics are studied.\\
        \midrule
        Sample 3: \\
    \midrule
        \textbf{Original input}: I kindly request a modification to the due date for my assignment. According to the previous thread, I was informed that I would receive an extension of 33 hours for this assignment. However, as of now, my Moodle account still indicates the original due date\\
         \textbf{Iteration 1}: I slowly ask for a change to the appropriate date for my task. According to the previous topic, I was informed that I would receive an extension of 33 hours for this task. However, from now on, my Moodle account still indicates the appropriate original date.\\
          \textbf{Iteration 2}: I slowly ask for a change to the appropriate date for my task. According to the previous subject, I was informed that I would receive a 33-hour extension for this task. However, from now on, my Moodle account still indicates the appropriate original date.\\
        \textbf{Iteration 3}: I slowly ask for a change to the appropriate date for my task. According to the previous topic, I was informed that I would receive a 33-hour extension for this task. However, from now on, my Moodle account still indicates the appropriate original date.\\
        \textbf{Iteration 4}: I slowly ask for a change to the appropriate date for my task. According to the previous topic, I was informed that I would receive a 33-hour extension for this task. However, from now on, my Moodle account still indicates the appropriate original date.\\
        \textbf{Iteration 5}: I slowly ask for a change to the appropriate date for my task. According to the previous topic, I was informed that I would receive an extension of 33 hours for this task. However, from now on, my Moodle account still indicates the appropriate original date.\\
        \bottomrule
    \end{tabular}
    \label{tab:example_m2m}
\end{table*}

\begin{table*}[]
    \centering
    \caption{Three original samples and their corresponding $k$-iteration from \mtwo.}
    \begin{tabular}{p{0.95\textwidth}}
    \toprule
    Sample 1: \\
    \midrule
        \textbf{Original input}: Working in groups of 2 or 4 (even numbers may work better later on), suggest a heuristic to evaluate the utility of a given board configuration.\\
         \textbf{Iteration 1}: If you work in groups of 2 or 4 (even numbers can work better later), suggest a mouse kit to evaluate the usefulness of a given configuration of the table.\\
          \textbf{Iteration 2}: If you work in groups of 2 or 4 (even numbers can work better later), suggest a mouse kit to evaluate the usefulness of a particular table configuration.\\
        \textbf{Iteration 3}: If you work in groups of 2 or 4 (even numbers can work better later), suggest a mouse kit to evaluate the usefulness of a particular table configuration.\\
        \textbf{Iteration 4}: If you work in groups of 2 or 4 (even numbers can work better later), suggest a mouse kit to evaluate the usefulness of a given table configuration.\\
        \textbf{Iteration 5}: If you work in groups of 2 or 4 (even numbers can work better later), suggest a mouse kit to evaluate the usefulness of a given table configuration.\\
        \midrule
     Sample 2: \\
    \midrule
        \textbf{Original input}: Therefore, to improve the ethical coping ability of Chinese NLP techniques and their application effectiveness in the field of ethics, the Chinese ethics knowledge base and KEPTMs for ethics domain are researched\\
        \textbf{Iteration 1}: For this reason, to improve the ethical adaptation capacity of Chinese NLP techniques and their effective application in the ethics field, the Chinese Ethics Knowledge Base and the KEPTMs for the ethics field are being studied.\\
        \textbf{Iteration 2}: Therefore, to improve the ethical adaptation capacity of Chinese NLP techniques and their effective application in the field of ethics, the Chinese ethics knowledge base and the ECEs for the field of ethics are being studied.\\
        \textbf{Iteration 3}: Consequently, in order to improve the ethical adaptation capacity of Chinese PLN techniques and their effective application in the field of ethics, the Chinese ethics knowledge base and ECE for the field of ethics are being studied.\\
         \textbf{Iteration 4}: Consequently, in order to improve the ethical adaptation capacity of Chinese PLN techniques and their effective application in the field of ethics, the Chinese ethics knowledge base and UNECE for the field of ethics are being studied.\\
         \textbf{Iteration 5}: Consequently, in order to improve the ethical adaptation capacity of Chinese PLN techniques and their effective application in the field of ethics, the Chinese ethics knowledge base and UNECE for the field of ethics are being studied.\\
        \midrule
        Sample 3: \\
    \midrule
        \textbf{Original input}: I kindly request a modification to the due date for my assignment. According to the previous thread, I was informed that I would receive an extension of 33 hours for this assignment. However, as of now, my Moodle account still indicates the original due date.\\
         \textbf{Iteration 1}: I am requesting a change to the due date of my assignment, and according to the previous thread, I have been informed that I will receive a 33-hour extension for this assignment, but at the moment my Moodle account still indicates the original due date.\\
          \textbf{Iteration 2}: I am requesting a change to my assignment due date and, according to the previous thread, I have been informed that I have received a 33-hour extension for this assignment, but at the moment, my Moodle account still indicates the original due date.\\
        \textbf{Iteration 3}: I am requesting a change to my assignment due date and, according to the previous thread, I have been informed that I have received a 33-hour extension for this assignment, but for the time being, my Moodle account still indicates the original due date.\\
        \textbf{Iteration 4}: I am requesting a change to the due date of my assignment and, according to the previous thread, I have been informed that I have received a 33-hour extension for this assignment, but for the time being, my Moodle account still indicates the original due date.\\
        \textbf{Iteration 5}: I am requesting a change to the due date of my assignment and, according to the previous thread, I was informed that I have received a 33-hour extension for this assignment, but for the time being, my Moodle account still indicates the original due date.\\
        \bottomrule
    \end{tabular}
    \label{tab:example_mbart}
\end{table*}

\begin{table*}[]
    \centering
    \caption{Three original samples and their corresponding $k$-iteration from \mthree.}
    \begin{tabular}{p{0.95\textwidth}}
    \toprule
    Sample 1: \\
    \midrule
        \textbf{Original input}: Working in groups of 2 or 4 (even numbers may work better later on), suggest a heuristic to evaluate the utility of a given board configuration.\\
         \textbf{Iteration 1}: Working in groups of 2 or 4 (even numbers might work better later on), propose a heuristic to evaluate the usefulness of a given board configuration.\\
          \textbf{Iteration 2}: Working in groups of 2 or 4 (even numbers might work better in the future), propose a heuristic to evaluate the usefulness of a given board configuration.\\
        \textbf{Iteration 3}: Working in groups of 2 or 4 (even numbers might work better in the future), propose a heuristic to evaluate the utility of a given board configuration.\\
        \textbf{Iteration 4}: Working in groups of 2 or 4 (even numbers might work better in the future), propose a heuristic to evaluate the usefulness of a given board configuration.\\
        \textbf{Iteration 5}: Working in groups of 2 or 4 (even numbers might work better in the future), propose a heuristic to evaluate the usefulness of a given board configuration.\\
        \midrule
     Sample 2: \\
    \midrule
        \textbf{Original input}: Therefore, to improve the ethical coping ability of Chinese NLP techniques and their application effectiveness in the field of ethics, the Chinese ethics knowledge base and KEPTMs for ethics domain are researched\\
        \textbf{Iteration 1}: Therefore, in order to improve the ethical adaptability of Chinese NLP techniques and their application efficiency in the field of ethics, the Chinese ethics knowledge base and KEPTMs for the field of ethics are studied.\\
        \textbf{Iteration 2}: Therefore, in order to improve the ethical adaptability of Chinese natural language processing techniques and their effectiveness in the field of ethics, Chinese ethical knowledge base and KEPTMs for the field of ethics are studied.\\
        \textbf{Iteration 3}: Therefore, in order to improve the ethical adaptability of Chinese natural language processing techniques and their effectiveness in the field of ethics, Chinese ethical knowledge base and KEPTMs for the field of ethics are studied.\\
         \textbf{Iteration 4}: Therefore, in order to improve the ethical adaptability of Chinese natural language processing techniques and their effectiveness in the field of ethics, Chinese ethical knowledge base and KEPTMs for the field of ethics are being studied.\\
         \textbf{Iteration 5}: Therefore, in order to improve the ethical adaptability of Chinese natural language processing techniques and their effectiveness in the field of ethics, Chinese ethical knowledge base and KEPTMs for the field of ethics are being studied.\\
        \midrule
        Sample 3: \\
    \midrule
        \textbf{Original input}: I kindly request a modification to the due date for my assignment. According to the previous thread, I was informed that I would receive an extension of 33 hours for this assignment. However, as of now, my Moodle account still indicates the original due date.\\
         \textbf{Iteration 1}: I politely request a modification of the due date for my assignment. According to the previous thread, I was informed that I would be granted a 33-hour extension for this assignment. However, at the moment, my Moodle account still shows the original due date.\\
          \textbf{Iteration 2}: I politely request a modification of the deadline for my assignment. According to the previous thread, I was informed that I would be granted an additional 33 hours for this assignment. However, at the moment, my Moodle account still displays the original deadline.\\
        \textbf{Iteration 3}: I politely request a change in the deadline for my assignment. According to the previous thread, I was informed that I would have an additional 33 hours for this assignment. However, currently, my Moodle account still displays the original deadline.\\
        \textbf{Iteration 4}: I politely request a change in deadline for my assignment. According to the previous thread, I was informed that I would have an additional 33 hours for this assignment. However, currently, my Moodle account still displays the original deadline.\\
        \textbf{Iteration 5}: I kindly request a change of deadline for my assignment. According to the previous thread, I was informed that I would have an additional 33 hours for this assignment. However, currently, my Moodle account still displays the original deadline.\\
        \bottomrule
    \end{tabular}
    \label{tab:example_chatgpt}
\end{table*}

\begin{table*}[]
    \centering
    \caption{Three original samples and their corresponding $k$-iteration from \mfour.}
    \begin{tabular}{p{0.95\textwidth}}
    \toprule
    Sample 1: \\
    \midrule
        \textbf{Original input}: Working in groups of 2 or 4 (even numbers may work better later on), suggest a heuristic to evaluate the utility of a given board configuration.\\
         \textbf{Iteration 1}: Working in groups of 2 or 4 (odd numbers may work better later), suggests a heuristic to evaluate the utility of a seating chart configuration.\\
          \textbf{Iteration 2}: Working in groups of 2 or 4 (odd numbers may work better later), suggests a heuristic for evaluating the utility of a configuration of the seating plan.\\
        \textbf{Iteration 3}: Working in groups of 2 or 4 (odd numbers can work better later), suggests an algorithm for evaluating the utility of a assistance plan configuration.\\
        \textbf{Iteration 4}: Working in groups of 2 or 4 (odd numbers can work better later), suggests an algorithm to evaluate the utility of a plan of assistance configuration.\\
        \textbf{Iteration 5}: Working in groups of 2 or 4 (odd numbers may work better later), suggests an algorithm to evaluate the utility of a configuration of assistance plans.\\
        \midrule
     Sample 2: \\
    \midrule
        \textbf{Original input}: Therefore, to improve the ethical coping ability of Chinese NLP techniques and their application effectiveness in the field of ethics, the Chinese ethics knowledge base and KEPTMs for ethics domain are researched\\
        \textbf{Iteration 1}: Therefore, to improve the ethical sweeping capacity of Chinese NLP techniques and their effectiveness in the ethical domain, Chinese knowledge bases on ethics and KEPTMs for the ethical domain are sought.\\
        \textbf{Iteration 2}: Therefore, to improve the ethical brushing capacity of Chinese NLP techniques and their effectiveness in the ethical domain, Chinese knowledge bases on ethics and KEPTMs for the ethical domain are sought.\\
        \textbf{Iteration 3}: Therefore, to improve the brushing capacity of Chinese NLP methods and their effectiveness in the ethical domain, Chinese knowledge bases on ethics and KEPTMs for the ethical domain are sought.\\
         \textbf{Iteration 4}: Therefore, to improve the brushing capacity of Chinese NLP methods and their effectiveness in the ethical domain, Chinese knowledge bases on ethics and KEPTMs for the ethical domain are sought.\\
         \textbf{Iteration 5}: Therefore, to improve the brushing capacity of Chinese NLP methods and their effectiveness in the ethical domain, Chinese knowledge bases on ethics and KEPTMs for the ethical domain are sought.\\
        \midrule
        Sample 3: \\
    \midrule
        \textbf{Original input}: I kindly request a modification to the due date for my assignment. According to the previous thread, I was informed that I would receive an extension of 33 hours for this assignment. However, as of now, my Moodle account still indicates the original due date.\\
         \textbf{Iteration 1}: Please extend the deadline for my assignment. I was informed that I will receive a 33 hour extension for this assignment, however, it still shows the initial date on my Moodle account.\\
          \textbf{Iteration 2}: Please extend the deadline for my assignment. I was informed that I will receive a 33 hour extension for this assignment, or this amount is still visible on my Moodle account.\\
        \textbf{Iteration 3}: Please extend the deadline for my assignment. I was informed that I will receive a 33 hour extension for this assignment, this quantity is always visible on my Moodle account.\\
        \textbf{Iteration 4}: Would you like to request an extension of the deadline for my mission? I am informed that I will obtain a 33 hour extension for this mission, this quantity is still visible on my Moodle account.\\
        \textbf{Iteration 5}: Did you want to request an extension of your mission deadline? I was informed that I will obtain a 33 hour extension for this mission, this amount is visible on my Moodle account.\\
        \bottomrule
    \end{tabular}
    \label{tab:example_cohere}
\end{table*}

\begin{table*}[]
    \centering
    \caption{Three original samples and their corresponding $k$-iteration from \mfive.}
    \begin{tabular}{p{0.95\textwidth}}
    \toprule
    Sample 1: \\
    \midrule
        \textbf{Original input}: Working in groups of 2 or 4 (even numbers may work better later on), suggest a heuristic to evaluate the utility of a given board configuration.\\
         \textbf{Iteration 1}: Working in groups of 2 or 4 (even numbers may work better later), propose a heuristic to evaluate the usefulness of a given board configuration.\\
          \textbf{Iteration 2}: Working in groups of 2 or 4 (even numbers may work better later), propose a heuristic to evaluate the usefulness of a given board configuration.\\
        \textbf{Iteration 3}: Working in groups of 2 or 4 (even numbers may work better later), propose a heuristic to evaluate the usefulness of a given board configuration.\\
        \textbf{Iteration 4}: Working in groups of 2 or 4 (even numbers may work better later), propose a heuristic to assess the usefulness of a given board configuration.\\
        \textbf{Iteration 5}: Working in groups of 2 or 4 (even numbers may work better later), propose a heuristic to evaluate the usefulness of a given board configuration.\\
        \midrule
     Sample 2: \\
    \midrule
        \textbf{Original input}: Therefore, to improve the ethical coping ability of Chinese NLP techniques and their application effectiveness in the field of ethics, the Chinese ethics knowledge base and KEPTMs for ethics domain are researched\\
        \textbf{Iteration 1}: Therefore, to improve the ethical management capacity of Chinese NLP techniques and their application effectiveness in the field of ethics, the Chinese ethical knowledge base and KEPTMs for the field of ethics are being studied.\\
        \textbf{Iteration 2}: Therefore, to improve the ethical management capacity of Chinese NLP techniques and their application effectiveness in the field of ethics, the Chinese ethical knowledge base and KEPTMs for the field of ethics are under study.\\
        \textbf{Iteration 3}: Therefore, to improve the ethical management capacity of Chinese NLP techniques and their application effectiveness in the field of ethics, the Chinese ethical knowledge base and KEPTMs for the field of ethics are currently under study.\\
         \textbf{Iteration 4}: Therefore, to improve the ethical management capacity of Chinese NLP techniques and their application efficiency in the field of ethics, the Chinese ethical knowledge base and KEPTMs for the field of ethics are currently under study.\\
         \textbf{Iteration 5}: Therefore, to improve the ethical management capacity of Chinese NLP techniques and their application efficiency in the field of ethics, the Chinese ethical knowledge base and KEPTMs for the field of ethics are currently under study.\\
        \midrule
        Sample 3: \\
    \midrule
        \textbf{Original input}: I kindly request a modification to the due date for my assignment. According to the previous thread, I was informed that I would receive an extension of 33 hours for this assignment. However, as of now, my Moodle account still indicates the original due date.\\
         \textbf{Iteration 1}: I kindly request a modification of the due date for my assignment. According to the previous thread, I was informed that I would receive a 33-hour extension for this assignment. However, at the present time, my Moodle account still shows the initial due date.\\
          \textbf{Iteration 2}: I kindly request a change in the due date for my assignment. According to the previous thread, I was informed that I would receive a 33-hour extension for this assignment. However, at present, my Moodle account still displays the initial due date.\\
        \textbf{Iteration 3}: I kindly request a change of the due date for my assignment. According to the previous thread, I was informed that I would receive a 33-hour extension for this assignment. However, at present, my Moodle account still displays the initial due date.\\
        \textbf{Iteration 4}: I am kindly requesting a change in the due date for my assignment. According to the previous thread, I was informed that I would receive a 33-hour extension for this task. However, at the present time, my Moodle account still displays the initial due date.\\
        \textbf{Iteration 5}: I kindly request a change of the due date for my assignment. According to the previous thread, I was informed that I would receive a 33-hour extension for this task. However, at the present time, my Moodle account still displays the initial due date.\\
        \bottomrule
    \end{tabular}
    \label{tab:example_gpt4}
\end{table*}